\documentclass[journal]{IEEEtran}
\usepackage{subfigure}
\usepackage{multirow}
\usepackage{color}
\usepackage{hyperref}
\usepackage[normalem]{ulem}
\usepackage{balance}
\usepackage{cases}
\usepackage{array}
\usepackage{bbding}
\usepackage{amssymb}
\usepackage{graphicx}
\usepackage{booktabs}
\usepackage{diagbox}
\usepackage{caption}

\ifCLASSINFOpdf
\else
\fi
\begin{document}

\title{Deep Unfolding Network for Image Compressed Sensing by Content-adaptive Gradient Updating and Deformation-invariant Non-local Modeling}

\author{Wenxue~Cui,
        Xiaopeng~Fan,~\IEEEmembership{Senior Member,~IEEE,}
        Jian Zhang,~\IEEEmembership{Member,~IEEE} \\
        Debin~Zhao,~\IEEEmembership{Member,~IEEE}
\thanks{This work was supported in part by the National Key Research and Development Program of China (2021YFF0900500), and the National Natural Science Foundation of China (NSFC) under grants U22B2035, 62272128. (Corresponding author: Debin Zhao.)} 
\thanks{Wenxue Cui, Xiaopeng Fan and Debin Zhao are with the Department
of Computer Science and Technology, Harbin Institute of Technology, Harbin, 150001, China and also with the Peng Cheng Laboratory, Shenzhen, 518052, China (e-mail: wxcui@hit.edu.cn; fxp@hit.edu.cn; dbzhao@hit.edu.cn).}
\thanks{Jian Zhang is with the School of Electronic and Computer Engineering, Peking University Shenzhen Graduate School, Shenzhen 518055, China and also with the Peng Cheng Laboratory, Shenzhen 518052, China (e-mail: zhangjian.sz@pku.edu.cn).}}

\markboth{IEEE Transactions on Multimedia}%
{Shell \MakeLowercase{\textit{et al.}}: Bare Demo of IEEEtran.cls for IEEE Journals}

\maketitle

\begin{abstract}
Inspired by certain optimization solvers, the deep unfolding network (DUN) has attracted much attention in recent years for image compressed sensing (CS). However, there still exist the following two issues: 1) In existing DUNs, most hyperparameters are usually content independent, which greatly limits their adaptability for different input contents. 2) In each iteration, a plain convolutional neural network is usually adopted, which weakens the perception of wider context prior and therefore depresses the expressive ability. In this paper, inspired by the traditional Proximal Gradient Descent (PGD) algorithm, a novel DUN for image compressed sensing (dubbed DUN-CSNet) is proposed to solve the above two issues. Specifically, for the first issue, a novel content adaptive gradient descent network is proposed, in which a well-designed step size generation sub-network is developed to dynamically allocate the corresponding step sizes for different textures of input image by generating a content-aware step size map, realizing a content-adaptive gradient updating. For the second issue, considering the fact that many similar patches exist in an image but have undergone a deformation, a novel deformation-invariant non-local proximal mapping network is developed, which can adaptively build the long-range dependencies between the nonlocal patches by deformation-invariant non-local modeling, leading to a wider perception on context priors. Extensive experiments manifest that the proposed DUN-CSNet outperforms existing state-of-the-art CS methods by large margins.

\end{abstract}

\begin{IEEEkeywords}
Image compressed sensing, proximal gradient descent (PGD), deep unfolding network, non-local neural network, convolutional neural networks (CNNs).
\end{IEEEkeywords}

\IEEEpeerreviewmaketitle

\section{Introduction}

\IEEEPARstart{C}{ompressed} sensing (CS)~\cite{donoho2006compressed, candes2008introduction}, as a powerful technique for signal acquisition, has attracted much attention over the past few years. Different from the sample-then-compress routine used in the traditional signal compression techniques, CS conducts a new paradigm for signal acquisition that performs signal sampling and compression simultaneously. The CS theory implies that if a signal is sparse in a certain domain, it can be reconstructed from much fewer linear measurements than that suggested by the Nyquist sampling theorem. Due to the simple and fast sampling, CS technique can effectively alleviate the demand for high transmission bandwidth and realize low-cost on-sensor signal compression. CS has been applied in diverse applications, including Magnetic Resonance Imaging (MRI)~\cite{Lustig2008Compressed}, sensor networks~\cite{Li2013Compressed}, snapshot compressive imaging~\cite{9363502}~\cite{9745932}.

Mathematically, given the input signal $\mathbf{x}\in\mathbb{R}^{N}$, the sampled linear measurements $\mathbf{y}\in\mathbb{R}^{M}$ can be acquired by $\mathbf{y}=\mathbf{\Phi} \mathbf{x}$, where $\mathbf{\Phi}\in \mathbb{R}^{M\times N}$ is called the sampling matrix and $\frac{M}{N}$ is the pre-defined CS sampling ratio. Emphatically, due to $M\ll N$, it is usually very hard to solve such ill-posed inverse problem, and some prior information about the signal is usually required to constrain the solution space. As noted above, the corresponding optimization model can be formulated as:
\vspace{-0.04in}
\begin{equation}\label{eq11}
\tilde{\mathbf{x}}=\mathop{\arg}\mathop{\min}_{\mathbf{x}}\mathcal{F}(\mathbf{x})+\lambda\mathbf{\Psi}(\mathbf{x})
\vspace{-0.02in}
\end{equation}
where the former item $\mathcal{F}(\mathbf{x})$ is the fidelity term and the latter one $\mathbf{\Psi}(\mathbf{x})$ indicates the regularization/prior term, and $\lambda$ is the regularization parameter to balance their contributions. In Eq.~\ref{eq11}, the fidelity term ensures the consistency between the possible solution and the target signal under CS sampling operation, and the regularization/prior term is used to guarantee that the possible solution satisfies the prior assumption. Specifically, for $\mathcal{F}(\mathbf{x})$, the Euclidean distance in the measurement domain is usually utilized, i.e.,
\vspace{-0.04in}
\begin{equation}\label{eq111}
\mathcal{F}(\mathbf{x})=\frac{1}{2} \| \mathbf{\Phi} \mathbf{x}-\mathbf{y} \|_{2}^{2}
\vspace{-0.02in}
\end{equation}

To solve Eq.~\ref{eq11}, many sparsity-regularized based methods have been proposed~\cite{gao2015block,Kim2010Compressed,Metzler2016From,6190204,7406719}, in which the prior term represents the sparsity in certain transform domains (such as DCT~\cite{6890254} and wavelet~\cite{7122281}). To further enhance the reconstructed performance, more well-designed regularizations are established, including minimal total variation~\cite{1580791,li2013tval3}, low rank~\cite{6827224,6288484} and non-local self-similarity image prior~\cite{9190055,zhang2014group}. By applying more sophisticated priors, many of these approaches have led to significant improvements. However, these optimization-based CS reconstruction algorithms usually require heavy computation, thus limiting CS applications.

Recently, fueled by the powerful learning ability of deep neural networks, many deep network-based image CS methods have been proposed. According to the interpretability, the existing CS networks can be roughly grouped into the following two categories: uninterpretable deep black box networks (DBNs) and interpretable deep unfolding networks (DUNs). \textbf{1) Uninterpretable DBNs:} this kind of method~\cite{2019DR2,7780424,cui2018efficient,8765626,shi2019scalable,9025255} usually trains the deep network as a black box, and builds a direct mapping from the compressed measurement domain to the original signal domain. Due to the simplicity and efficiency of such kind of algorithm, it has been widely studied in the early stage of deep network-based CS research. Unfortunately, this rude mapping strategy usually lacks a theoretical interpretation, thus weakening the interpretability and limiting the reconstructed quality. \textbf{2) Interpretable DUNs:} this kind of method~\cite{2017Learned,8578294,9019857,9298950,9467810} usually unfolds certain optimization algorithms, such as iterative shrinkage-thresholding algorithm (ISTA)~\cite{DBLP-ISTA}, half quadratic splitting (HQS)~\cite{8099783} and approximate message passing (AMP)~\cite{10124848}, into deep networks to enjoy a good interpretability. Specifically, inspired by the perspective of the iterative optimization, DUNs usually inherit a well-designed cascaded multi-stage structure to gradually reconstruct the target signal. By unfolding the optimization solvers, this kind of method apparently enjoys solid theoretical support and better interpretability.

Compared to DBNs, the recent DUNs have become the mainstream for CS reconstruction. However, there still exist the following two issues: 1) In existing DUNs, once the training is completed, most hyperparameters (e.g., the step size~\cite{9019857,2021Memory} and the control parameter~\cite{9298950}) are fixed for any input content, which limits the adaptive ability of these models. 2) For each iteration in DUNs, a plain stacked convolutional network is usually adopted, which weakens the perception of wider context prior and therefore depresses the expressiveness of these DUNs for image reconstruction.

\begin{figure*}
\begin{center}
\includegraphics[width=6.9in]{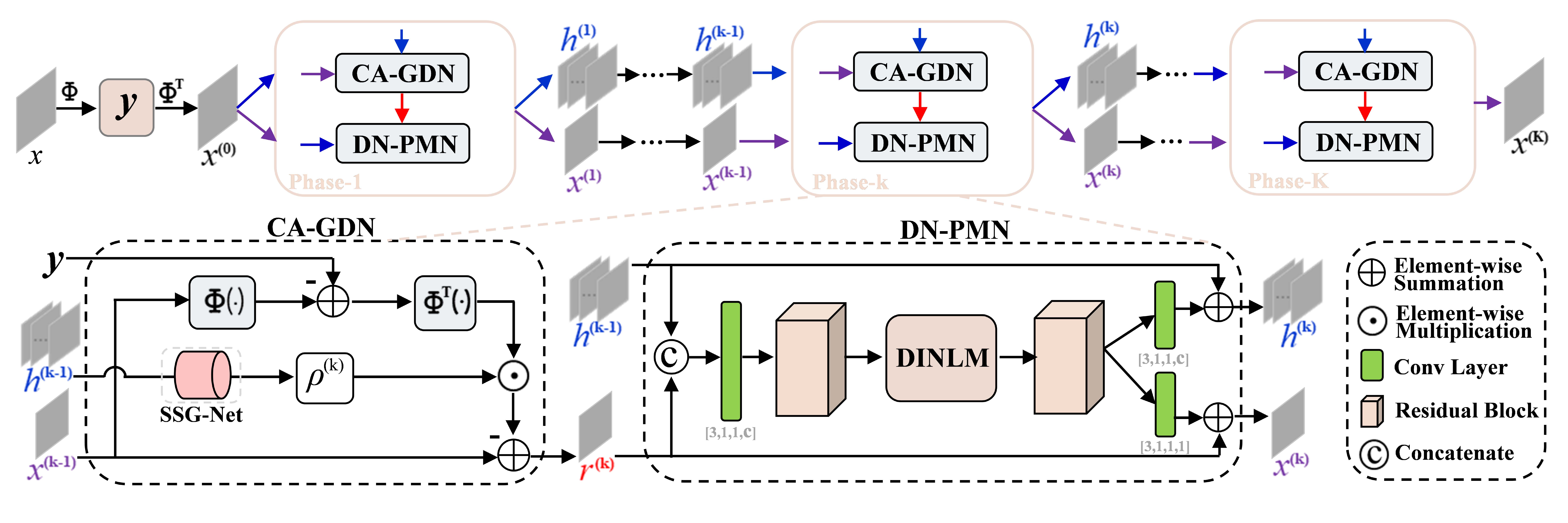}
\end{center}
\vskip -0.16in
   \caption{Illustration of the proposed DUN-CSNet. The overall architecture is shown in the first row, which cascades multiple phases, and each phase corresponds to an iteration in the PGD algorithm. More details of $k$-th phase are presented in the second row, containing a content adaptive gradient descent network (CA-GDN) and a deformation-invariant non-local proximal mapping network (DN-PMN). Specifically, in CA-GDN, SSG-Net represents the designed step size generation sub-network. In DN-PMN, DINLM indicates the proposed deformation-invariant non-local sub-module. The elements in the tuples next to the convolutional layers signify: [kernel size, stride size, padding size, kernel number]. Blue and purple arrows respectively indicate the flow of intermediate feature $\mathbf{h}^{(\cdot)}$ and reconstructed results $\mathbf{x}^{(\cdot)}$. Red arrows indicate the flow of intermediate result $\mathbf{r}^{(\cdot)}$ output from the network CA-GDN.}.

\vskip -0.22in
\label{fig:1}
\end{figure*}

To address the above issues, a novel deep unfolding image CS network (DUN-CSNet) is proposed (as shown in Fig.~\ref{fig:1}) in this paper. Inspired by the traditional Proximal Gradient Descent (PGD) algorithm, the proposed DUN-CSNet unfolds PGD into multiple phases and cascades them together. Each phase consists of two interactive networks: content adaptive gradient descent network (CA-GDN) and deformation-invariant non-local proximal mapping network (DN-PMN). Specifically, in CA-GDN, a novel step size generation sub-network (SSG-Net) is designed, which is able to dynamically allocate the corresponding step sizes for different textures of input image by generating a content-aware step size map, realizing a content-adaptive gradient updating. In DN-PMN, a novel deformation-invariant deep non-local sub-module (DINLM) is presented, which can adaptively build the long-range dependencies between the non-local patches by deformation-invariant non-local modeling, leading to a wider perception on context priors.

The main contributions are summarized as follows:

\textbf{1)} Inspired by the proximal gradient descent (PGD) algorithm, a novel deep unfolding image CS network (DUN-CSNet) is proposed, in which the newly designed networks, i.e., CA-DGN and DN-PMN, improve the CS performance significantly by content-adaptive gradient updating and deformation-invariant non-local modeling. 

\textbf{2)} For gradient descent, a novel content adaptive gradient descent network (CA-GDN) is proposed, in which a well-designed step size generation sub-network (SSG-Net) is developed to dynamically allocate the corresponding step sizes for different textures of input image, realizing a content-adaptive gradient updating and a powerful adaptability.

\textbf{3)} For proximal mapping, a novel deformation-invariant non-local proximal mapping network (DN-PMN) is designed, which can adaptively build the long-range dependencies between the nonlocal patches by deformation-invariant non-local modeling, leading to a wider perception on context priors.

The remainder of this paper is organized as follows: Section~\ref{section:a2} reviews the recent related works. Section~\ref{section:a3} elaborates the proposed framework. Section~\ref{section:a4} illustrates the experimental details and Section~\ref{section:a6} concludes the paper. 

\vspace{-0.03in}
\section{Background and Related Works}
In this paper, we mainly focus on the deep network-based image compressed sensing task. Besides, we further explore the deep non-local knowledge to improve the expressiveness of our CS model. As noted above, the related works in this section are summarized from the following two aspects: 1) image compressed sensing using deep networks and 2) non-local prior exploiting using deep networks.
\label{section:a2}

\vspace{-0.1in}
\subsection{Image Compressed Sensing Using Deep Networks}

According to the interpretability, the existing image CS networks can be roughly divided into the following two groups: 1) Uninterpretable deep black box networks (DBNs) and 2) Interpretable deep unfolding networks (DUNs).

\textbf{Uninterpretable DBNs:} With the powerful learning ability of deep networks, this kind of CS method~\cite{2019DR2,7780424,7447163} usually builds a direct mapping from the compressed measurement domain to the original image domain. Due to the simplicity and efficiency, this kind of method is widely favored by many researchers. Specifically, the early works~\cite{7780424,2018Convolutional,2019DR2} usually first reconstruct the image blocks from the corresponding measurements and then splice all these reconstructed image blocks together into a final image. However, these block-by-block reconstruction methods usually suffer from serious block artifacts (especially at low sampling rates)~\cite{cui2018efficient}. In order to solve this problem, some works~\cite{7780424,2018Convolutional,2019DR2} try to append a de-blocking module (such as BM3D~\cite{4271520}) after these methods, which still cannot obtain satisfactory reconstructed quality in most cases.

To further remove the block artifacts and enhance the reconstruction performance, some CS literatures~\cite{8765626,cui2018efficient,9199540,8019428} attempt to explore the latent deep image priors in the whole image space. Specifically, these CS methods still perform the image sampling in a block-by-block manner, while during the reconstruction, they first concatenate all image blocks together in the initial reconstruction, and then carry out a deep reconstruction in the whole image space. More recently, to enhance the flexibility of CS model, several scalable network architectures~\cite{shi2019scalable,xu2018lapran,9467810} are designed, which are able to realize scalable sampling and reconstruction with only single model. By inferring deep networks in the whole image space, these CS algorithms mitigate the block artifacts and achieve much higher reconstructed quality.

However, the aforementioned CS networks usually train the deep network as a black box, which apparently makes these methods lack the theoretical interpretation. Besides, in these deep black box CS networks, the sampling matrix is not used in the reconstruction process, which results in an insufficient guidance for image reconstruction, thus limiting the reconstructed quality.

\textbf{Interpretable DUNs:} Inspired by some model-based solvers, this kind of CS method usually unfolds certain iterative optimizers into deep neural networks to enjoy a
better interpretability. For instance, ADMM-CSNet~\cite{8550778} casts the iterative Alternating Direction Method of Multipliers (ADMM) algorithm into a deep network architecture for image CS reconstruction. AMP-Net~\cite{9298950} solves the image CS problem by unrolling the iterative denoising process of the approximate message passing (AMP) algorithm~\cite{donoho2009message}. Recently, some DUNs~\cite{8578294,9467810,10098557} unfold the traditional Proximal Gradient Descent (PGD) algorithm into network forms to enjoy the interpretability. Mathematically, the PGD algorithm solves Eq.~\ref{eq11} through the following iterative steps:
\vspace{-0.03in}
\begin{numcases}{}
\mathbf{r}^{(k)} = \mathbf{x}^{(k-1)} - \rho\mathbf{\Phi}^{\rm T}(\mathbf{\Phi} \mathbf{x}^{(k-1)}-\mathbf{y}) \label{eq22} \\
\mathbf{x}^{(k)} = \mathcal{\rm{prox}}_{\lambda,\mathbf{\Psi}}(\mathbf{r}^{(k)}) \label{eq33}
\end{numcases}
where Eq.~\ref{eq22} is responsible for the gradient descent of the fidelity term (in Eq.~\ref{eq11}) and $\rho$ is the pre-defined step size. In Eq.~\ref{eq33}, ${\rm prox}_{\lambda, \mathbf{\Psi}}(\cdot)$ indicates the corresponding proximal operator, which is highly related to the regularization term of Eq.~\ref{eq11}. Inspired by Eqs.~\ref{eq22}~\ref{eq33}, the existing PGD-based DUNs attempt to embed deep networks into the PGD algorithm to solve CS problem by iterating the following updating steps:
\vspace{-0.03in}
\begin{numcases}{}
\mathbf{r}^{(k)} = \mathbf{x}^{(k-1)} - \rho^{(k)}\mathbf{\Phi}^{\rm T}(\mathbf{\Phi} \mathbf{x}^{(k-1)}-\mathbf{y}) \label{eq44} \\
\mathbf{x}^{(k)} = \mathcal{H}_{\lambda,\mathbf{\Psi}}^{(k)}(\mathbf{r}^{(k)}) \label{eq55}
\end{numcases}
where Eq.~\ref{eq44} indicates the gradient descent process of the current $k$-th iteration and $\rho^{(k)}$ is the corresponding learnable step size. Corresponding to the proximal operator (in ~\ref{eq33}), Eq.~\ref{eq55} indicates a specific deep neural network (i.e., $\mathcal{H}_{\lambda,\mathbf{\Psi}}^{(k)}$) to learn a deep proximal mapping. 

Specifically, depending on Eqs.~\ref{eq44} and~\ref{eq55}, Zhang \emph{et al.}~\cite{8578294} propose a novel deep unfolding network (dubbed ISTA-Net) based on the Iterative Shrinkage-Thresholding Algorithm (ISTA\footnote{ISTA is a typical PGD-based algorithm, in which the regularization term of the Eq.~\ref{eq11} is defined as an $L_{1}$ norm, i.e., $\mathbf{\Psi}(\mathbf{x})=\|\mathbf{x}\|_{1}$.}) for image CS reconstruction. However, the performance of~\cite{8578294} is greatly limited because of its random sampling and block-by-block reconstruction. To further enhance the reconstructed quality, several DUN variants~\cite{9019857,9467810,2021Memory} on the basis of~\cite{8578294} are subsequently proposed, which attempt to optimize the sampling matrix in the sampling process and embed a well-designed de-blocking strategy into the reconstruction module. Apparently, by unfolding the optimization-based solvers, these deep unfolding methods have better interpretability, but these algorithms usually adopt a simple stacked convolutional network, which weakens the perception of wider context information and therefore limits the expressiveness of these models for image reconstruction. Besides, in these DUNs, once the training is completed, most hyperparameters (e.g., the step size~\cite{9019857,9467810,2021Memory} and the control parameter~\cite{9298950}) remain unchanged for any input content, which limits the adaptive ability of these models.

\vspace{-0.1in}
\subsection{Non-local Prior Exploiting Using Deep Networks}
\vspace{-0.01in}

\begin{figure*}
\begin{center}
\includegraphics[width=6.8in]{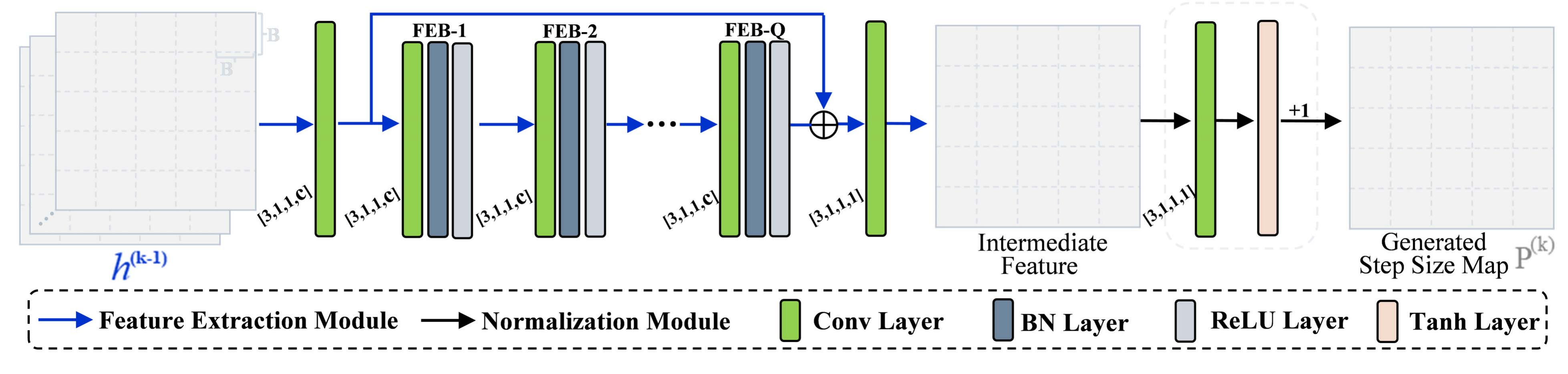}
\end{center}
\vskip -0.15in
   \caption{The network architecture details of the proposed step size generation sub-network (SSG-Net), in which two modules, i.e., feature extraction module and normalization module are included. FEB is short for feature extraction block, in which three layers, i.e., a Conv layer, a Batch Norm layer and a ReLU layer are included. The elements in the tuples next to the convolutional layers signify: [kernel size, stride size, padding size, kernel number].}
\vskip -0.15in
\label{fig:2}
\end{figure*}

Inspired by the non-local means operation~\cite{1467423}, the non-local self-similarity image prior has been extensively studied for diverse low-level vision tasks~\cite{2011Nonlocal,5459452}, which depicts the repetitiveness of higher level patterns (e.g., textures and structures) globally positioned in images. Recently, the non-local self-similarity image prior is also applied in some image CS literatures~\cite{dong2014compressive,6341094}. For example, Zhang \emph{et al.}~\cite{zhang2014group} establish a novel sparse representation model of natural images by exploring the non-local patches with the similar structures. Zhao \emph{et al.}~\cite{7786160} exploit the non-local self-similarity patches and propose a low-rank based CS reconstruction model. However, these non-local prior based algorithms are all optimization-based CS methods, which still run very slowly because of their hundreds of iterations.

Recently, the non-local neural networks have been proposed. For example, Wang \emph{et al.}~\cite{2018Non} propose a differentiable non-local neural network, which able to capture the long-range dependencies among non-local information in a feed-forward fashion. Motivated by this work, Li \emph{et al.}~\cite{8721638} propose a residual network with nonlocal constraint for image CS reconstruction. The non-local operator in~\cite{8721638} only perceives the self-similarities inside the current image block. Subsequently, Sun \emph{et al.}~\cite{8999514} propose a non-locally regularized CS network, in which the non-local prior and the deep network prior~\cite{8579082} are both utilized to enhance the reconstruction performance. Unfortunately, the network in~\cite{8999514} needs to be trained online in an iterative mode for each input image, which leads to a deficient flexibility, thus hindering its application. More recently, Cui \emph{et al.}~\cite{9635679} develop a novel image CS framework using non-local neural network (NL-CSNet), which utilizes the non-local priors in the measurement domain and multi-scale feature domain to improve the reconstruction quality. However, the sampling matrix of~\cite{9635679} is not used in its reconstruction process, therefore influences its performance.

In addition to the non-local neural network~\cite{2018Non} or its recent variants~\cite{9008832,9133304}, many self-attention based deep non-local models, such as Transformer-based non-local model~\cite{vit,9710580} and Graph convolutional network-based non-local model~\cite{9159868,8803367}, also attract extensive attentions and achieve impressive effects in diverse computer vision tasks~\cite{9710580,detrs}. It is worth noting that the non-local models mentioned above usually directly build the long-range dependencies between non-local patches (in embedding space) by measuring their explicit similarity, resulting in a weak modeling ability of non-local self-similarity dependences in implicit space.

\vspace{-0.05in}
\section{The Proposed Deep Unfolding Image CS Network}

\label{section:a3}

In this section, we first give an overview of the proposed deep unfolding CS framework DUN-CSNet, and then detail the sampling process and initial reconstruction. After that, the content-adaptive gradient updating is presented. Finally, the deformation-invariant non-local modeling is described.

\begin{figure*}
\begin{center}
\includegraphics[width=6.8in]{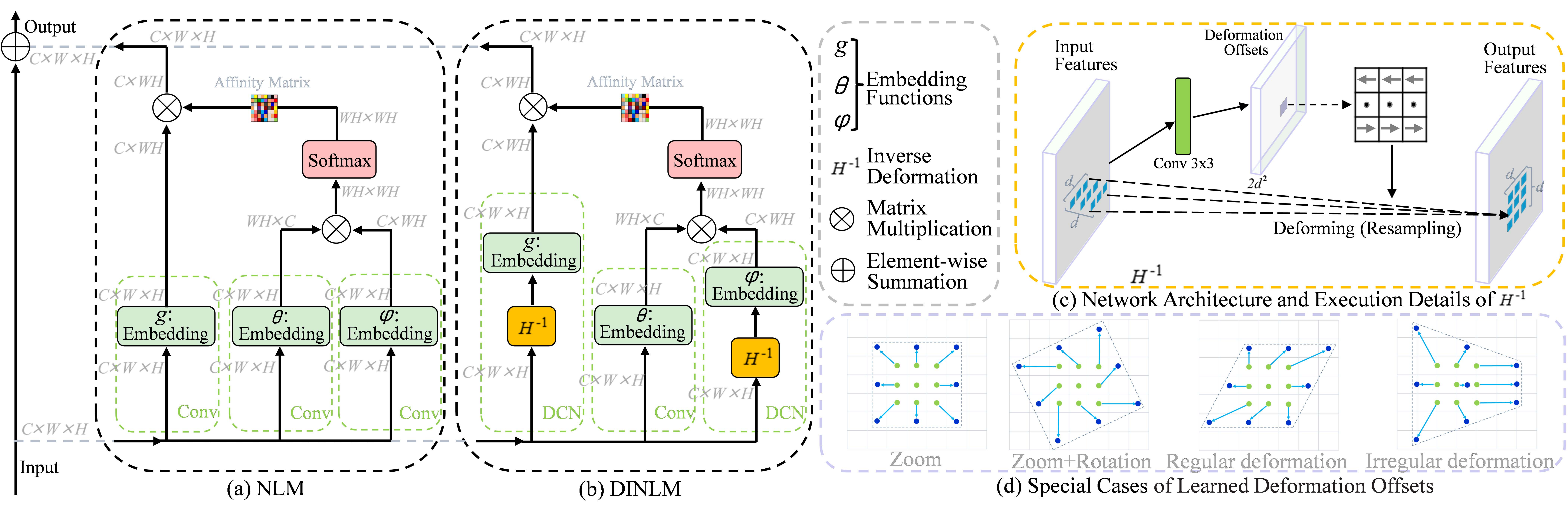}
\end{center}
\vskip -0.16in
   \caption{The figures (a) and (b) show the network architecture details of the traditional non-local sub-module (NLM) and the proposed deformation-invariant non-local sub-module (DINLM). The figure (c) shows the network architecture and execution details of $H^{-1}$ in our proposed DINLM, and the figure (d) shows some special cases of learned deformation offsets, in which the green patches (consisting of green points) and blue patches (consisting of blue points) are similar in a certain deformation domain, and the arrows represent corresponding deformation offsets.}
\vskip -0.15in
\label{fig:3}
\end{figure*}

\vspace{-0.1in}
\subsection{Overview of \rm{DUN-CSNet}}
\vspace{-0.01in}
Figure~\ref{fig:1} shows the whole network structure of the proposed DUN-CSNet. For image sampling, the measurements are acquired by using a sampling matrix $\mathbf{\Phi}$ in a block-by-block sampling manner. For image reconstruction, inspired by the traditional iterative proximal gradient descent (PGD) algorithm, the proposed CS reconstruction network cascades multiple ($K$) phases to gradually reconstruct the target image. Corresponding to the inherent iterative steps, i.e., gradient descent and proximal operator, in PGD algorithm, each phase of the proposed DUN-CSNet consists of two interactive networks: Content Adaptive Gradient Descent Network (CA-GDN) and Deformation-invariant Non-local Proximal Mapping Network (DN-PMN).

Specifically, CA-GDN is responsible for the gradient descent of the fidelity term in the Eq.~\ref{eq11}, and DN-PMN aims to fit the proximal operator using neural networks for exploring non-local image priors. To enhance the expressiveness and adaptability of the proposed DUN-CSNet, some well-designed deep modules are delicately embedded into the proposed framework. More specifically, in CA-GDN, considering the different learning abilities of deep networks on different textures in an image, a novel step size generation sub-network (SSG-Net) is designed, which is able to dynamically allocate the corresponding step sizes for different textures of input image by generating a content-aware step size map, realizing a content-adaptive gradient updating. In DN-PMN, we first aggregate the deep features of contiguous phases to facilitate the information transmission within the entire cascaded structure, and then the proposed deformation-invariant deep non-local network is followed, which can adaptively exploit the long-range dependencies between the non-local patches by deformation-invariant non-local modeling.

\vspace{-0.04in}
\subsection{Sampling and Initial Reconstruction}

For image sampling, the image $\mathbf{x}$ is first divided into non-overlapping blocks $\mathbf{x}_{(i,j)}$ of size $B\times B$, where $i$ and $j$ are the position indexes of the image blocks. Then a sampling matrix $\mathbf{\Phi}$ of size $n_{B}\times B^{2}$ is used to acquire the measurements, i.e., $\mathbf{y}_{(i,j)} = \mathbf{\Phi} \mathbf{x}_{(i,j)}$, where $n_{B} = \frac{M}{N}B^{2}$ and $\frac{M}{N}$ is the pre-defined sampling ratio. Because each row of the sampling matrix can be considered as a filter~\cite{8765626}, we can use the convolutional operation to perform the sampling process. Specifically, the convolutional kernel actually is the reshaped version of the sampling matrix (kernel size is $B\times B$ and kernel number is $n_{B}$), and the stride is set as $B\times B$, which ensures the non-overlapping sampling of all image blocks for the given input image $\mathbf{x}$. The process can be expressed as $\mathbf{y}=\mathbf{\Phi} * \mathbf{x}$, where $*$ is the convolutional operator and output $\mathbf{y}$ contains all measurements $\{\mathbf{y}_{(i,j)}\}$ of all image blocks.

After sampling process, an initial reconstruction operation is subsequently performed to produce the initial reconstructed image. Specifically, given the measurement $\mathbf{y}_{(i,j)}$ of image block $\mathbf{x}_{(i,j)}$, an upsampling operation is first performed by $\mathbf{x}_{(i,j)}^{(0)}=\mathbf{\Phi}^{\rm T}\mathbf{y}_{(i,j)}$, where $\rm T$ indicates the transposition of a matrix and the response $\mathbf{x}_{(i,j)}^{(0)}$ is obviously an upsampled vector of size $1\times B^{2}$. After this upsampling process, we then reshape each $\mathbf{x}_{(i,j)}^{(0)}$ (for all image blocks) into a $B\times B$ tensor block and finally concatenate all these reshaped blocks together to output the final initial reconstructed image $\mathbf{x}^{(0)}$. After the initial reconstruction, the proposed deep unfolding network with multiple cascaded phases is followed to further enhance the reconstructed quality.

\vspace{-0.09in}
\subsection{Content-adaptive Gradient Updating}
\vspace{-0.01in}

For gradient descent, the step size hyperparameter is mainly used for controlling the intensity of gradient updating, which actually affects the stability of the entire iterative process, thus influencing the convergence speed and reconstructed quality to a certain extent. However, in most recent PGD-based DUNs~\cite{8578294,9019857,9467810,2021Memory}, the step size $\rho^{(k)}$ (in Eq.~\ref{eq44}) is usually content independent. That is to say, once the training is completed, the step sizes $\rho^{(k)}$ of these methods usually remain unchanged for any input content, which greatly limits their adaptability. Besides, the learned step size of each phase in existing DUNs is a scalar value, which implies that the same step size is actually utilized for all different textures of input image. In fact, different textural contents in an image usually have different learning attributes, for example, the smooth area is generally easier to learn than the textural area. As above, a novel step size generation sub-network (SSG-Net) is developed in our framework to dynamically generate the step sizes for different textures of input image:
\vspace{-0.02in}
\begin{equation}\label{eq66}
\mathbf{P}^{(k)} = \mathcal{G}^{(k)}(\mathbf{h}^{(k-1)})
\vspace{-0.04in}
\end{equation}
where $\mathcal{G}^{(k)}(\cdot)$ indicates the designed sub-network SSG-Net, and $\mathbf{P}^{(k)}$ is the generated step size map, in which different elements indicate the step sizes of different textures. As shown in Fig.~\ref{fig:1}, $\mathbf{h}^{(k-1)}$ signifies the intermediate feature output from the previous ($k$-$1$)-th phase, and the utilizing of $\mathbf{h}^{(k-1)}$ in the current $k$-th phase strengthens the inferential cooperation among cascaded phases. As above, by bring Eq.~\ref{eq66} into Eq.~\ref{eq44}, a novel content adaptive gradient descent operation is obtained, which can be expressed as:
\vspace{-0.02in}
\begin{equation}\label{eq77}
\mathbf{r}^{(k)} = \mathbf{x}^{(k-1)} - \mathcal{G}^{(k)}(\mathbf{h}^{(k-1)})\mathbf{\Phi}^{\rm T}(\mathbf{\Phi} \mathbf{x}^{(k-1)}-\mathbf{y}) \\
\vspace{-0.04in}
\end{equation}
Obviously, different from the existing DUNs, the developed step size generation sub-network $\mathcal{G}^{(k)}(\cdot)$ not only can dynamically tune up the step sizes of current phase depending on the intermediate feature $\mathbf{h}^{(k-1)}$ of the previous phase, but also is able to adaptively allocate the corresponding step sizes for different textures of input image.

For the network structure of SSG-Net (as shown in Fig.~\ref{fig:2}), two modules, i.e., feature extraction module and normalization module, are included, which are respectively responsible for feature extraction and step size normalization. Specifically, in the feature extraction module, a convolutional layer with $c$ kernels is first performed, and then several (${\rm Q}$) feature extraction blocks (FEB) are followed with a skip connection, finally another convolutional layer with a single kernel is appended to produce the output intermediate feature. In the normalization module, a convolutional layer and a Tanh layer are sequentially executed. It is noted that because the lower bound of $Tanh$ function is -1, we add 1 after Tanh layer to ensure the nonnegativity of the generated step sizes. It is worth noting that there is no any up/down sampling in the designed step size generation sub-network (SSG-Net), which ensures the output is a step size map with the same size as the input, and the elements in this map correspond to the step sizes of different textures of input image.

\vspace{-0.05in}
\subsection{Deformation-invariant Non-local Modeling}

For proximal mapping, a novel deformation-invariant non-local proximal mapping network is proposed in our framework, which is able to adaptively build the long-range dependencies between the nonlocal context information under certain automatically learned deformations. For simplicity, we first model the proposed deformation-invariant non-local operation, and then attempt to package it into a specific network module. Finally, more structural details of the entire proximal mapping network are explained.

Given the current signal patch ${\mathbf{x}}_{i}$ and according to the existing non-local operation~\cite{2018Non}, the referenced information of $\mathbf{x}_{i}$ by referencing the other signal patches can be expressed as:
\vspace{-0.09in}
\begin{equation}
\hat{\mathbf{x}}_{i} = \frac{1}{\mathcal{C}({\mathbf{x}})}\sum_{\forall j}{W}_{\hspace{-0.03in}i\hspace{-0.01in}j}g({\mathbf{x}}_{j}), \ { W}_{\hspace{-0.03in}i\hspace{-0.01in}j}=f({\mathbf{x}}_{i}, {\mathbf{x}}_{j})
\label{eq:1}
\vspace{-0.1in}
\end{equation}
where $f$ is a pairwise function to compute the affinity coefficient ${ W}_{\hspace{-0.03in}i\hspace{-0.01in}j}$ between ${\mathbf{x}}_{i}$ and ${\mathbf{x}}_{j}$. The unary function $g$ is used to compute a new representation of ${\mathbf{x}}_{j}$ and $\mathcal{C}({\mathbf{x}})$ is the normalization factor. Obviously, when ${\mathbf{x}}_{j}$ is similar to ${\mathbf{x}}_{i}$, a higher affinity coefficient ${W}_{\hspace{-0.03in}i\hspace{-0.01in}j}$ can be obtained, so that more information can be referenced from ${\mathbf{x}}_{j}$. It is noted that in~\cite{2018Non}, all affinity coefficients ${ W}_{\hspace{-0.03in}i\hspace{-0.01in}j}$ can be integrated into a matrix, and we call this matrix as affinity matrix in our paper.

To explicate the proposed deformation-invariant non-local operation, we first define that the patch ${\mathbf{X}}_{\hspace{-0.01in}j}$ is a deformed version of ${\mathbf{x}}_{i}$ in the current image, i.e., ${\mathbf{X}}_{\hspace{-0.01in}j}$$ =$$ \mathcal{H}_{ij}({\mathbf{x}}_{i})$, where $\mathcal{H}_{ij}$ is the corresponding deformation operator. The proposed deformation-invariant non-local can be expressed as:
\vspace{-0.06in}
\begin{equation}
\hat{\mathbf{x}}_{i}\hspace{-0.04in} =\hspace{-0.04in} \frac{1}{\mathcal{C}({\mathbf{x}})}\hspace{-0.03in}\sum_{\forall j}{ W}_{\hspace{-0.03in}i\hspace{-0.01in}j}g(\mathcal{H}_{ij}^{-1}\hspace{-0.02in}({\mathbf{X}}_{j})), { W}_{\hspace{-0.03in}i\hspace{-0.01in}j}\hspace{-0.04in} =\hspace{-0.04in} f({\mathbf{x}}_{i},\hspace{-0.03in} \mathcal{H}_{ij}^{-1}\hspace{-0.02in}({\mathbf{X}}_{j}))
\label{eq:2}
\vspace{-0.08in}
\end{equation}
where $\mathcal{H}_{ij}^{-1}$ indicates the approximate inverse deformation of $\mathcal{H}_{ij}$, i.e., $\check{\mathbf{x}}_{ij}$$ =$$ \mathcal{H}_{ij}^{-1}({\mathbf{X}}_{j})$, and $\check{\mathbf{x}}_{ij}$ is the deformed response of ${\mathbf{X}}_{j}$. Clearly, the Eq.~\ref{eq:1} is a special example of the Eq.~\ref{eq:2} in the case of $\mathcal{H}_{ij}^{-1}$ is an identity transformation. Therefore, the proposed deformation-invariant non-local operation (shown in Eq.~\ref{eq:2}) can be considered as an extended universal formulation for exploiting non-local self-similarity prior.

\begin{table*}[t]
\centering
\caption{Average PSNR and SSIM comparisons of different deep network-based CS algorithms using learned sampling matrix at diverse sampling rates on dataset Set11. Bold indicates the best result, and underline signifies the second-best result.}
\label{tab:1}
\vspace{-0.1in}
\small
\begin{tabular}{p{3.25cm}<{\centering} | p{0.75cm}<{\centering} p{0.8cm}<{\centering} | p{0.75cm}<{\centering}  p{0.8cm}<{\centering} | p{0.75cm}<{\centering} p{0.8cm}<{\centering} | p{0.75cm}<{\centering} p{0.8cm}<{\centering} | p{0.75cm}<{\centering} p{0.8cm}<{\centering} | p{0.75cm}<{\centering} p{0.8cm}<{\centering}}
\toprule
\multirow{2}*{Algorithms} & \multicolumn{2}{c}{Rate=0.01} & \multicolumn{2}{c}{Rate=0.10} & \multicolumn{2}{c}{Rate=0.25} & \multicolumn{2}{c}{Rate=0.30} & \multicolumn{2}{c}{Rate=0.40} & \multicolumn{2}{c}{Avg.}\\
\cline{2-13}
&PSNR&SSIM&PSNR&SSIM&PSNR&SSIM&PSNR&SSIM&PSNR&SSIM&PSNR&SSIM\\

\midrule

\footnotesize{CSNet}\tiny{$_{\rm \textcolor{red}{(ICME2017)}}$}\footnotesize{~\cite{8019428}}&21.01&0.5560&28.10&0.8514&32.10&0.9221&33.86&0.9448&35.88&0.9605&30.19&0.8470\\
\footnotesize{LapCSNet}\tiny{$_{\rm \textcolor{red}{(ICASSP2018)}}$}\footnotesize{~\cite{cui2018efficient}}&\underline{21.54}&0.5659&28.34&0.8571&--\ --&--\ --&--\ --&--\ --&--\ --&--\ --&--\ --&--\ --\\
\footnotesize{SCSNet}\tiny{$_{\rm \textcolor{red}{(CVPR2019)}}$}\footnotesize{~\cite{shi2019scalable}}&21.04&0.5562&28.52&0.8616&33.43&0.9373&34.64&0.9511&36.92&0.9666&30.91&0.8546\\
\footnotesize{CSNet$^{+}$}\tiny{$_{\rm \textcolor{red}{(TIP2020)}}$}\footnotesize{~\cite{8765626}}&21.03&0.5566&28.34&0.8580&33.34&0.9387&34.27&0.9492&36.44&0.9690&30.68&0.8543\\
\footnotesize{NL-CSNet}\tiny{$_{\rm \textcolor{red}{(TMM2021)}}$}\footnotesize{~\cite{9635679}}&\textbf{21.96}&\textbf{0.6005}&\underline{30.05}&\underline{0.8995}&34.45&0.9513&35.68&0.9606&37.71&0.9753&31.97&\underline{0.8774}\\
\hline
\footnotesize{BCS-Net}\tiny{$_{\rm \textcolor{red}{(TMM2020)}}$}\footnotesize{~\cite{9159912}}&20.88&0.5505&29.43&0.8676&34.20&0.9408&35.63&0.9495&37.27&0.9706&31.48&0.8558\\
\footnotesize{OPINENet$^{+}$}\tiny{$_{\rm \textcolor{red}{(JSTSP2020)}}$}\footnotesize{~\cite{9019857}}&20.02&0.5362&29.81&0.8904&34.81&0.9514&36.30&0.9615&38.32&0.9722&31.85&0.8623\\
\footnotesize{AMP-Net$^{+}$}\tiny{$_{\rm \textcolor{red}{(TIP2021)}}$}\footnotesize{~\cite{9298950}}&20.20&0.5581&29.42&0.8782&34.60&0.9469&35.91&0.9576&38.25&0.9714&31.68&0.8624\\
\footnotesize{COAST}\tiny{$_{\rm \textcolor{red}{(TIP2021)}}$}\footnotesize{~\cite{9467810}}&20.74&0.5681&30.02&0.8990&35.33&0.9587&36.50&0.9638&38.48&0.9729&32.21&0.8725\\
\footnotesize{MADUN}\tiny{$_{\rm \textcolor{red}{(ACMMM2021)}}$}\footnotesize{~\cite{2021Memory}}&20.51&0.5647&29.91&0.8986&\underline{35.66}&\underline{0.9601}&\underline{36.94}&\underline{0.9676}&\underline{39.15}&\underline{0.9772}&\underline{32.43}&\underline{0.8736}\\
\midrule
DUN-CSNet&21.45&\underline{0.5893}&\textbf{30.87}&\textbf{0.9075}&\textbf{36.10}&\textbf{0.9617}&\textbf{37.39}&\textbf{0.9685}&\textbf{39.53}&\textbf{0.9777}&\textbf{33.07}&\textbf{0.8809}\\
\bottomrule

\end{tabular}
\label{tab:1}
\end{table*}

\begin{table*}[t]
\centering
\caption{Average PSNR and SSIM comparisons of different deep network-based CS algorithms using learned sampling matrix at diverse sampling rates on dataset Set14. Bold indicates the best result, and underline signifies the second-best result.}
\label{tab:3}
\vspace{-0.1in}
\small
\begin{tabular}{p{3.25cm}<{\centering} | p{0.75cm}<{\centering} p{0.8cm}<{\centering} | p{0.75cm}<{\centering}  p{0.8cm}<{\centering} | p{0.75cm}<{\centering} p{0.8cm}<{\centering} | p{0.75cm}<{\centering} p{0.8cm}<{\centering} | p{0.75cm}<{\centering} p{0.8cm}<{\centering} | p{0.75cm}<{\centering} p{0.8cm}<{\centering}}
\toprule
\multirow{2}*{Algorithms} & \multicolumn{2}{c}{Rate=0.01} & \multicolumn{2}{c}{Rate=0.10} & \multicolumn{2}{c}{Rate=0.25} & \multicolumn{2}{c}{Rate=0.30} & \multicolumn{2}{c}{Rate=0.40} & \multicolumn{2}{c}{Avg.}\\
\cline{2-13}
&PSNR&SSIM&PSNR&SSIM&PSNR&SSIM&PSNR&SSIM&PSNR&SSIM&PSNR&SSIM\\

\midrule

\footnotesize{CSNet}\tiny{$_{\rm \textcolor{red}{(ICME2017)}}$}\footnotesize{~\cite{8019428}}&22.79&0.5628&28.91&0.8119&32.86&0.9057&34.00&0.9276&35.84&0.9481&30.88&0.8312\\
\footnotesize{LapCSNet}\tiny{$_{\rm \textcolor{red}{(ICASSP2018)}}$}\footnotesize{~\cite{cui2018efficient}}&23.12&0.5762&29.07&0.8149&--\ --&--\ --&--\ --&--\ --&--\ --&--\ --&--\ --&--\ --\\
\footnotesize{SCSNet}\tiny{$_{\rm \textcolor{red}{(CVPR2019)}}$}\footnotesize{~\cite{shi2019scalable}}&22.87&0.5631&29.22&0.8181&33.24&0.9073&34.51&0.9311&36.54&0.9525&31.28&0.8344\\
\footnotesize{CSNet$^{+}$}\tiny{$_{\rm \textcolor{red}{(TIP2020)}}$}\footnotesize{~\cite{8765626}}&22.83&0.5630&29.13&0.8169&33.19&0.9064&34.34&0.9297&36.16&0.9502&31.13&0.8332\\
\footnotesize{NL-CSNet}\tiny{$_{\rm \textcolor{red}{(TMM2021)}}$}\footnotesize{~\cite{9635679}}&\textbf{23.61}&\textbf{0.5862}&30.16&\underline{0.8527}&33.84&0.9270&34.88&0.9405&36.86&0.9573&31.87&\underline{0.8527}\\
\hline
\footnotesize{BCS-Net}\tiny{$_{\rm \textcolor{red}{(TMM2020)}}$}\footnotesize{~\cite{9159912}}&22.68&0.5624&29.47&0.8105&34.02&0.9164&34.79&0.9312&36.68&0.9550&31.53&0.8351\\
\footnotesize{OPINENet$^{+}$}\tiny{$_{\rm \textcolor{red}{(JSTSP2020)}}$}\footnotesize{~\cite{9019857}}&22.30&0.5508&29.94&0.8415&34.31&0.9268&35.18&0.9369&37.51&0.9572&31.85&0.8426\\
\footnotesize{AMP-Net$^{+}$}\tiny{$_{\rm \textcolor{red}{(TIP2021)}}$}\footnotesize{~\cite{9298950}}&22.60&0.5723&29.87&0.8130&34.27&0.9218&35.23&0.9364&37.42&0.9561&31.88&0.8399\\
\footnotesize{COAST}\tiny{$_{\rm \textcolor{red}{(TIP2021)}}$}\footnotesize{~\cite{9467810}}&22.81&0.5764&\underline{30.26}&0.8507&34.72&0.9335&35.66&0.9404&37.86&0.9598&32.26&0.8522\\
\footnotesize{MADUN}\tiny{$_{\rm \textcolor{red}{(ACMMM2021)}}$}\footnotesize{~\cite{2021Memory}}&22.44&0.5675&30.17&0.8483&\underline{34.98}&\underline{0.9362}&\underline{36.03}&\underline{0.9473}&\underline{38.27}&\underline{0.9641}&\underline{32.38}&\underline{0.8527}\\
\midrule
DUN-CSNet&\underline{23.31}&\underline{0.5805}&\textbf{30.83}&\textbf{0.8598}&\textbf{35.43}&\textbf{0.9385}&\textbf{36.58}&\textbf{0.9500}&\textbf{38.63}&\textbf{0.9651}&\textbf{32.96}&\textbf{0.8588}\\
\bottomrule

\end{tabular}
\vspace{-0.16in}
\label{tab:3}
\end{table*}

In order to efficiently construct the references between the non-local patches ${\mathbf{x}}_{i}$ and ${\mathbf{X}}_{j}$, our main challenge is to seek an excellent inverse deformation $\mathcal{H}_{ij}^{-1}$ so that $\check{\mathbf{x}}_{ij}$ and ${\mathbf{x}}_{i}$  are similar enough to ensure to obtain more referenced knowledge from ${\mathbf{X}}_{j}$. In fact, the deformation from ${\mathbf{x}}_{i}$ to ${\mathbf{X}}_{j}$ means that the coordinate position undergoes a corresponding deviation. Reversibly, resampling the corresponding coordinate positions of ${\mathbf{X}}_{j}$ can actually realize an approximate inverse deformation. Therefore, the deformation $\mathcal{H}_{ij}^{-1}$ in our paper can be empirically designed via a resampling strategy~\cite{8237351}. For easier description, we first set the locations of patches ${\mathbf{x}}_{i}$ and ${\mathbf{X}}_{\hspace{-0.01in}j}$ as ${P}_{\hspace{-0.01in}i}$ and ${P}_{\hspace{-0.02in}j}$ in the current image or feature map. Because $\check{\mathbf{x}}_{ij}$ is the deformed version of patch ${\mathbf{X}}_{j}$, the location of $\check{\mathbf{x}}_{ij}$ is also set as ${P}_{\hspace{-0.02in}j}$. It is clear that the patch sizes of ${\mathbf{x}}_{i}$ and $\check{\mathbf{x}}_{ij}$ are the same, and we set them as $d$$\times$$d$. As above, the elements of $\check{\mathbf{x}}_{ij}$ (or ${\mathbf{x}}_{i}$) can be easily accessed through their locations and a regular grid $\mathcal{R}$~\cite{8237351}. The grid $\mathcal{R}$ is related to the patch size ($d$). For example, when $d$$=$$3$, $\mathcal{R}$$=$$\{(-1, -1),$ $(-1, 0),..., (0, 1), (1, 1)\}$. Based on the above definitions, we perform $\mathcal{H}_{ij}^{-1}$ in the following resampling manner:
\vspace{-0.05in}
\begin{equation}
\check{\mathbf{x}}_{ij}({P}_{\hspace{-0.02in}j}\hspace{-0.02in} +\hspace{-0.02in} { P}_{\hspace{-0.02in}n})\hspace{-0.02in} =\hspace{-0.02in} {\mathbf{X}}_{\hspace{-0.01in}j}({ P}_{\hspace{-0.02in}j}\hspace{-0.02in} +\hspace{-0.02in} {P}_{\hspace{-0.02in}n} +\hspace{-0.02in} \Delta {P}_{\hspace{-0.02in}ij})
\label{eq:3}
\vspace{-0.04in}
\end{equation}
where ${P}_{\hspace{-0.02in}n}\hspace{-0.04in}$ enumerates the elements of $\mathcal{R}$, and $\Delta {P}_{\hspace{-0.02in}ij}$ is the learnable offset to optimize the indexes of resampling. It is noted that the dimension of $\Delta {P}_{\hspace{-0.02in}ij}$ is 2$d^{2}$ for each pair of signal patches ${\mathbf{x}}_{i}$ and ${\mathbf{X}}_{j}$, which corresponds to $d^{2}$ 2D offsets in $X$ and $Y$ coordinate directions. Since the offset $\Delta { P}_{\hspace{-0.02in}ij}$ is typically fractional, we perform Eq.~\ref{eq:3} via a bilinear interpolation same as~\cite{8237351}, which ensures the back propagation of gradients during the training process. As above, the resampling strategy in our paper optimizes the resampling indexes by learning the offset $\Delta { P}_{\hspace{-0.02in}ij}$, realizing a flexible resampling operation.

Considering the function $f$ in Eq.~\ref{eq:2}, the following embedded Gaussian version is mainly analyzed to compute the similarities in an embedding space:
\vspace{-0.07in}
\begin{equation}
f({\mathbf{x}}_{i},\hspace{-0.03in} \mathcal{H}_{ij}^{-1}\hspace{-0.02in}({\mathbf{X}}_{j}))\hspace{-0.02in}=\hspace{-0.02in}e^{\theta({\mathbf{x}}_{i})^{\rm T}\cdot \varphi(\mathcal{H}_{ij}^{-1}\hspace{-0.02in}({\mathbf{X}}_{j}))}
\label{eq:5}
\vspace{-0.05in}
\end{equation}
where $\theta(\cdot)$ and $\varphi(\cdot)$ indicate two linear embedding functions to generate two embeddings. Specifically, these two embedding functions $\theta$ and $\varphi$ are defined as:

\vspace{-0.18in}
\begin{numcases}{}
\theta({\mathbf{x}}_{i})\hspace{-0.02in}=\hspace{-0.1in}\sum\limits_{{P}_{\hspace{-0.02in}n}\in \mathcal{R}}\hspace{-0.08in}{\mathbf{w}}_{\hspace{-0.01in}\theta}\hspace{-0.01in}({ P}_{\hspace{-0.02in}n})\hspace{-0.02in}\cdot\hspace{-0.02in} {\mathbf{x}}_{i}\hspace{-0.01in}({ P}_{\hspace{-0.02in}i}\hspace{-0.02in} +\hspace{-0.02in} {P}_{\hspace{-0.02in}n}) \label{eq13} \\
\varphi(\mathcal{H}_{ij}^{-1}\hspace{-0.02in}({\mathbf{X}}_{j}))\hspace{-0.02in}=\hspace{-0.1in}\sum\limits_{\hspace{-0.02in}{ P}_{\hspace{-0.02in}n}\in \mathcal{R}}\hspace{-0.08in}{\mathbf{w}}_{\hspace{-0.01in}\varphi}\hspace{-0.01in}({ P}_{\hspace{-0.02in}n})\hspace{-0.02in}\cdot\hspace{-0.02in} \check{\mathbf{x}}_{\hspace{-0.01in}ij}\hspace{-0.01in}({P}_{\hspace{-0.02in}j}\hspace{-0.02in} +\hspace{-0.02in} {P}_{\hspace{-0.02in}n}) \label{eq14}
\end{numcases}
\vskip -0.12in
\hspace{-0.2in} where $\check{\mathbf{x}}_{\hspace{-0.01in}ij}$=$\mathcal{H}_{ij}^{-1}({\mathbf{X}}_{j})$ is the deformed version of ${\mathbf{X}}_{j}$, which is computed through the Eq.~\ref{eq:3}. ${\mathbf{w}}_{\hspace{-0.01in}\theta}$ and ${\mathbf{w}}_{\hspace{-0.01in}\varphi}$ are the learnable weights in functions $\theta$ and $\varphi$ to generate the final responses. Analogously, the function $g$ is defined as:

\vspace{-0.1in}
\begin{equation}\label{eq15}
g(\mathcal{H}_{ij}^{-1}\hspace{-0.02in}({\mathbf{X}}_{j}))\hspace{-0.02in}=\hspace{-0.1in}\sum_{\hspace{-0.02in}{P}_{\hspace{-0.02in}n}\in \mathcal{R}}\hspace{-0.06in}{\mathbf{w}}_{\hspace{-0.01in}g}\hspace{-0.01in}({ P}_{\hspace{-0.02in}n})\hspace{-0.02in}\cdot\hspace{-0.02in} \check{\mathbf{x}}_{\hspace{-0.01in}ij}\hspace{-0.01in}({P}_{\hspace{-0.02in}j}\hspace{-0.02in} +\hspace{-0.02in} {P}_{\hspace{-0.02in}n})
\vspace{-0.06in}
\end{equation}
where ${\mathbf{w}}_{\hspace{-0.01in}g}$ indicates the learnable weight to produce the final embedding of the deformed patch $\check{\mathbf{x}}_{ij}$. Finally, we set $\mathcal{C}({\mathbf{x}})=\sum_{\hspace{-0.01in}\forall \hspace{-0.01in}j}\hspace{-0.02in}f({\mathbf{x}}_{i},\hspace{-0.03in} \mathcal{H}_{ij}^{-1}\hspace{-0.02in}({\mathbf{X}}_{j}))$ in Eq.~\ref{eq:2} to normalize the final response.

According to the existing deep non-local sub-module (NLM)~\cite{2018Non} (shown in Fig.~\ref{fig:3}(a)), we map the above data flow graph of our non-local operation into a specific network module (deformation-invariant non-local sub-module, DINLM), and Fig.~\ref{fig:3}(b) shows more details of its network structure. Specifically, given the input feature map, the deformation operator $\mathcal{H}^{-1}$ aims to deform the feature patches ($d$$\times$$d$) centered on each position. For more details of $\mathcal{H}^{-1}$ as shown in Fig.~\ref{fig:3}(c), the deformation offsets $\Delta {P}_{\hspace{-0.02in}ij}$ (with channel of $2d^{2}$) are first generated through a convolutional layer (kernel size is 3$\times$3), which are then utilized for feature deformation through the resampling strategy as shown in Eq.~\ref{eq:3}. In fact, Eq.~\ref{eq14} and Eq.~\ref{eq15} respectively perform the linear embeddings upon the deformed patches, which actually correspond to the operations of DCN (Deformable Convolutional Network) layer~\cite{8237351,huang2021learning}. For simplicity, we directly draw on the ideas of DCN to perform deformation. Specifically, corresponding to Eq.~\ref{eq14} and Eq.~\ref{eq15}, two DCN layers (no bias) with kernel size of $d$$\times$$d$ are utilized, and the deformable convolutional kernels of these two DCN layers are used to learn the embedding weights ${\rm \mathbf{w}}_{\hspace{-0.01in}\varphi}$ and ${\rm \mathbf{w}}_{\hspace{-0.01in}g}$. For the Eq.~\ref{eq13}, a convolutional layer with kernel size of $d\times d$ is utilized to learn weight ${\rm \mathbf{w}}_{\hspace{-0.01in}\theta}$. By introducing $\mathcal{H}^{-1}$, the proposed DINLM can exploit the non-local priors in certain learned deformation spaces and some special cases of learned deformation offsets are shown in Fig.~\ref{fig:3}(d). In addition, similar with the vanilla NLM, an affinity matrix is also generated in our DINLM, which is composed of all affinity coefficients computed from different feature patches.

For the entire network architecture of network DN-PMN as shown in Fig.~\ref{fig:1}, we first aggregate (concatenation operator is used) the two input entities ($\mathbf{r}^{(k)}$) and ($\mathbf{h}^{(k-1)}$). Then, the proposed proximal mapping network is appended. Specifically, for the internal network details of the proposed non-local network DN-PMN, two residual blocks are developed with a DINLM in the middle, and considering the residual block, several convolutional layers (with ReLU behind) are stacked by dense connection. After network DN-PMN, two output entities, i.e., $\mathbf{x}^{(k)}$ and $\mathbf{h}^{(k)}$, will be sent into the next phase for the cascaded image CS reconstruction.

In the inference process of our proposed deformation-invariant non-local module DINLM, a subsampling trick (similar with~\cite{2018Non}) is utilized to reduce the resource consumptions. As stated in~\cite{2018Non}, this subsampling trick does not alter the non-local behavior, but only makes the computation sparser. Besides, for function $f$ in Eq.~\ref{eq:5}, the other forms of metric functions, such as the vanilla Gaussian version and dot-product version, can also be used for affinity measuring.

\vspace{-0.05in}
\section{EXPERIMENTAL RESULTS}

\label{section:a4}

In this section, we first elaborate the loss function, and then demonstrate the experimental settings, implementation details and the experimental comparisons with the existing state-of-the-art CS methods. Finally, more ablation studies and discussions are provided in detail.
\vspace{-0.08in}
\subsection{Loss Function}
\vspace{-0.01in}

Given the input image $\mathbf{x}_{i}$ and the sampling matrix $\mathbf{\Phi}$, the measurement $\mathbf{y}_{i}$ can be acquired through the sampling process. To obtain the faithful reconstruction, the proposed DUN-CSNet takes $\mathbf{y}_{i}$ and $\mathbf{\Phi}$ as inputs and aims to narrow down the gap between the output and the target image $\mathbf{x}_{i}$. Besides, due to the cascaded multi-phase structure of the proposed network, multiple intermediate reconstructed results $\{\mathbf{x}_{i}^{(k)}\}$ are generated through the pipeline of the entire framework, where $k=\{1,2,...,K\}$ indicates the index of the phases.

\begin{table*}[t]
\centering
\caption{Average PSNR and SSIM comparisons of different deep unfolding CS networks using learned sampling matrix at diverse sampling rates on dataset Set5. Bold indicates the best result, and underline signifies the second-best result.}
\label{tab:25}
\vspace{-0.1in}
\small
\begin{tabular}{p{3.25cm}<{\centering} | p{0.75cm}<{\centering} p{0.8cm}<{\centering} | p{0.75cm}<{\centering}  p{0.8cm}<{\centering} | p{0.75cm}<{\centering} p{0.8cm}<{\centering} | p{0.75cm}<{\centering} p{0.8cm}<{\centering} | p{0.75cm}<{\centering} p{0.8cm}<{\centering} | p{0.75cm}<{\centering} p{0.8cm}<{\centering}}
\toprule
\multirow{2}*{Algorithms} & \multicolumn{2}{c}{Rate=0.01} & \multicolumn{2}{c}{Rate=0.10} & \multicolumn{2}{c}{Rate=0.25} & \multicolumn{2}{c}{Rate=0.30} & \multicolumn{2}{c}{Rate=0.40} & \multicolumn{2}{c}{Avg.}\\
\cline{2-13}
&PSNR&SSIM&PSNR&SSIM&PSNR&SSIM&PSNR&SSIM&PSNR&SSIM&PSNR&SSIM\\

\midrule

\footnotesize{BCS-Net}\tiny{$_{\rm \textcolor{red}{(TMM2020)}}$}\footnotesize{~\cite{9159912}}&22.98&0.6103&32.71&0.9030&37.90&0.9576&38.64&0.9694&39.88&0.9785&34.42&0.8838\\
\footnotesize{OPINENet$^{+}$}\tiny{$_{\rm \textcolor{red}{(JSTSP2020)}}$}\footnotesize{~\cite{9019857}}&21.88&0.5162&27.81&0.8040&31.50&0.9062&32.79&0.9278&34.73&0.9521&34.95&0.8911\\
\footnotesize{AMP-Net$^{+}$}\tiny{$_{\rm \textcolor{red}{(TIP2021)}}$}\footnotesize{~\cite{9298950}}&22.30&0.5391&27.86&0.7928&31.75&0.9050&32.84&0.9242&34.86&0.9509&34.87&0.8938\\
\footnotesize{COAST}\tiny{$_{\rm \textcolor{red}{(TIP2021)}}$}\footnotesize{~\cite{9467810}}&\underline{23.31}&\underline{0.6514}&\underline{33.90}&0.9266&38.21&0.9648&39.23&0.9706&41.36&0.9780&35.20&0.8983\\
\footnotesize{MADUN}\tiny{$_{\rm \textcolor{red}{(ACMMM2021)}}$}\footnotesize{~\cite{2021Memory}}&23.12&0.6503&33.86&\underline{0.9267}&\underline{38.44}&\underline{0.9660}&\underline{39.57}&\underline{0.9723}&\underline{41.72}&\underline{0.9808}&\underline{35.34}&\underline{0.8992}\\
\midrule

DUN-CSNet&\textbf{24.35}&\textbf{0.6668}&\textbf{34.50}&\textbf{0.9360}&\textbf{39.00}&\textbf{0.9691}&\textbf{40.08}&\textbf{0.9743}&\textbf{42.17}&\textbf{0.9817}&\textbf{36.02}&\textbf{0.9056}\\
\bottomrule

\end{tabular}
\vspace{-0.08in}
\label{tab:25}
\end{table*}

\begin{table*}[t]
\centering
\caption{Average PSNR and SSIM comparisons of different deep unfolding CS networks using learned sampling matrix at diverse sampling rates on dataset BSD68. Bold indicates the best result, and underline signifies the second-best result.}
\label{tab:2}
\vspace{-0.1in}
\small
\begin{tabular}{p{3.25cm}<{\centering} | p{0.75cm}<{\centering} p{0.8cm}<{\centering} | p{0.75cm}<{\centering}  p{0.8cm}<{\centering} | p{0.75cm}<{\centering} p{0.8cm}<{\centering} | p{0.75cm}<{\centering} p{0.8cm}<{\centering} | p{0.75cm}<{\centering} p{0.8cm}<{\centering} | p{0.75cm}<{\centering} p{0.8cm}<{\centering}}
\toprule
\multirow{2}*{Algorithms} & \multicolumn{2}{c}{Rate=0.01} & \multicolumn{2}{c}{Rate=0.10} & \multicolumn{2}{c}{Rate=0.25} & \multicolumn{2}{c}{Rate=0.30} & \multicolumn{2}{c}{Rate=0.40} & \multicolumn{2}{c}{Avg.}\\
\cline{2-13}
&PSNR&SSIM&PSNR&SSIM&PSNR&SSIM&PSNR&SSIM&PSNR&SSIM&PSNR&SSIM\\

\midrule

\footnotesize{BCS-Net}\tiny{${\rm \textcolor{red}{(TMM2020)}}$}\footnotesize{\cite{9159912}}&22.16&0.5287&27.78&0.7864&31.14&0.9006&32.15&0.9167&33.90&0.9473&29.43&0.8159\\
\footnotesize{OPINENet$^{+}$}\tiny{${\rm \textcolor{red}{(JSTSP2020)}}$}\footnotesize{\cite{9019857}}&21.88&0.5162&27.81&0.8040&31.50&0.9062&32.78&0.9278&34.73&0.9521&29.74&0.8213\\
\footnotesize{AMP-Net$^{+}$}\tiny{${\rm \textcolor{red}{(TIP2021)}}$}\footnotesize{\cite{9298950}}&21.94&0.5253&\underline{27.86}&0.7928&31.75&0.9050&32.84&0.9242&34.86&0.9509&29.85&0.8196\\
\footnotesize{COAST}\tiny{${\rm \textcolor{red}{(TIP2021)}}$}\footnotesize{\cite{9467810}}&\underline{22.30}&\underline{0.5391}&27.80&0.8091&31.81&0.9128&32.78&0.9331&34.90&0.9565&\underline{29.92}&\underline{0.8301}\\
\footnotesize{MADUN}\tiny{${\rm \textcolor{red}{(ACMMM2021)}}$}\footnotesize{\cite{2021Memory}}&21.65&0.5249&27.74&\underline{0.8108}&\underline{31.90}&\underline{0.9165}&\underline{32.96}&\underline{0.9353}&\underline{35.02}&\underline{0.9584}&29.86&0.8293\\
\midrule

DUN-CSNet&\textbf{22.71}&\textbf{0.5400}&\textbf{28.39}&\textbf{0.8231}&\textbf{32.31}&\textbf{0.9207}&\textbf{33.40}&\textbf{0.9371}&\textbf{35.46}&\textbf{0.9595}&\textbf{30.45}&\textbf{0.8361}\\
\bottomrule

\end{tabular}
\vspace{-0.16in}
\label{tab:2}
\end{table*}

In fact, each phase of algorithm PGD can obtain an intermediate reconstructed result, and then the result is transferred to the next phase for recursive iteration. Inspired by this inferential conception, in our proposed DUN-CSNet, the outputs $\{\mathbf{x}_{i}^{(k)}\}$ of different phases are all constrained. More specifically, we directly use the L2 norm to restrain the distance between the output $\mathbf{x}_{i}^{(k)}$ and the ground truth image $\mathbf{x}_{i}$, i.e.,
\vspace{-0.06in}
\begin{equation}
\mathcal{L}({\mathbf{\Theta}}) =  \frac{1}{KN_{a}} \sum_{i=1}^{N_{a}}\sum_{k=1}^{K}\| \mathbf{x}_{i}^{(k)} - \mathbf{x}_{i} \|_{2}^{2}
\vspace{-0.03in}
\end{equation}
where $\mathbf{\Theta}$ denotes the trainable parameter set of our proposed DUN-CSNet, including the learnable parameters of the networks CA-GDN and DN-PMN in all cascaded phases. $N_{a}$ and $K$ respectively represent the number of training images and the phase number of the proposed CS framework. It is worth noting that similar with the existing representative CS networks, the sampling matrix $\mathbf{\Phi}$ in our framework can also be jointly optimized with the reconstruction process.

\begin{figure}[b]
\begin{center}
\vspace{-0.17in}
\includegraphics[width=1.7in]{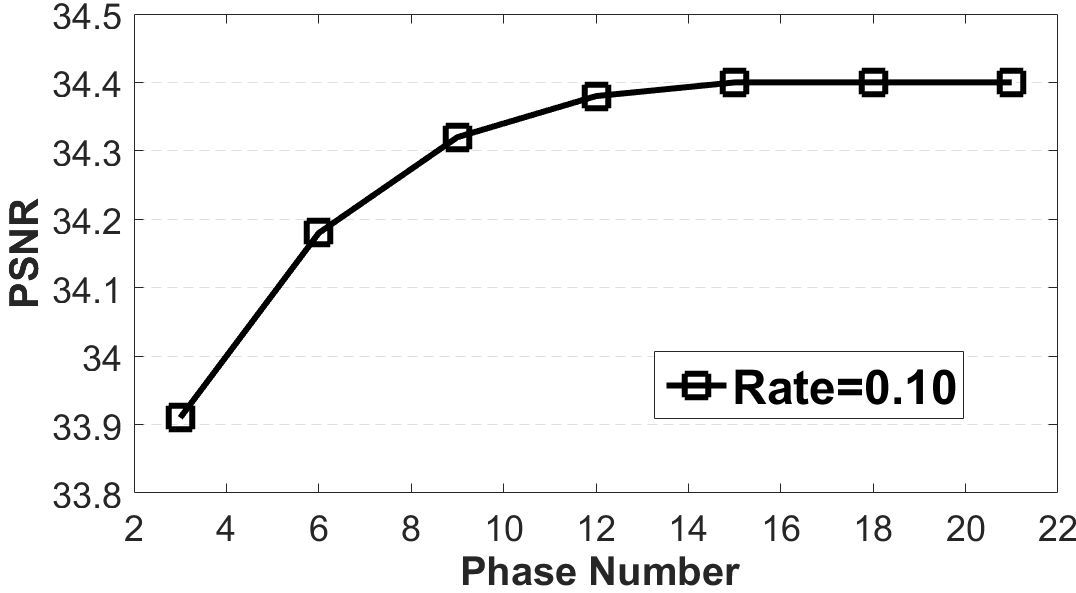}
\includegraphics[width=1.7in]{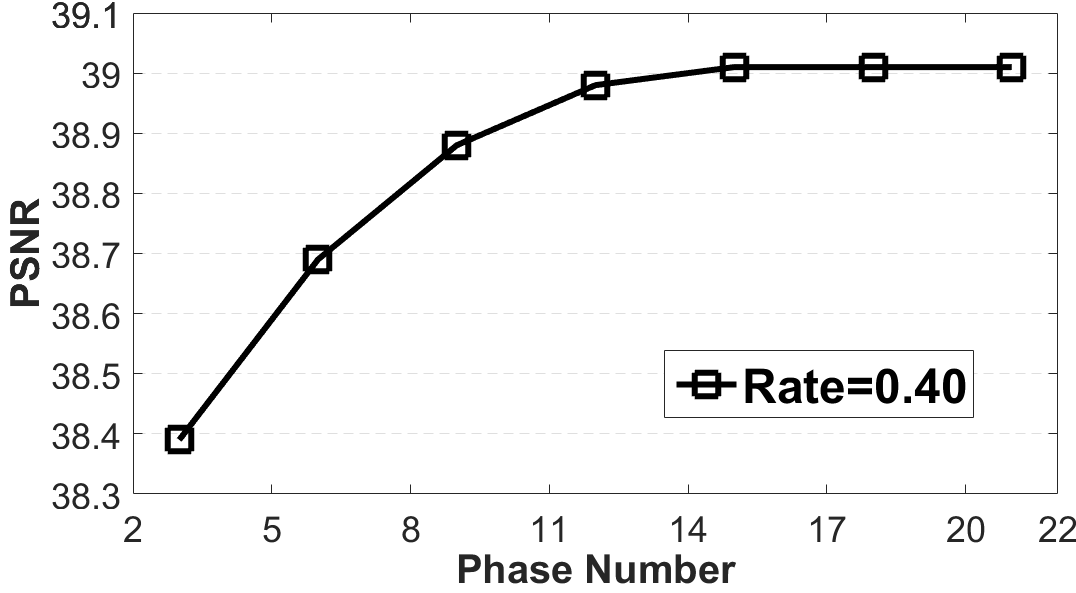}
\end{center}
\vskip -0.1in
   \caption{The relationship between the phase number and the reconstructed quality (PSNR) of the proposed DUN-CSNet on the dataset Set5 under two sampling rates 0.10 and 0.40.}
\label{fig:9}
\end{figure}

\vspace{-0.15in}
\subsection{Implementation and Training Details}
\vspace{-0.01in}

\begin{figure}[h]
\begin{center}
\vspace{-0.1in}
\includegraphics[width=1.7in]{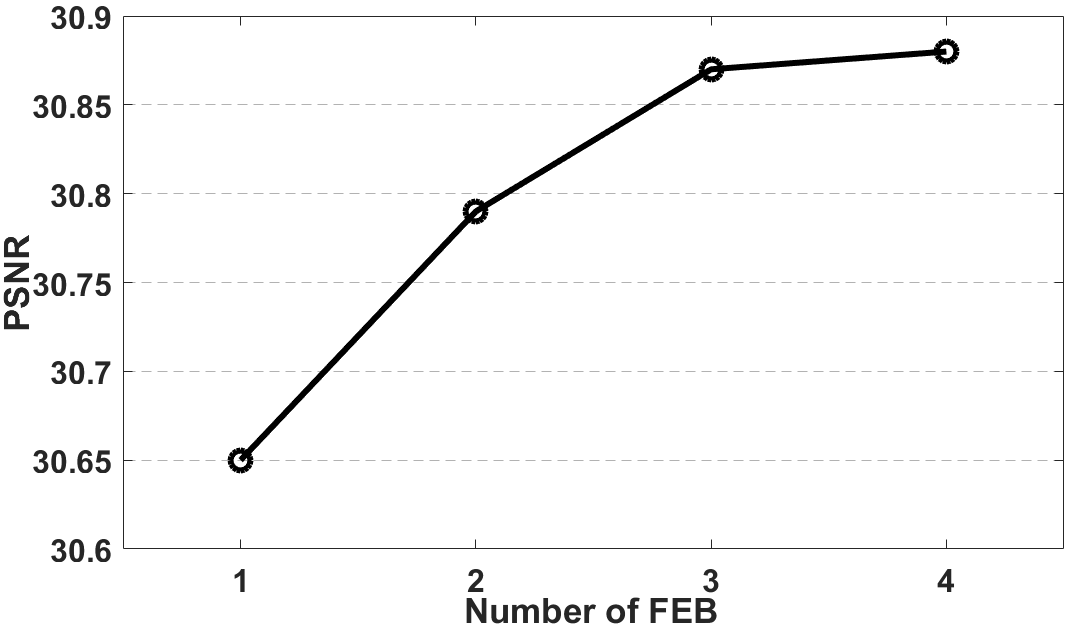}
\includegraphics[width=1.7in]{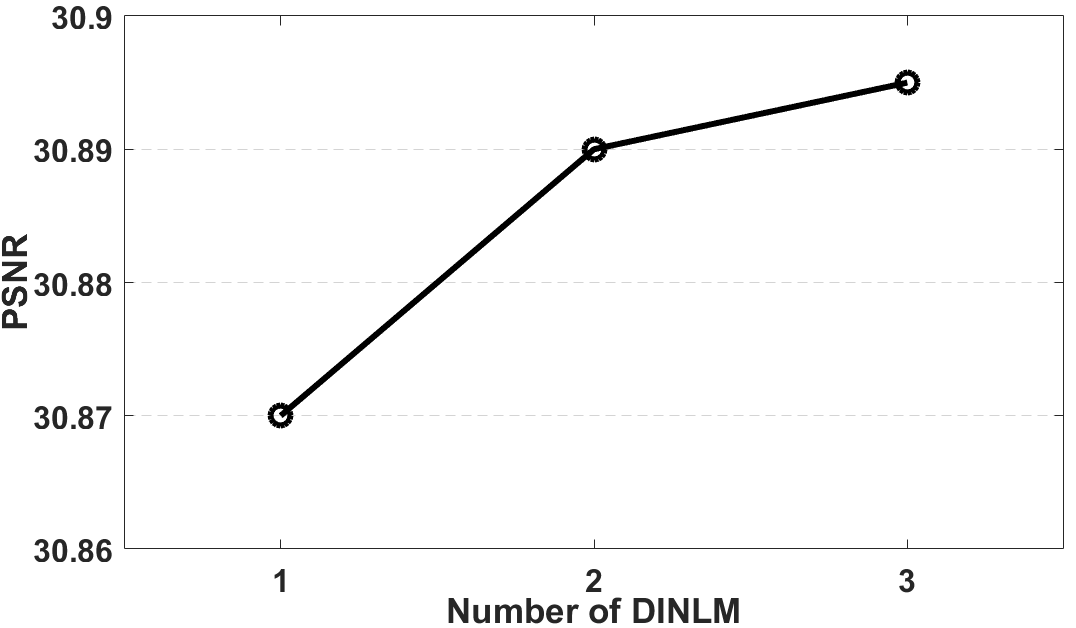}
\end{center}
\vskip -0.1in
   \caption{Left figure shows the relationship between the FEB number and the reconstructed quality. Right figure shows the relationship between the DINLM number and the reconstruction performance. The results are based on the dataset Set11 at sampling rate 0.10.}
\vspace{-0.08in}
\label{fig:94}
\end{figure}

In the proposed CS framework DUN-CSNet, we set block size $B=33$ (same as the recent CS literatures~\cite{8765626,8578294}). Considering the phase number $K$, Fig.~\ref{fig:9} shows the relationship between the phase number and the reconstructed image quality, from which we can get that with the phase number increases, the reconstruction performance tends to converge, and finally we set $K=15$ in our model. For more configuration details of the proposed CS network, we set the channel number of the intermediate feature maps as 32, i.e., ${\rm c}=32$. For the number of FEB in sub-network SSG-Net, left figure of Fig.~\ref{fig:94} shows the relationship between the number of FEB and reconstructed quality, from which we can observe that with the increase of the number of FEB, the performance of the model gradually improves and tends to be stable. As above, in our proposed DUN-CSNet, the number of FEB in SSG-Net is set as 3. Similar to the analysis of FEB, right figure of Fig.~\ref{fig:94} shows the analysis results about the number of DINLM in DN-PMN, and finally we set the number of DINLM as 1. In network DN-PMN, the number of the convolutional layer for each residual block is set as 3, and the kernels of size $3\times 3$ are utilized in these convolutional layers. For more details about DINLM, we set the patch size as $3\times 3$ (i.e., $d=3$). In the training process, we initialize all the convolutional filters using the same method as~\cite{9635679} and pad zeros around the boundaries to keep the size of feature maps the same as the input.

\begin{figure*}[t]

\begin{minipage}[t]{0.135\textwidth}
\centering
\includegraphics[width=0.95in]{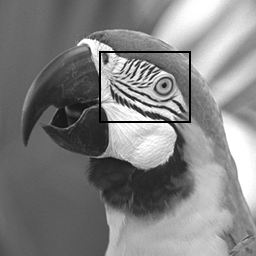}
\begin{scriptsize}
\centering
\vskip -0.52 cm \begin{tiny}Parrots$\backslash$PSNR$\backslash$SSIM\end{tiny}
\end{scriptsize}
\end{minipage}
\hspace{-0.012in}
\begin{minipage}[t]{0.135\textwidth}
\centering
\includegraphics[width=0.95in]{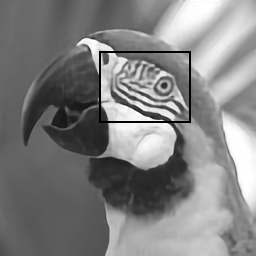}
\begin{scriptsize}
\centering
\vskip -0.52 cm \begin{tiny}CSNet$^{+}$~\cite{8765626}$\backslash$28.10$\backslash$0.8921\end{tiny}
\end{scriptsize}
\end{minipage}
\hspace{-0.012in}
\begin{minipage}[t]{0.135\textwidth}
\centering
\includegraphics[width=0.95in]{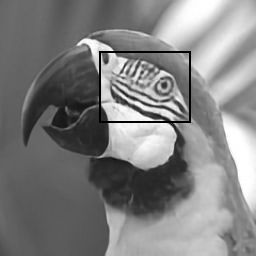}
\begin{scriptsize}
\centering
\vskip -0.52 cm \begin{tiny}NL-CSNet~\cite{9635679}$\backslash$29.20$\backslash$0.9203\end{tiny}
\end{scriptsize}
\end{minipage}
\hspace{-0.012in}
\begin{minipage}[t]{0.135\textwidth}
\centering
\includegraphics[width=0.95in]{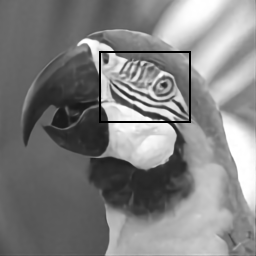}
\begin{scriptsize}
\centering
\vskip -0.52 cm \begin{tiny}OPINE-Net~\cite{9019857}$\backslash$29.34$\backslash$0.9155\end{tiny}
\end{scriptsize}
\end{minipage}
\hspace{-0.012in}
\begin{minipage}[t]{0.135\textwidth}
\centering
\includegraphics[width=0.95in]{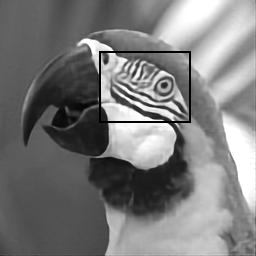}
\begin{scriptsize}
\centering
\vskip -0.52 cm \begin{tiny}AMP-Net~\cite{9298950}$\backslash$29.20$\backslash$0.9055\end{tiny}
\end{scriptsize}
\end{minipage}
\hspace{-0.012in}
\begin{minipage}[t]{0.135\textwidth}
\centering
\includegraphics[width=0.95in]{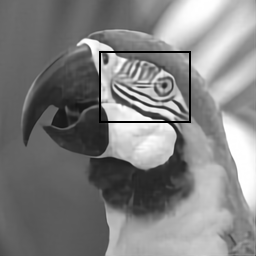}
\begin{scriptsize}
\centering
\vskip -0.52 cm \begin{tiny}MADUN~\cite{2021Memory}$\backslash$29.30$\backslash$0.9198\end{tiny}
\end{scriptsize}
\end{minipage}
\hspace{-0.012in}
\begin{minipage}[t]{0.135\textwidth}
\centering
\includegraphics[width=0.95in]{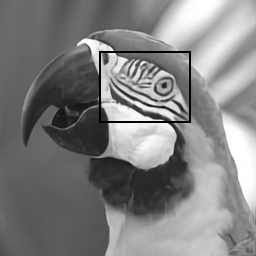}
\begin{scriptsize}
\centering
\vskip -0.52 cm \begin{tiny}DUN-CSNet$\backslash$\textbf{30.07}$\backslash$\textbf{0.9271}\end{tiny}
\end{scriptsize}
\end{minipage}

\begin{minipage}[t]{0.135\textwidth}
\centering
\includegraphics[width=0.95in]{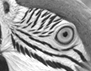}
\begin{scriptsize}
\end{scriptsize}
\end{minipage}
\hspace{-0.012in}
\begin{minipage}[t]{0.135\textwidth}
\centering
\includegraphics[width=0.95in]{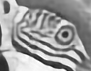}
\begin{scriptsize}
\end{scriptsize}
\end{minipage}
\hspace{-0.012in}
\begin{minipage}[t]{0.135\textwidth}
\centering
\includegraphics[width=0.95in]{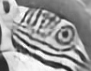}
\begin{scriptsize}
\end{scriptsize}
\end{minipage}
\hspace{-0.012in}
\begin{minipage}[t]{0.135\textwidth}
\centering
\includegraphics[width=0.95in]{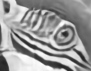}
\begin{scriptsize}
\end{scriptsize}
\end{minipage}
\hspace{-0.012in}
\begin{minipage}[t]{0.135\textwidth}
\centering
\includegraphics[width=0.95in]{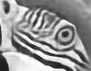}
\begin{scriptsize}
\end{scriptsize}
\end{minipage}
\hspace{-0.012in}
\begin{minipage}[t]{0.135\textwidth}
\centering
\includegraphics[width=0.95in]{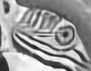}
\begin{scriptsize}
\end{scriptsize}
\end{minipage}
\hspace{-0.012in}
\begin{minipage}[t]{0.135\textwidth}
\centering
\includegraphics[width=0.95in]{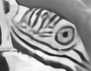}
\begin{scriptsize}
\end{scriptsize}
\end{minipage}

\vspace{-0.19in}
\caption{Visual quality comparisons of different deep network-based CS methods on the image $\emph{Parrots}$ from dataset Set11 under the sampling rate 0.10.}
\vspace{-0.02in}
\label{fig:444}
\vspace{-0.06in}
\end{figure*}

\begin{figure*}[t]

\begin{minipage}[t]{0.135\textwidth}
\centering
\includegraphics[width=0.95in]{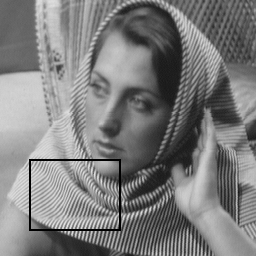}
\begin{scriptsize}
\centering
\vskip -0.52 cm \begin{tiny}GT$\backslash$PSNR$\backslash$SSIM\end{tiny}
\end{scriptsize}
\end{minipage}
\hspace{-0.012in}
\begin{minipage}[t]{0.135\textwidth}
\centering
\includegraphics[width=0.95in]{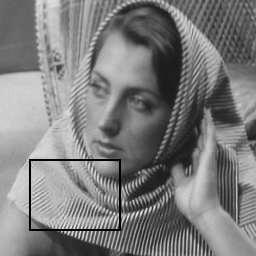}
\begin{scriptsize}
\centering
\vskip -0.52 cm \begin{tiny}CSNet$^{+}$~\cite{8765626}$\backslash$31.22$\backslash$0.9256\end{tiny}
\end{scriptsize}
\end{minipage}
\hspace{-0.012in}
\begin{minipage}[t]{0.135\textwidth}
\centering
\includegraphics[width=0.95in]{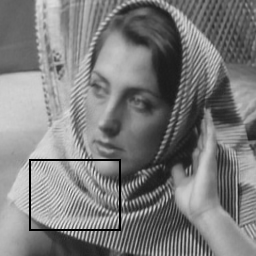}
\begin{scriptsize}
\centering
\vskip -0.52 cm \begin{tiny}NL-CSNet~\cite{9635679}$\backslash$34.85$\backslash$0.9685\end{tiny}
\end{scriptsize}
\end{minipage}
\hspace{-0.012in}
\begin{minipage}[t]{0.135\textwidth}
\centering
\includegraphics[width=0.95in]{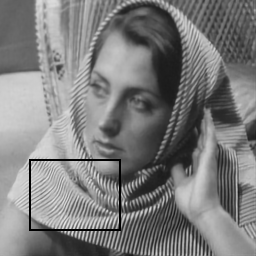}
\begin{scriptsize}
\centering
\vskip -0.52 cm \begin{tiny}OPINE-Net~\cite{9019857}$\backslash$34.55$\backslash$0.9666\end{tiny}
\end{scriptsize}
\end{minipage}
\hspace{-0.012in}
\begin{minipage}[t]{0.135\textwidth}
\centering
\includegraphics[width=0.95in]{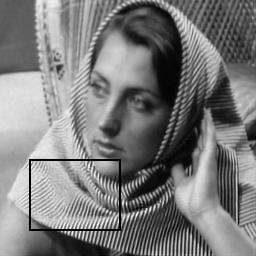}
\begin{scriptsize}
\centering
\vskip -0.52 cm \begin{tiny}AMP-Net~\cite{9298950}$\backslash$33.53$\backslash$0.9522\end{tiny}
\end{scriptsize}
\end{minipage}
\hspace{-0.012in}
\begin{minipage}[t]{0.135\textwidth}
\centering
\includegraphics[width=0.95in]{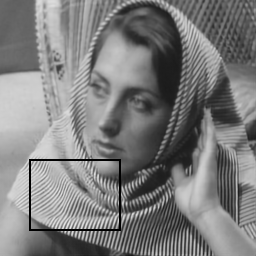}
\begin{scriptsize}
\centering
\vskip -0.52 cm \begin{tiny}MADUN~\cite{2021Memory}$\backslash$35.65$\backslash$0.9726\end{tiny}
\end{scriptsize}
\end{minipage}
\hspace{-0.012in}
\begin{minipage}[t]{0.135\textwidth}
\centering
\includegraphics[width=0.95in]{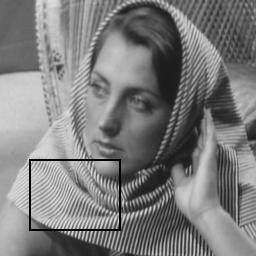}
\begin{scriptsize}
\centering
\vskip -0.52 cm \begin{tiny}DUN-CSNet$\backslash$\textbf{36.59}$\backslash$\textbf{0.9779}\end{tiny}
\end{scriptsize}
\end{minipage}

\begin{minipage}[t]{0.135\textwidth}
\centering
\includegraphics[width=0.95in]{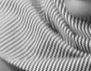}
\begin{scriptsize}
\end{scriptsize}
\end{minipage}
\hspace{-0.012in}
\begin{minipage}[t]{0.135\textwidth}
\centering
\includegraphics[width=0.95in]{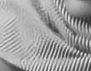}
\begin{scriptsize}
\end{scriptsize}
\end{minipage}
\hspace{-0.012in}
\begin{minipage}[t]{0.135\textwidth}
\centering
\includegraphics[width=0.95in]{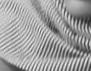}
\begin{scriptsize}
\end{scriptsize}
\end{minipage}
\hspace{-0.012in}
\begin{minipage}[t]{0.135\textwidth}
\centering
\includegraphics[width=0.95in]{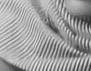}
\begin{scriptsize}
\end{scriptsize}
\end{minipage}
\hspace{-0.012in}
\begin{minipage}[t]{0.135\textwidth}
\centering
\includegraphics[width=0.95in]{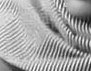}
\begin{scriptsize}
\end{scriptsize}
\end{minipage}
\hspace{-0.012in}
\begin{minipage}[t]{0.135\textwidth}
\centering
\includegraphics[width=0.95in]{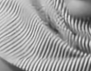}
\begin{scriptsize}
\end{scriptsize}
\end{minipage}
\hspace{-0.012in}
\begin{minipage}[t]{0.135\textwidth}
\centering
\includegraphics[width=0.95in]{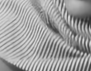}
\begin{scriptsize}
\end{scriptsize}
\end{minipage}
\vspace{-0.16in}
\caption{Visual quality comparisons of different deep network-based CS methods on the image $\emph{Barbara}$ from dataset Set11 under the sampling rate 0.30.}
\vspace{-0.02in}
\label{fig:555}
\vspace{-0.15in}
\end{figure*}

For training dataset, we use the training set (400 images) of dataset BSD500 as our training data, which has been widely used in many existing deep network-based CS methods~\cite{8765626,9019857}. Specifically, to expand the diversity of training images, we augment the training data in the following two ways: 1) Rotate the training images by 90$^{\circ}$, 180$^{\circ}$ and 270$^{\circ}$ randomly. 2) Flip the training images horizontally and vertically randomly. In the training process, we first convert the RGB image into the grayscale format and then randomly crop the size of image patches to $99\times 99$. For more training details, we use the PyTorch toolbox and train our model using the Adam optimizer ($\beta_{1}$=0.9 and $\beta_{2}$=0.999) on a NVIDIA GTX 3090 GPU. Besides, during the training procedure, we set the batch size as 16, and the learning rate is initialized to 1e-4 in the beginning and is halved every 30 epochs. We train our model for 200 epochs totally and 2000 iterations are performed for each epoch. Therefore 200$\times$2000 iterations are completed in the whole training process.

\vspace{-0.1in}
\subsection{Comparisons with State-of-the-art Methods}
\vspace{-0.03in}
In recent deep network-based CS methods, the sampling matrix is usually optimized jointly with the reconstruction process. Compared with the Gaussian random sampling matrix, the learned sampling matrix is usually able to achieve better reconstructed quality. To evaluate the performance of the proposed CS framework, we mainly compare our proposed DUN-CSNet with the recent deep network-based CS schemes that use learned sampling matrix. Specifically, depending on the related works analyzed in Section~\ref{section:a2}, the compared CS methods can be roughly grouped into the following two categories: deep black box CS networks (DBNs) and deep unfolding CS networks (DUNs). For DBNs, five CS algorithms are considered, including CSNet~\cite{8019428}, LapCSNet~\cite{cui2018efficient}, SCSNet~\cite{shi2019scalable}, CSNet$^{+}$~\cite{8765626} and NL-CSNet~\cite{9635679}. For DUNs, five representative CS methods, i.e., BCS-Net~\cite{9159912}, OPINE-Net$^{+}$~\cite{9019857}, AMP-Net~\cite{9298950} COAST~\cite{9467810} and MADUN~\cite{2021Memory} participate in the comparison in our experiments.

\begin{figure*}[t]
\begin{center}

\hspace{1.384in}
\includegraphics[width=1.385in]{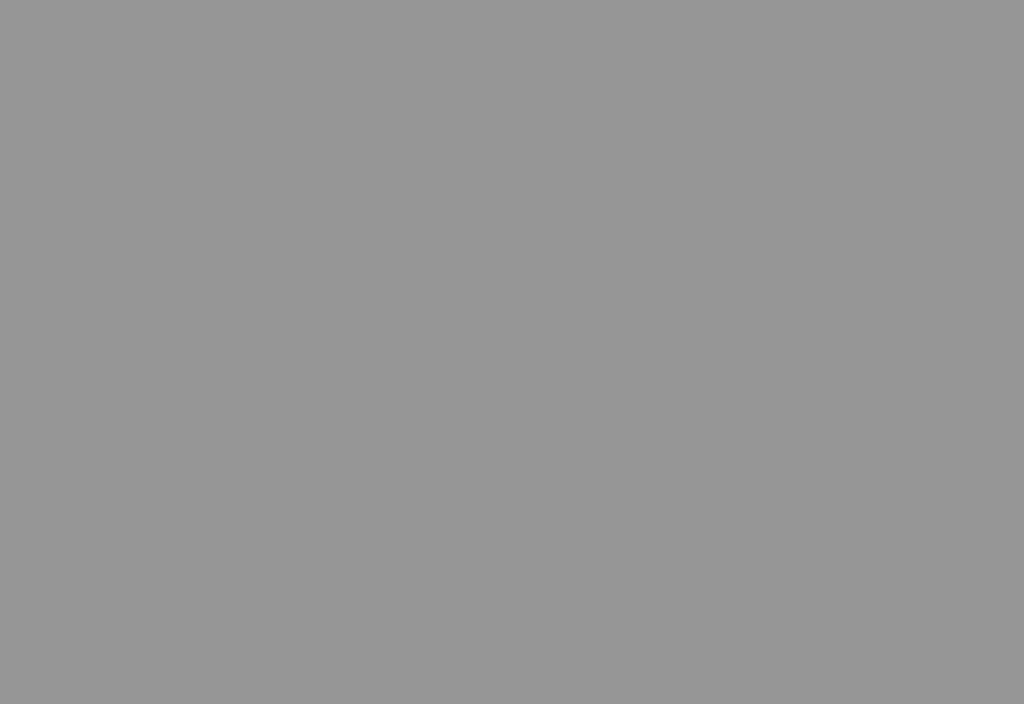} 
\includegraphics[width=1.385in]{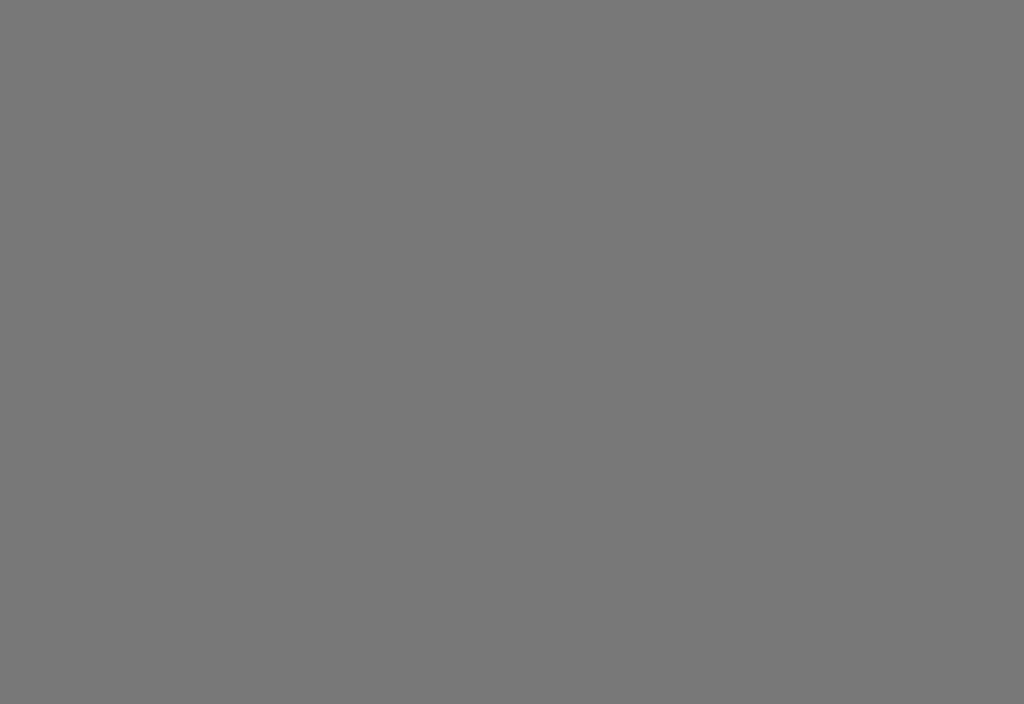} 
\includegraphics[width=1.385in]{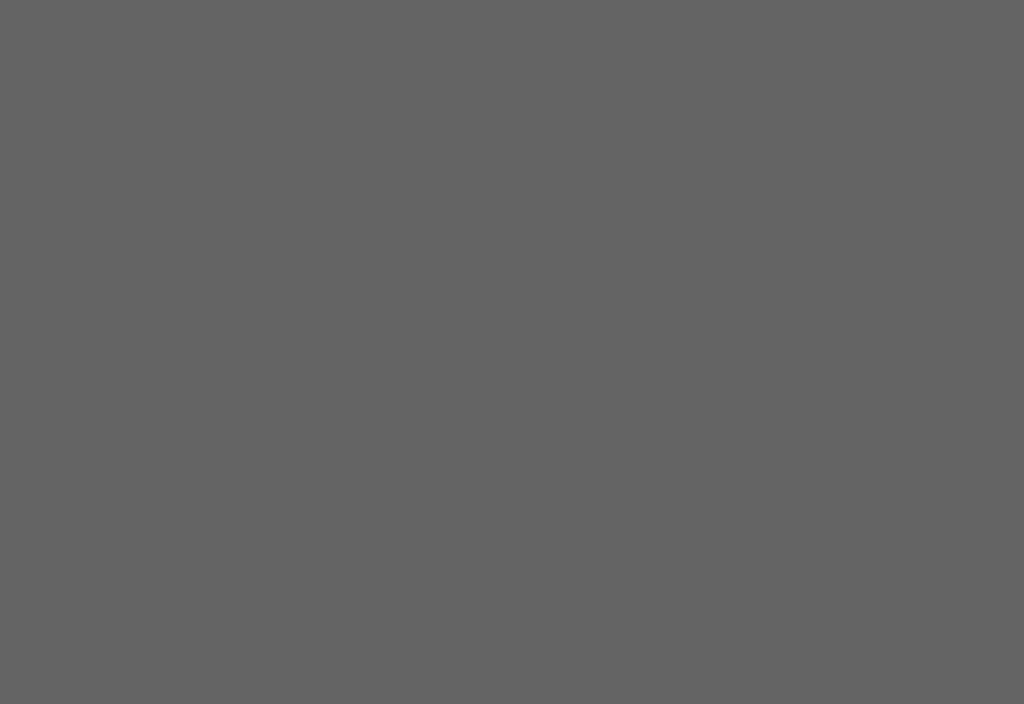} 
\includegraphics[width=1.385in]{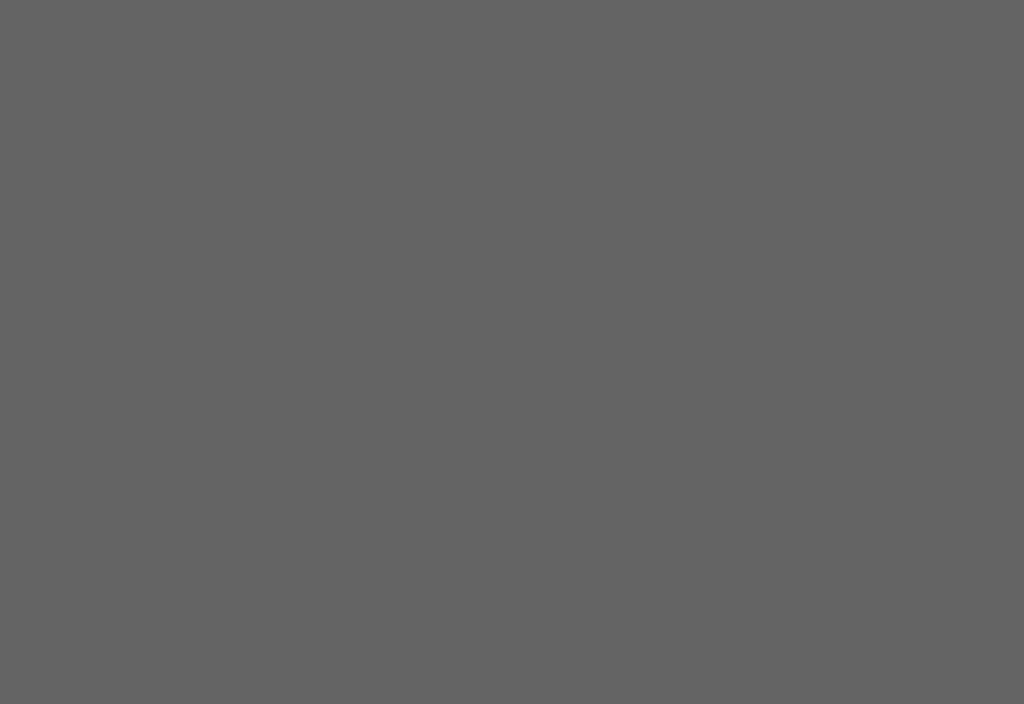} 
\\
\vspace{0.03in}
\hspace{1.384in}
\includegraphics[width=1.385in]{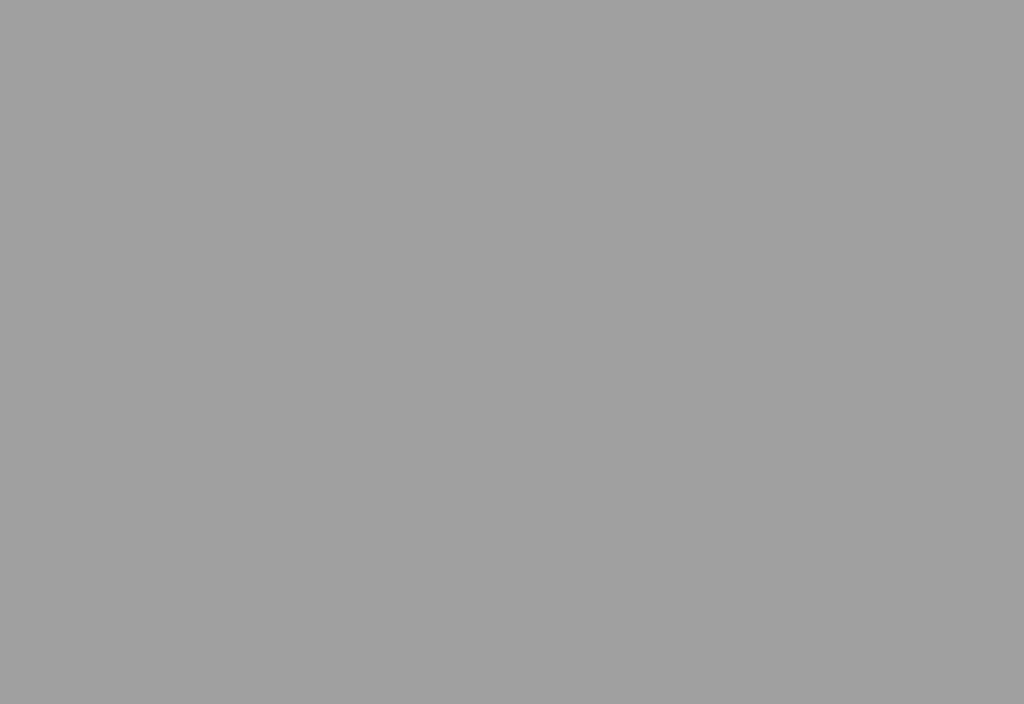} 
\includegraphics[width=1.385in]{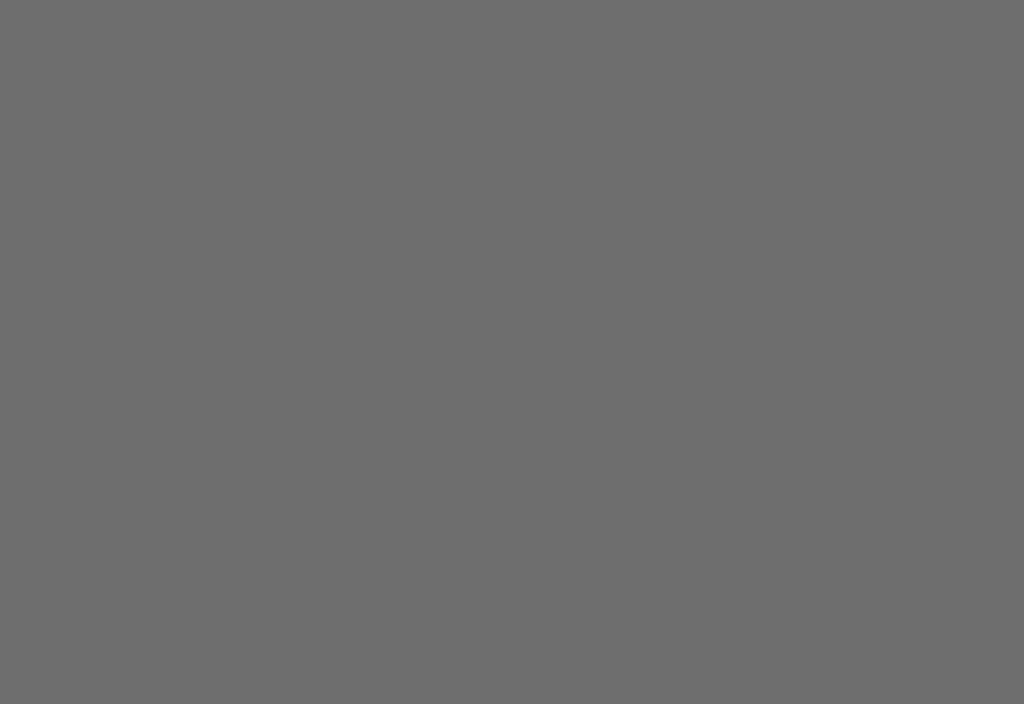} 
\includegraphics[width=1.385in]{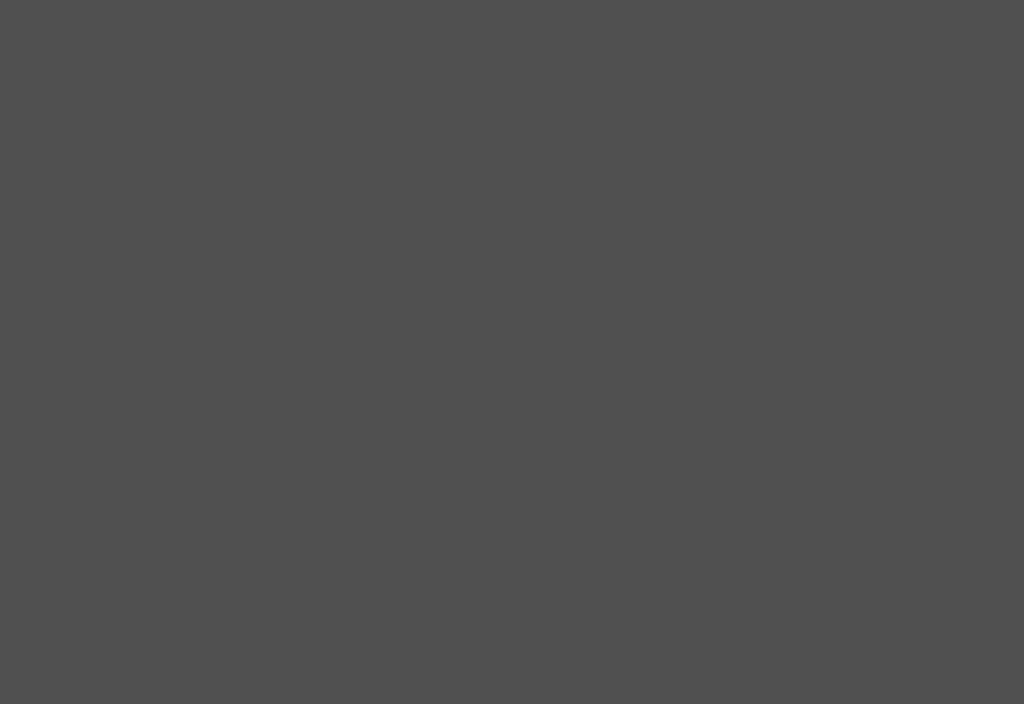} 
\includegraphics[width=1.385in]{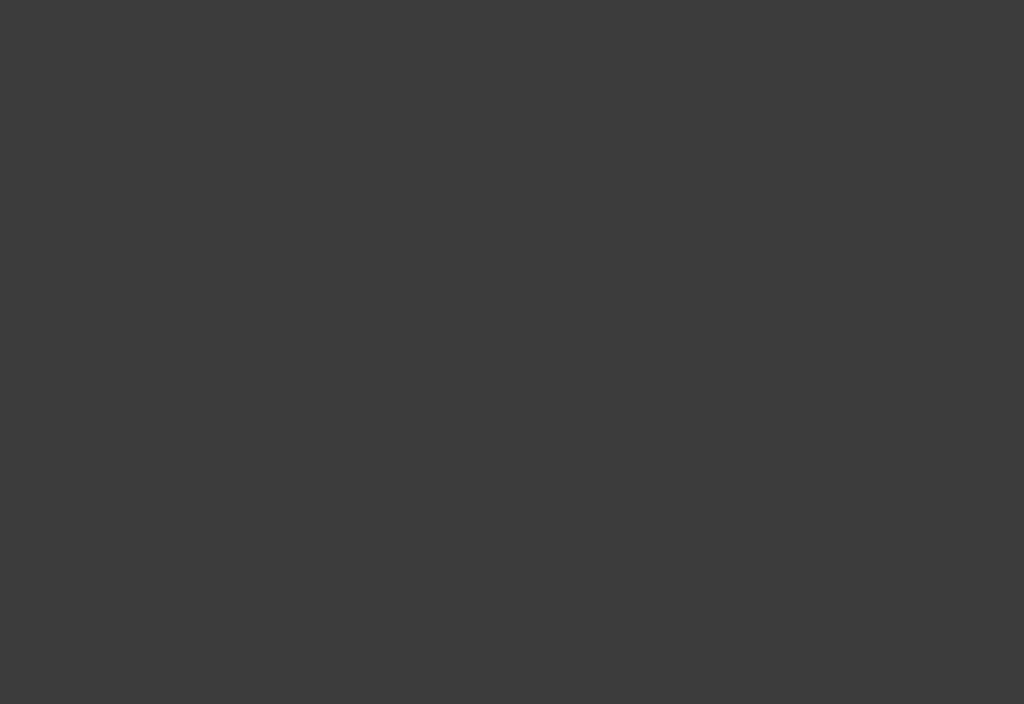}
\\
\vspace{0.03in}
\hspace{1.384in}
\includegraphics[width=1.385in]{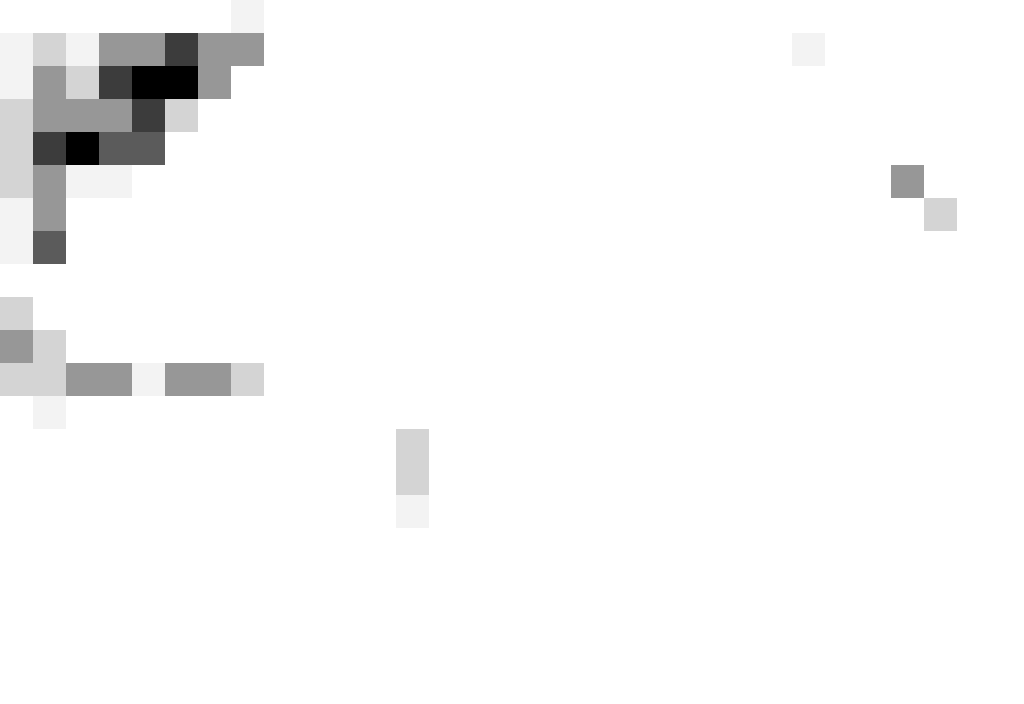} 
\includegraphics[width=1.385in]{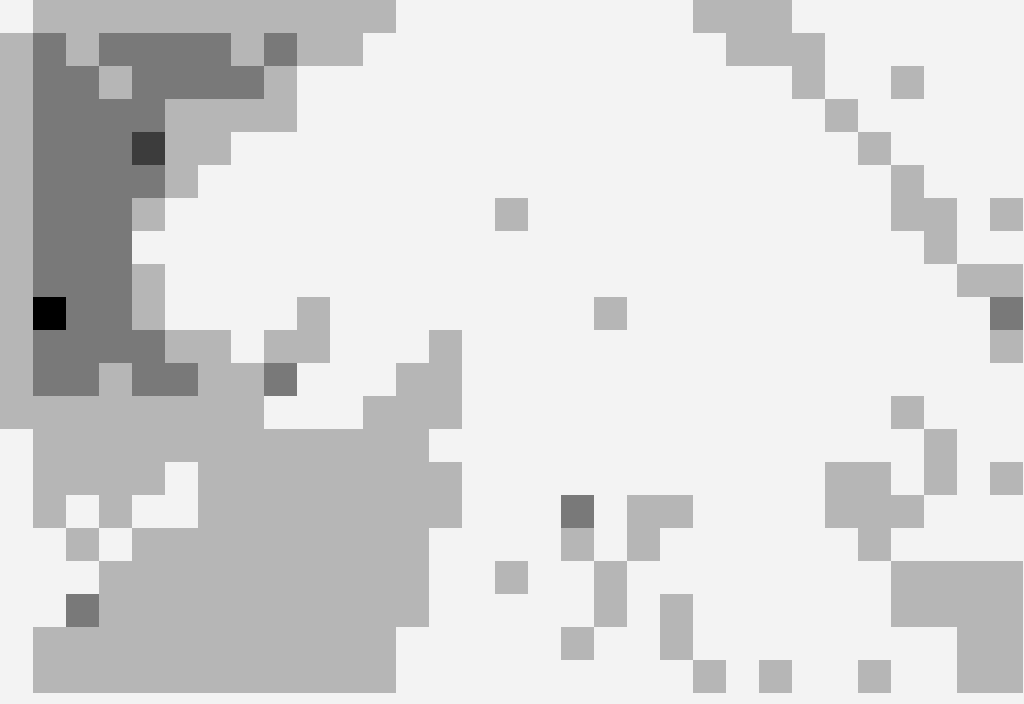} 
\includegraphics[width=1.385in]{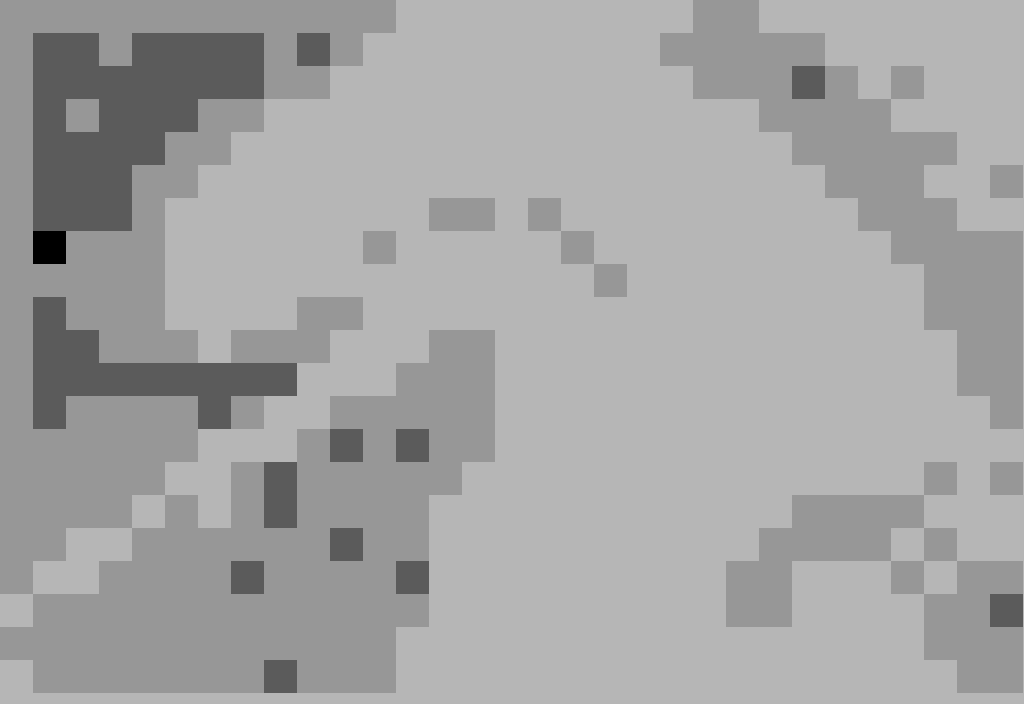} 
\includegraphics[width=1.385in]{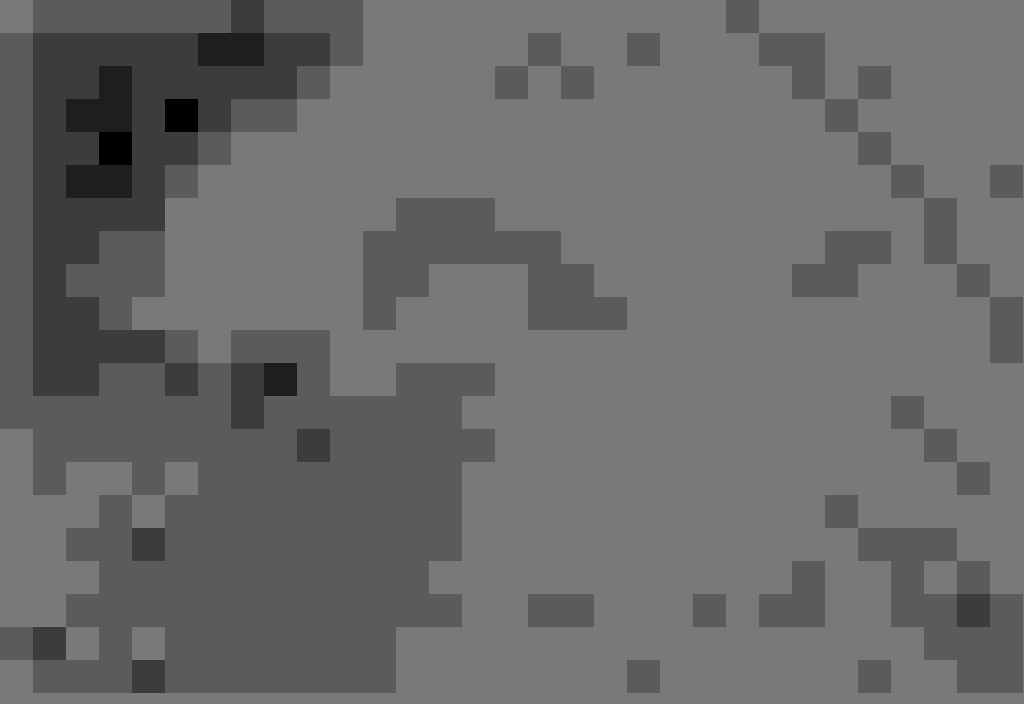}
\\
\vspace{0.03in}
\hspace{1.384in}
\includegraphics[width=1.385in]{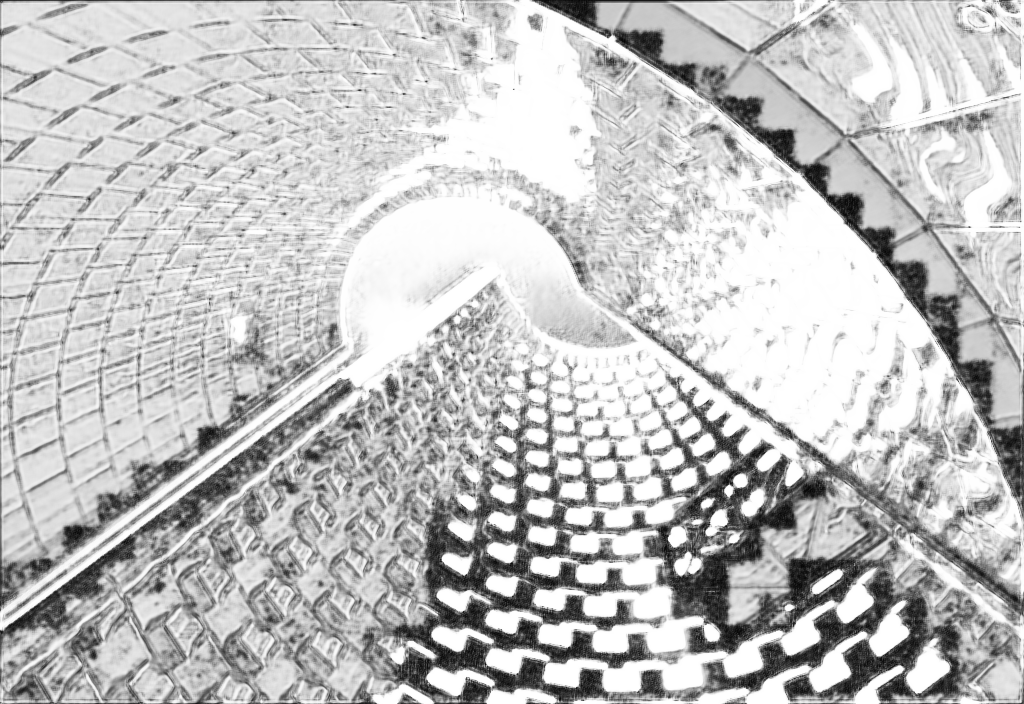} 
\includegraphics[width=1.385in]{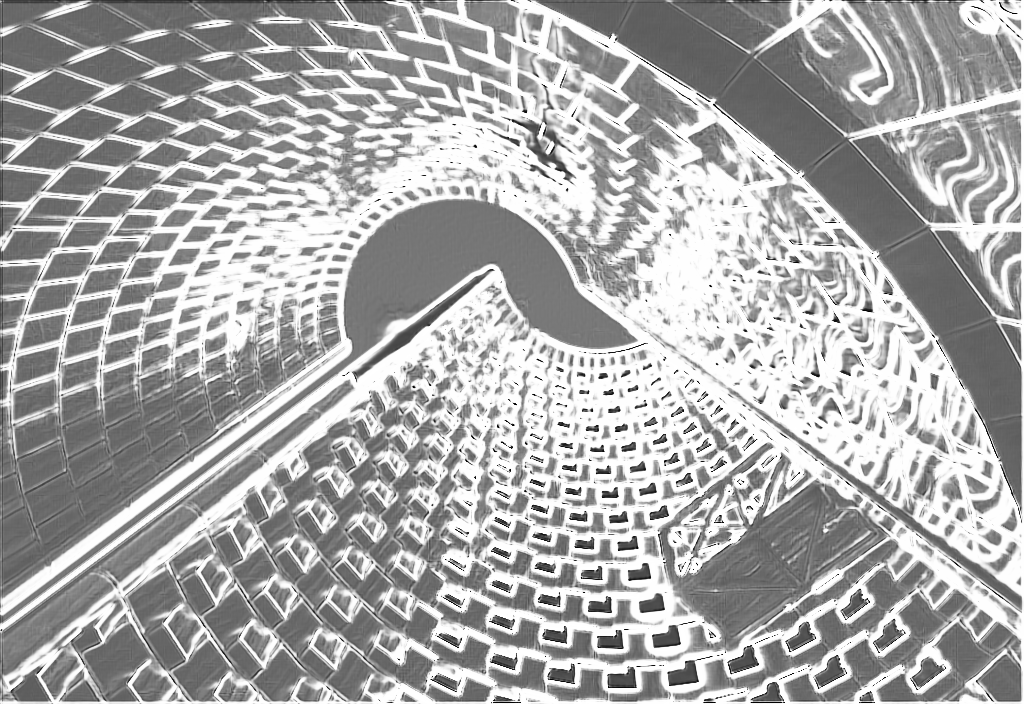} 
\includegraphics[width=1.385in]{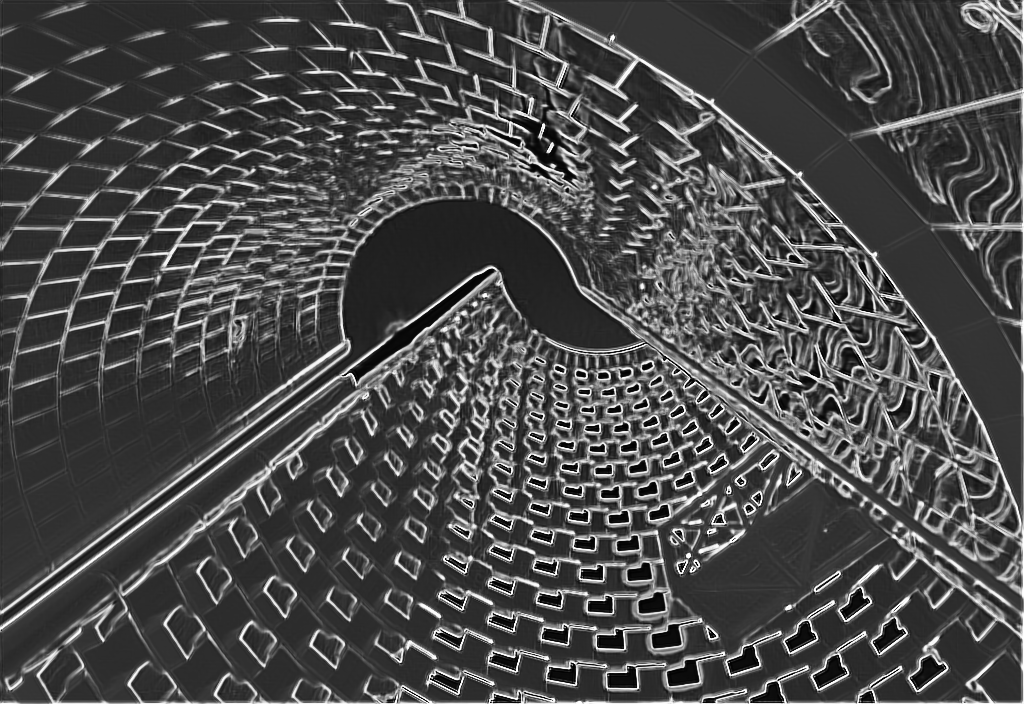} 
\includegraphics[width=1.385in]{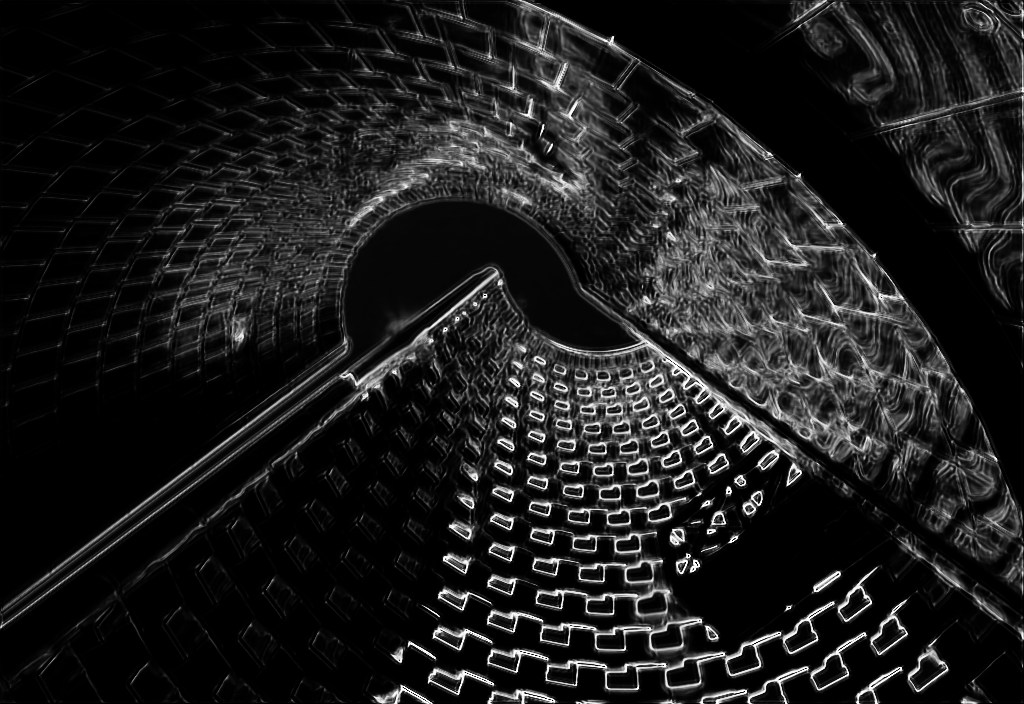}
\\
\vspace{-2.4in}
\hspace{-5.78in}
\includegraphics[width=1.385in]{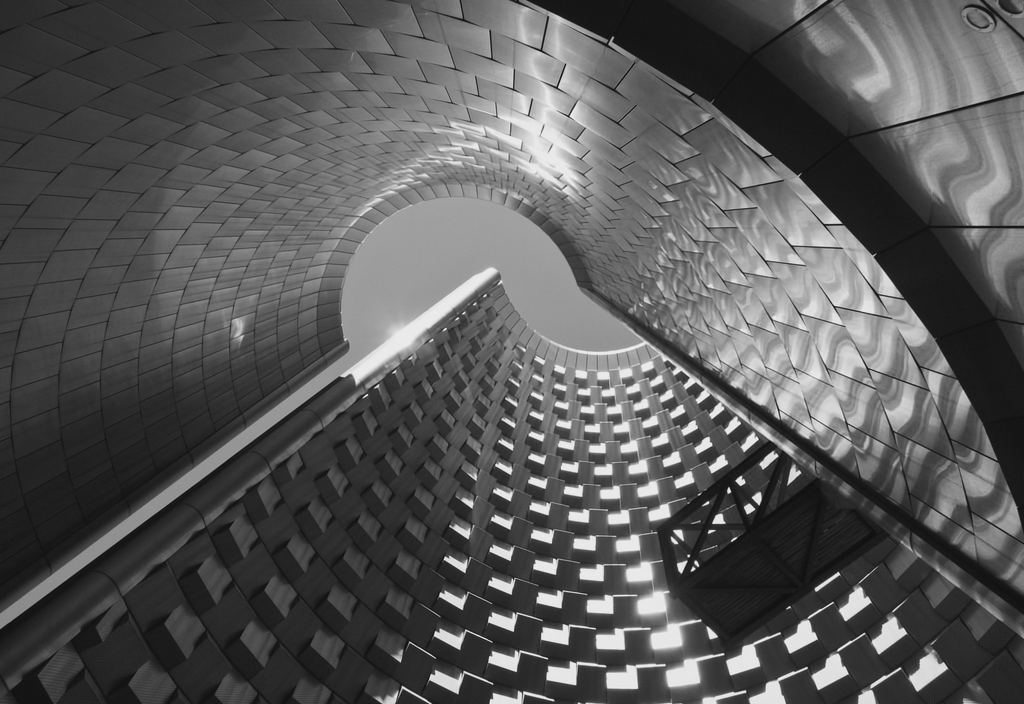}  

\vskip -0.01in \hskip -5.7in \scriptsize{(a)}

\vskip 1.34in \hskip 1.45in \scriptsize{(b) \quad\quad\quad\quad\quad\quad \quad\quad\quad\quad\quad\quad\quad\quad (c) \quad\quad\quad\quad\quad\quad\quad\quad\quad\quad\quad\quad \quad (d) \quad\quad\quad\quad\quad\quad\quad\quad\quad \quad\quad\quad\quad (e)}

\end{center}
\vskip -0.07in
\caption{The visualizations of the generated step sizes for different model variants. Figure (a) is the ground truth image. In (b)-(e), different rows indicate the step sizes generated by different model variants. Specifically, the first row is the learned step sizes of existing DUNs, second row indicates the step sizes generated by SSG-Net-G, third row is the step size maps generated by SSG-Net-B and the last row indicates the step size maps generated by SSG-Net. In addition, different columns of (b)-(e) indicate the step sizes from different phases (2-th, 6-th, 10-th, 14-th) of CS model.}
\vskip -0.15in
\label{fig:5}
\end{figure*}

For more comparison details, we conduct extensive experiments on the condition of five different sampling ratios: 0.01, 0.10, 0.25, 0.30 and 0.40. Considering testing data, we carry out extensive experiments on several benchmark datasets: Set5~\cite{shi2019scalable}, Set14~\cite{cui2018efficient}, Set11~\cite{eccvcs} and BSD68~\cite{8765626}, which have been widely used in many recent CS literatures. For fairness of comparison, we use the same training data and augmentation policy to fine-tune or retrain the compared CS models. Specifically, for the compared CS algorithms, when there is a pre-trained model at a given sampling rate, we directly fine tune the model using the same training data and augmentation policy. While when there is no pre-trained model, we directly train the model from scratch. Furthermore, we evaluate the reconstruction performance with two extensively used quality evaluation metrics: PSNR and SSIM in terms of various sampling ratios.

\begin{table}[h]
\centering
\vspace{-0.03in}
\caption{Average PSNR comparisons of different representative deep network-based CS algorithms using gaussian random sampling matrix on dataset Set11.}
\label{tab:45}
\vspace{-0.07in}
\small
\begin{tabular}{p{2.16cm}<{\centering} | p{0.81cm}<{\centering} | p{0.81cm}<{\centering}  | p{0.81cm}<{\centering} | p{0.81cm}<{\centering} | p{0.81cm}<{\centering}}
\toprule
\diagbox{Alg.}{PSNR}{Rate} & 0.10 & 0.25 & 0.30 & 0.40 & Avg.\\

\midrule
ReconNet~\cite{7780424}&24.07&26.38&28.72&30.59&27.44\\
I-Recon~\cite{2018Convolutional}&25.97&28.52&31.45&32.26&29.55\\
DR$^{2}$-Net~\cite{2019DR2}&24.71&--\ --&30.52&31.20&--\ --\\
DPA-Net~\cite{9199540}&26.99&32.38&33.35&35.21&31.98\\
NL-CSNet~\cite{9635679}&27.24&31.86&33.41&35.73&32.06\\
\hline
IRCNN~\cite{8099783}&23.05&28.42&29.55&31.30&28.08\\
LDAMP~\cite{2017Learned}&24.94&--\ --&32.01&34.07&--\ --\\
ISTA-Net~\cite{8578294}&26.49&32.48&33.81&36.02&32.20\\
DPDNN~\cite{8481558}&26.23&31.71&33.16&35.29&31.60\\
NN~\cite{8878159}&23.90&29.20&30.26&32.31&28.92\\
MAC-Net~\cite{eccvcs}&27.68&32.91&33.96&36.18&32.68\\
iPiano-Net~\cite{SU2020115989}&28.05&33.53&34.78&37.00&33.34\\

\midrule

DUN-CSNet&\textbf{29.64}&\textbf{34.97}&\textbf{36.15}&\textbf{38.12}&\textbf{34.72}\\
\bottomrule

\end{tabular}
\vspace{-0.02in}
\label{tab:45}
\end{table}

\subsubsection{Comparisons with DBNs}

For the compared DBNs, the target image is directly reconstructed from the measurements by using a well-designed deep neural network in a rude manner. The experimental results against the compared DBNs on the given testing datasets are shown in Tables~\ref{tab:1} and~\ref{tab:3}, from which we can observe that the proposed network outperforms these black box CS methods by a large margin. In the compared DBNs, since the methods SCSNet, CSNet$^{+}$ and NL-CSNet achieve the best reconstructed quality, we mainly analyze the experimental results compared to these three CS algorithms. Specifically, 1) On the dataset Set11, the proposed DUN-CSNet achieves on average 2.16dB, 2.39dB, 1.10dB and 0.0263, 0.0266, 0.0035 gains in PSNR and SSIM compared against these three DBNs at the given sampling ratios. 2) On the dataset Set14, our proposed framework achieves on average 1.68dB, 1.83dB, 1.09dB and 0.0244, 0.0256, 0.0061 gains in PSNR and SSIM under different sampling ratios. The visual comparisons are displayed in Figs.~\ref{fig:444}~\ref{fig:555}~\ref{fig:111111}, from which we observe that the proposed DUN-CSNet is capable of preserving more structural details compared with these representative deep black box CS methods.

\begin{figure*}[t]
\begin{center}
\hspace{1.384in}
\includegraphics[width=1.385in]{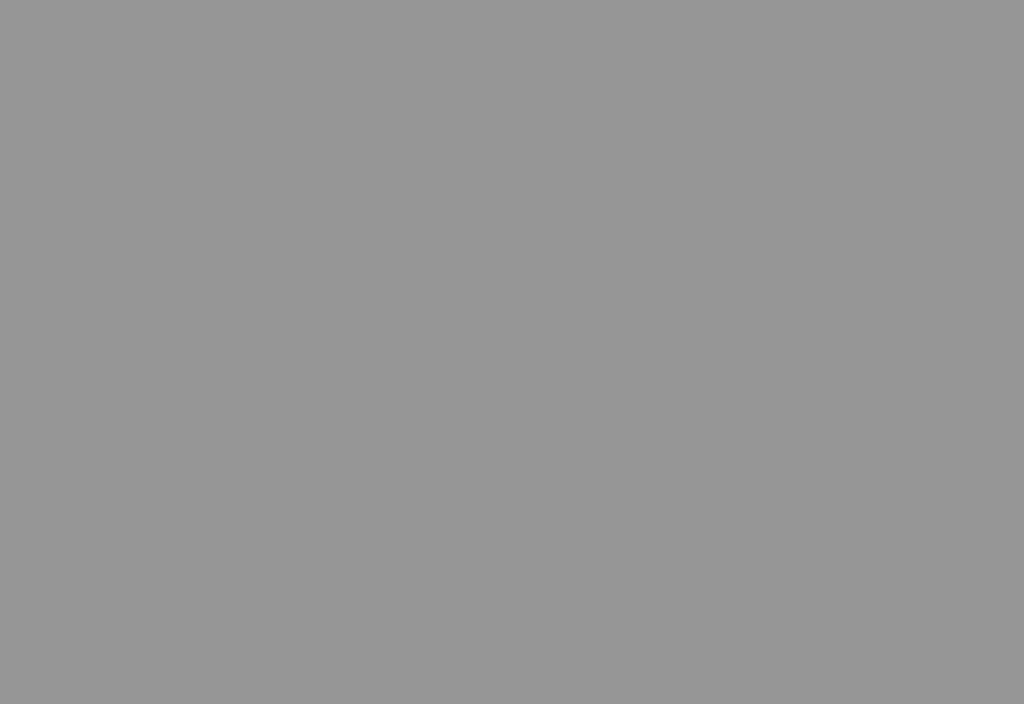} 
\includegraphics[width=1.385in]{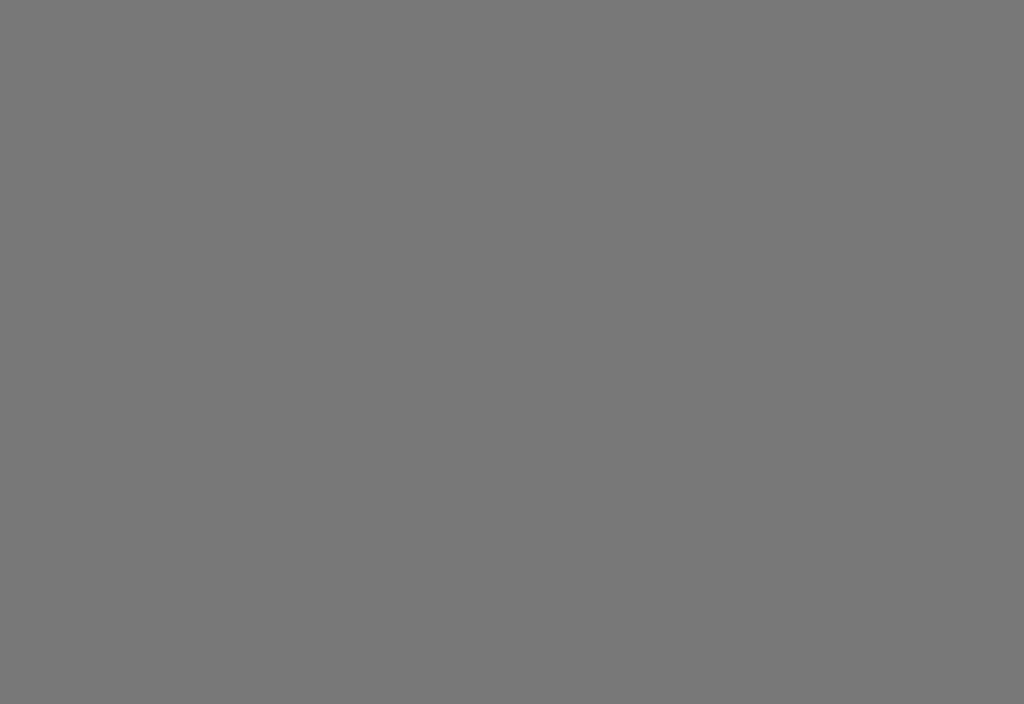} 
\includegraphics[width=1.385in]{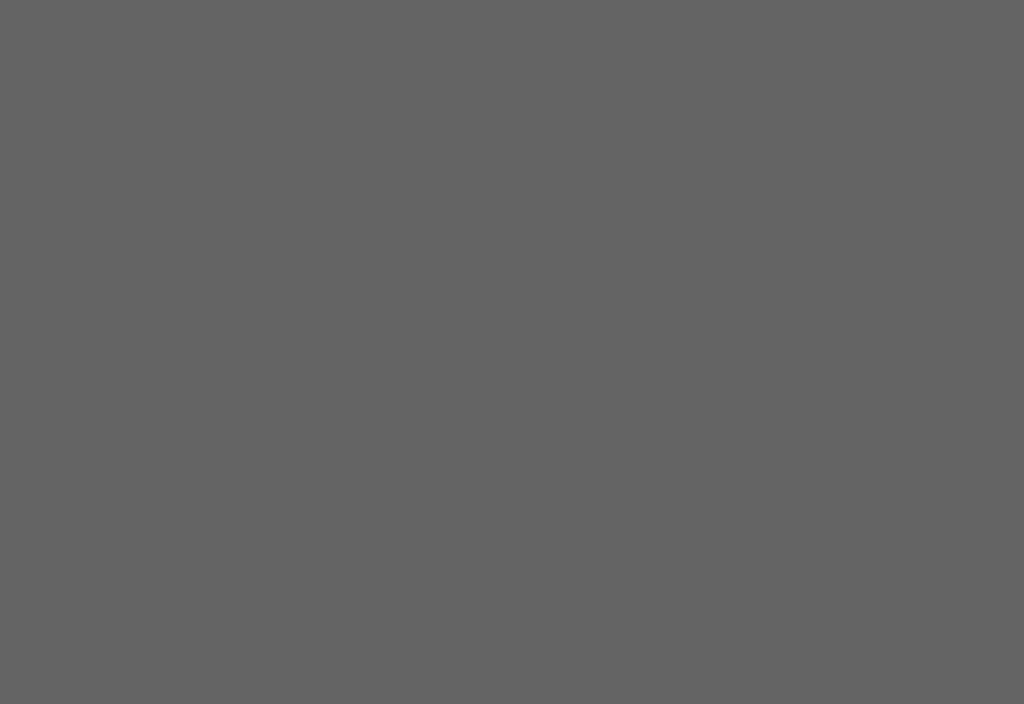} 
\includegraphics[width=1.385in]{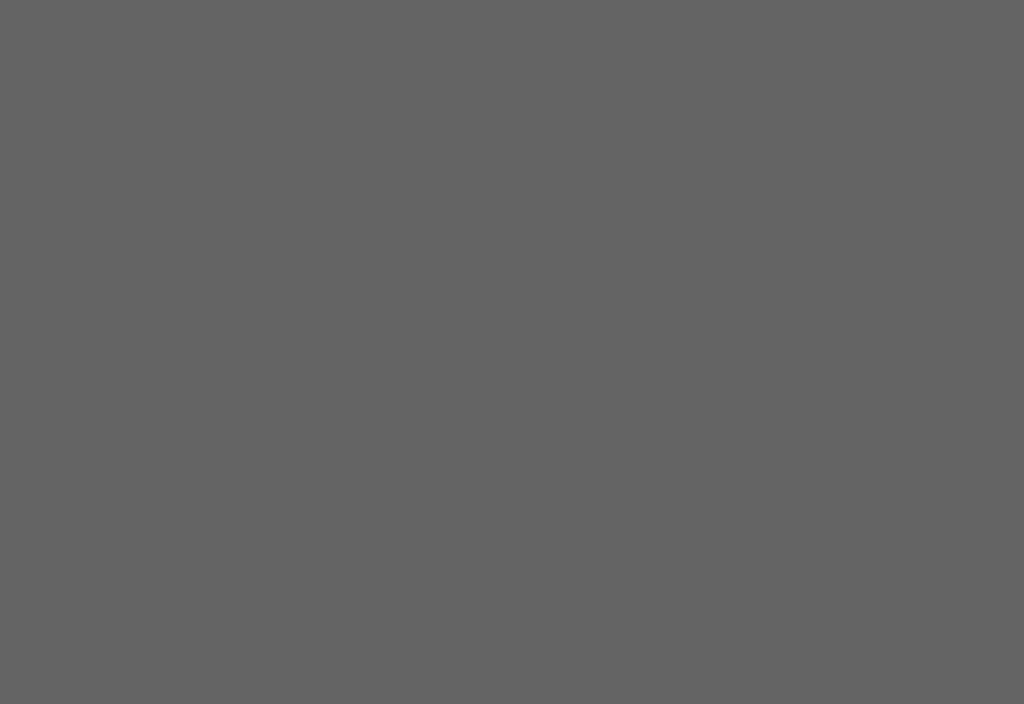} 
\\
\vspace{0.03in}
\hspace{1.384in}
\includegraphics[width=1.385in]{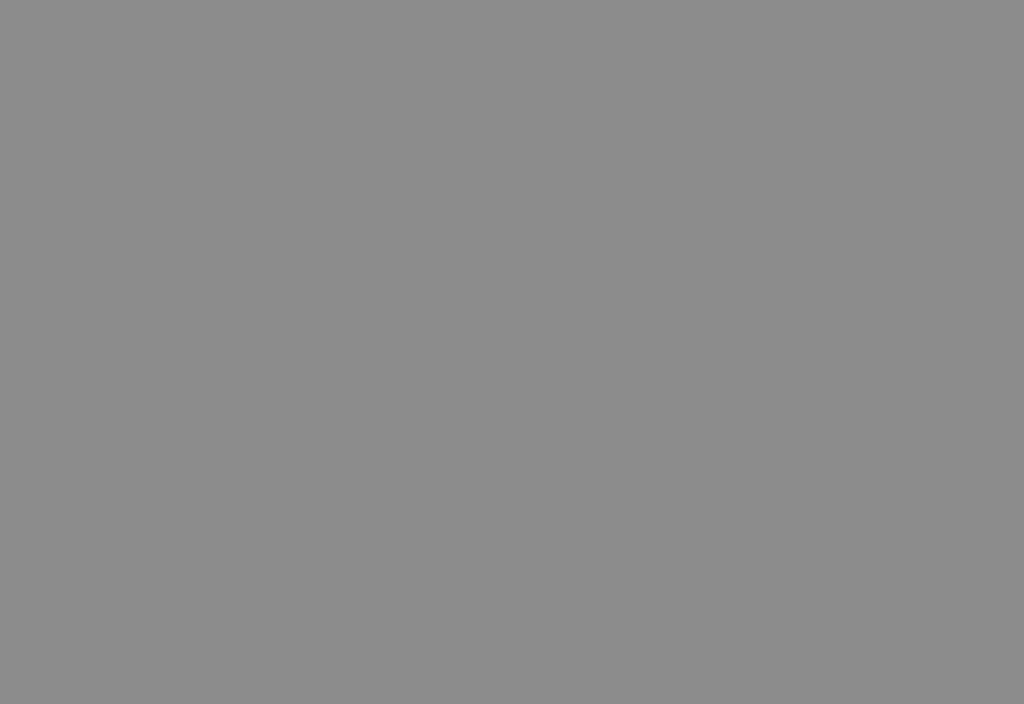} 
\includegraphics[width=1.385in]{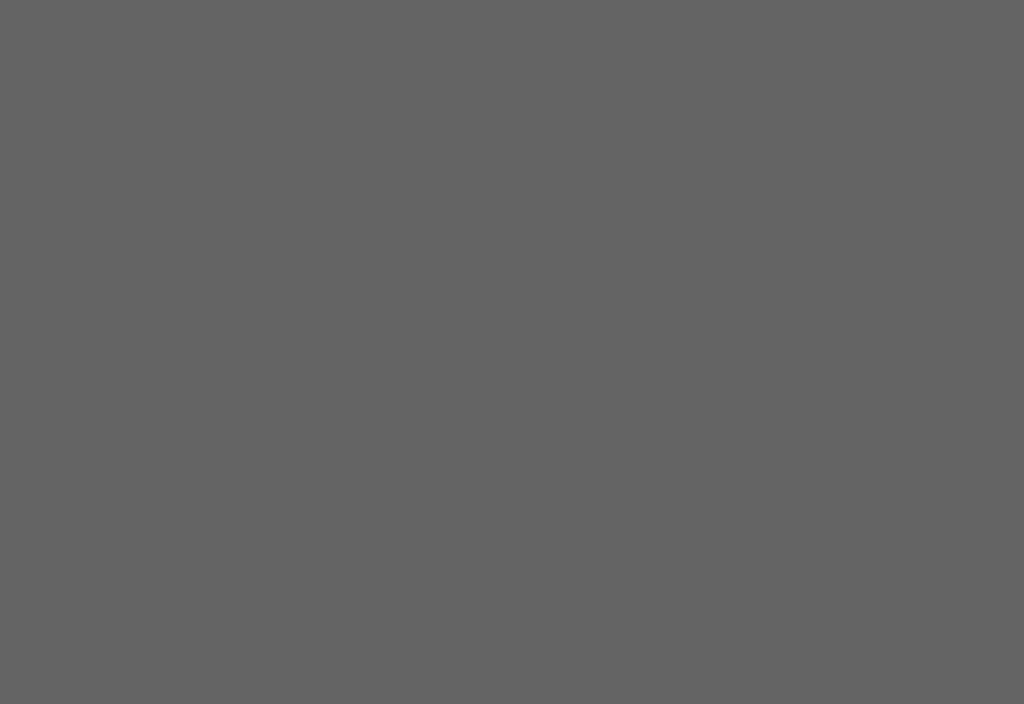} 
\includegraphics[width=1.385in]{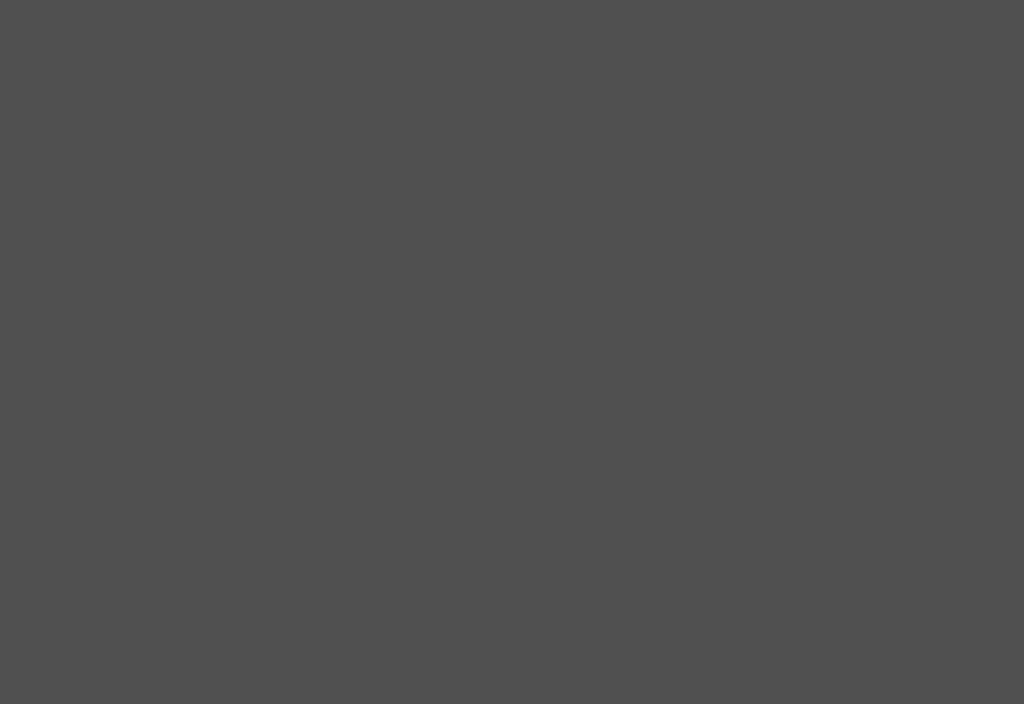} 
\includegraphics[width=1.385in]{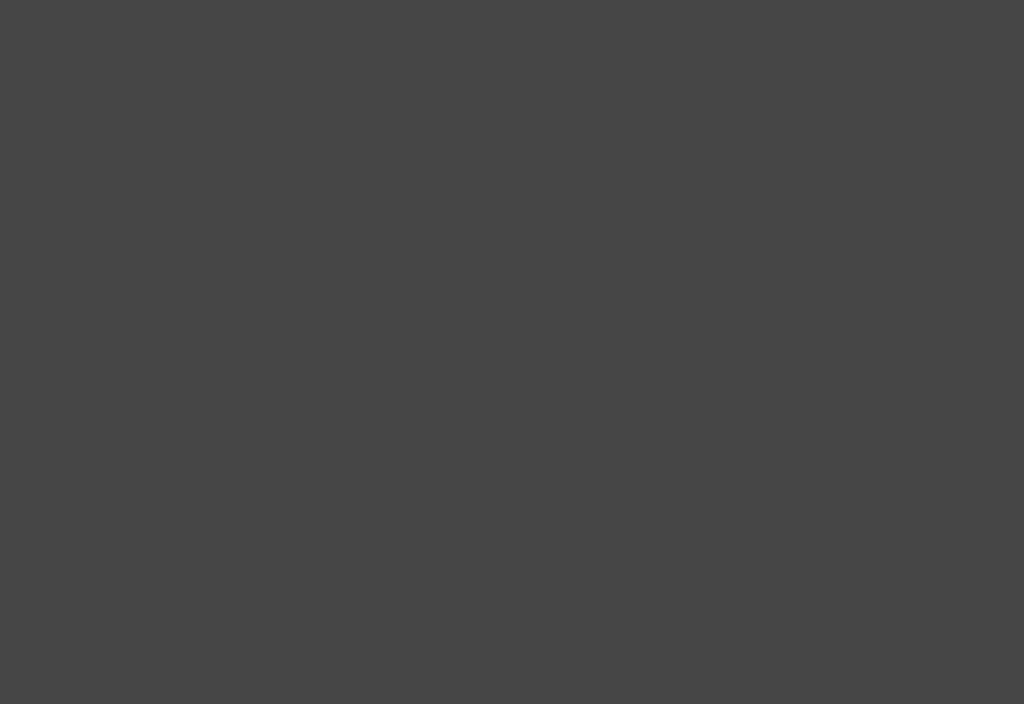}
\\
\vspace{0.03in}
\hspace{1.384in}
\includegraphics[width=1.385in]{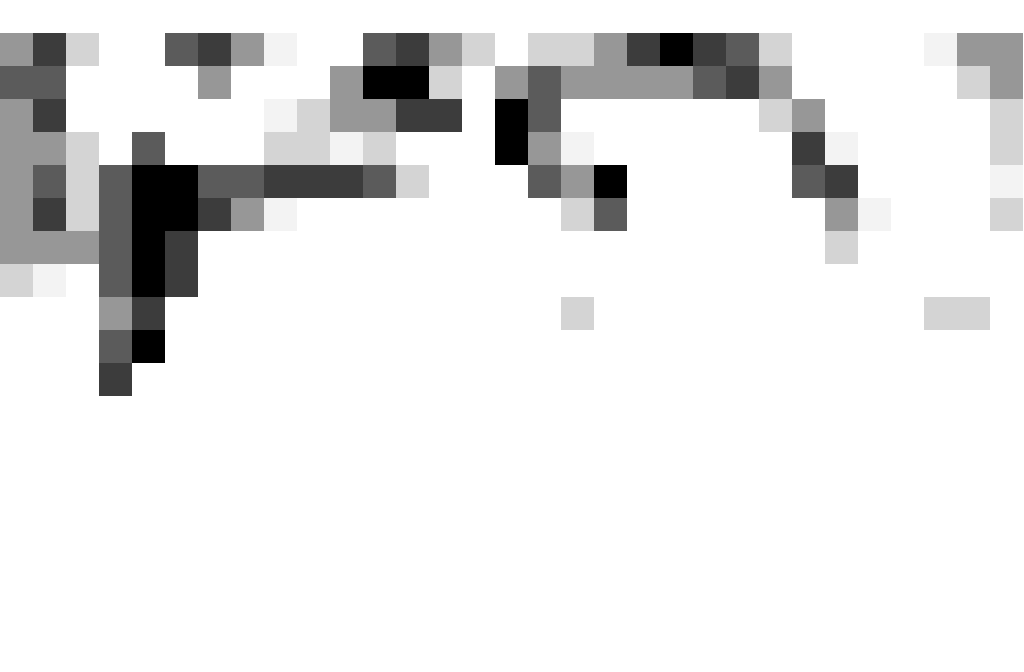} 
\includegraphics[width=1.385in]{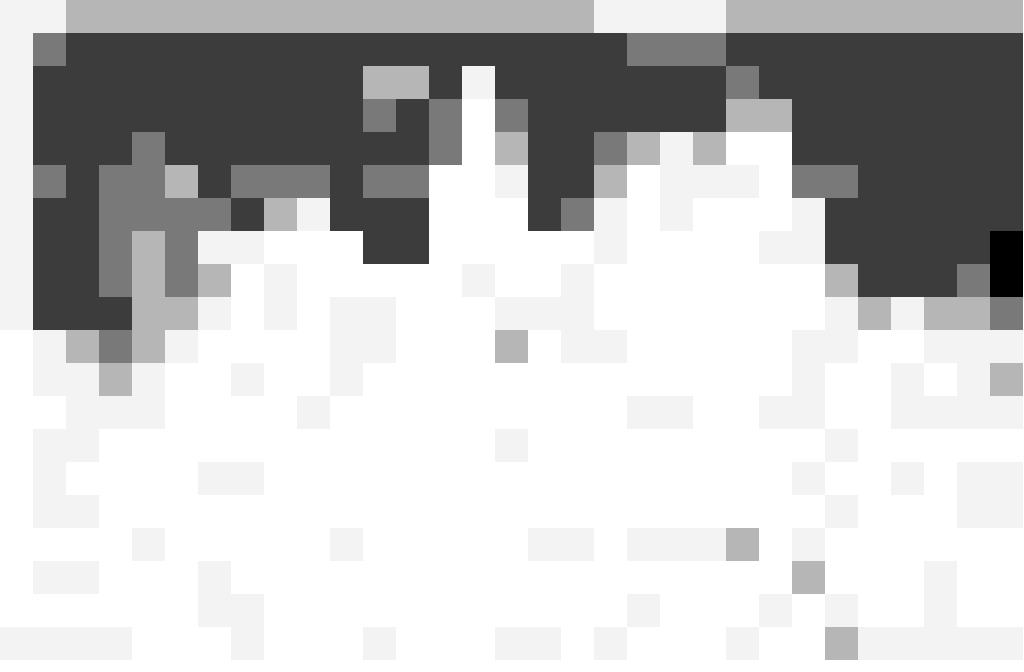} 
\includegraphics[width=1.385in]{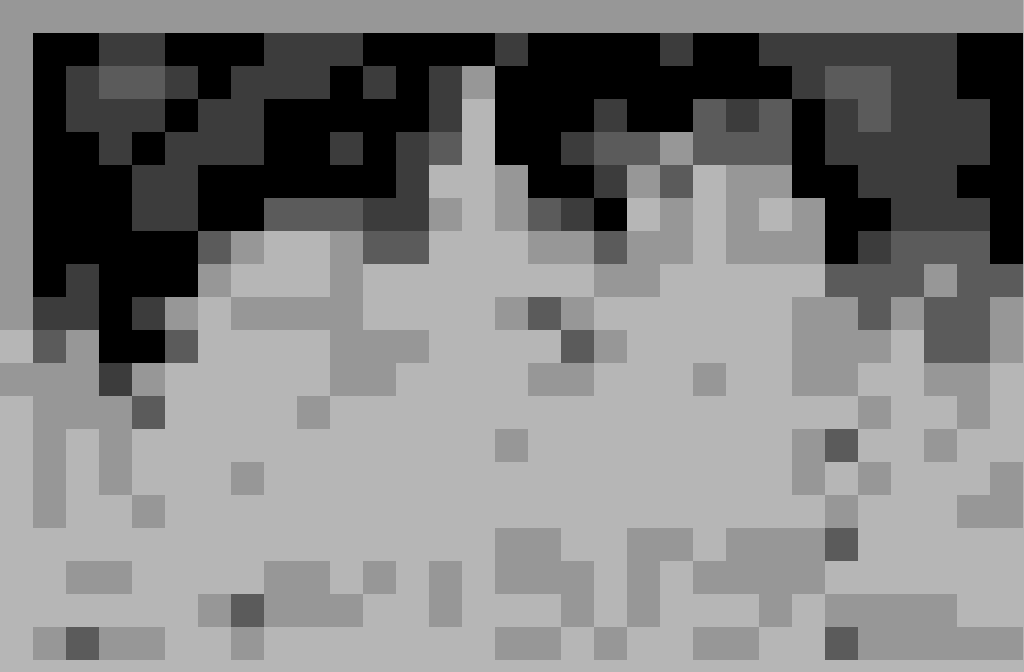} 
\includegraphics[width=1.385in]{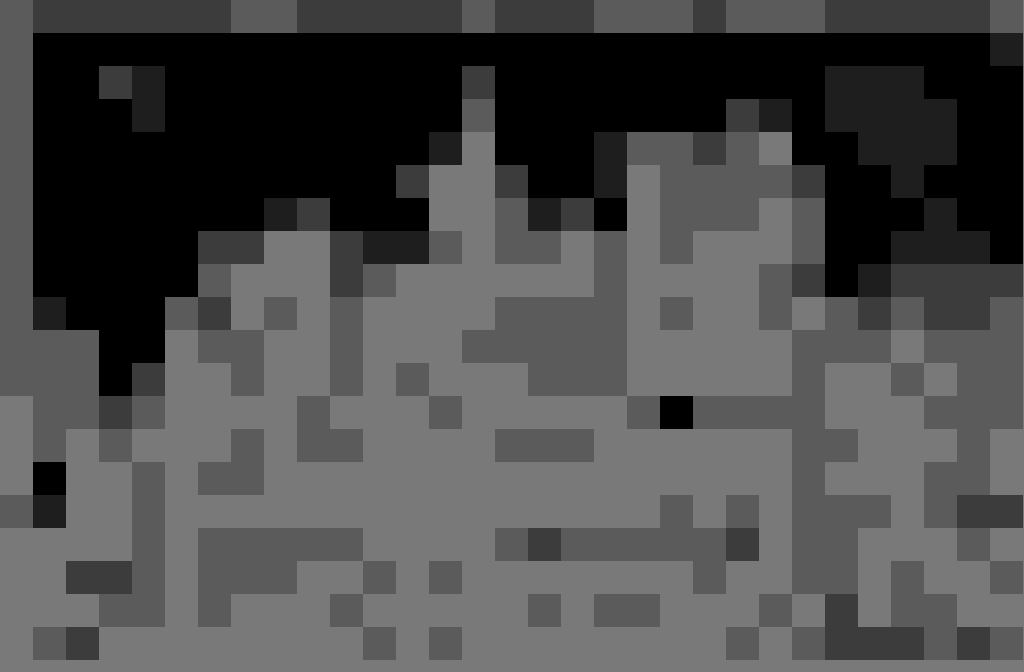}
\\
\vspace{0.03in}
\hspace{1.384in}
\includegraphics[width=1.385in]{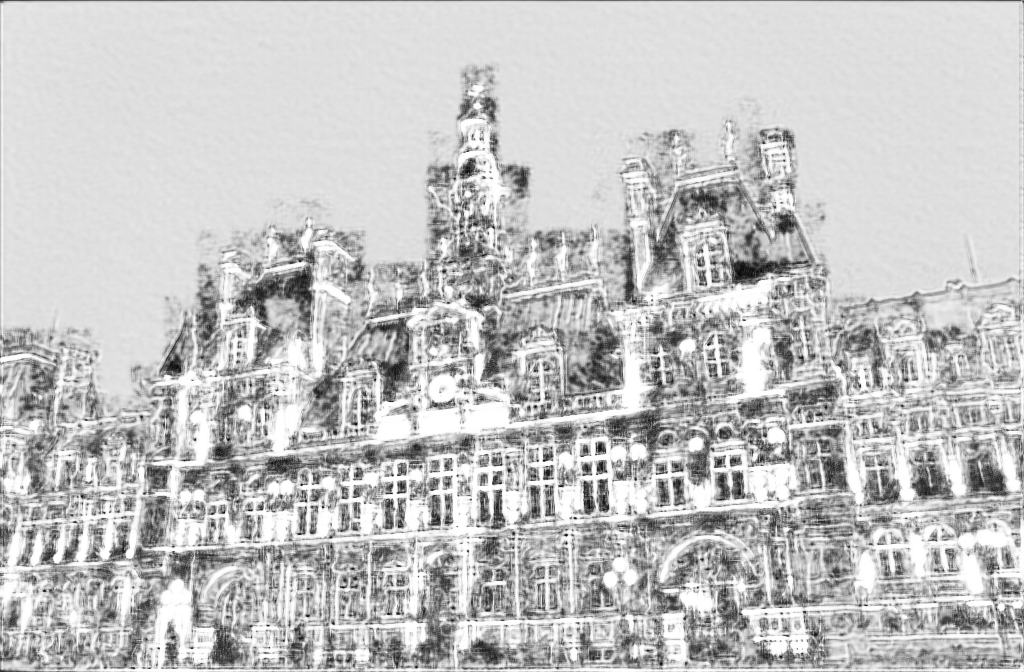} 
\includegraphics[width=1.385in]{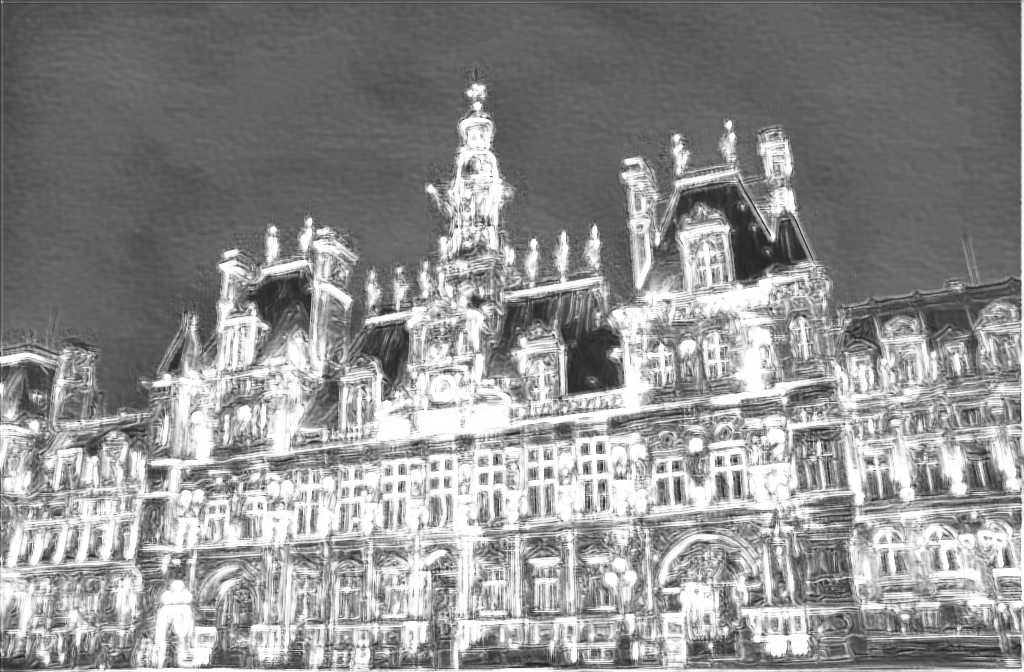} 
\includegraphics[width=1.385in]{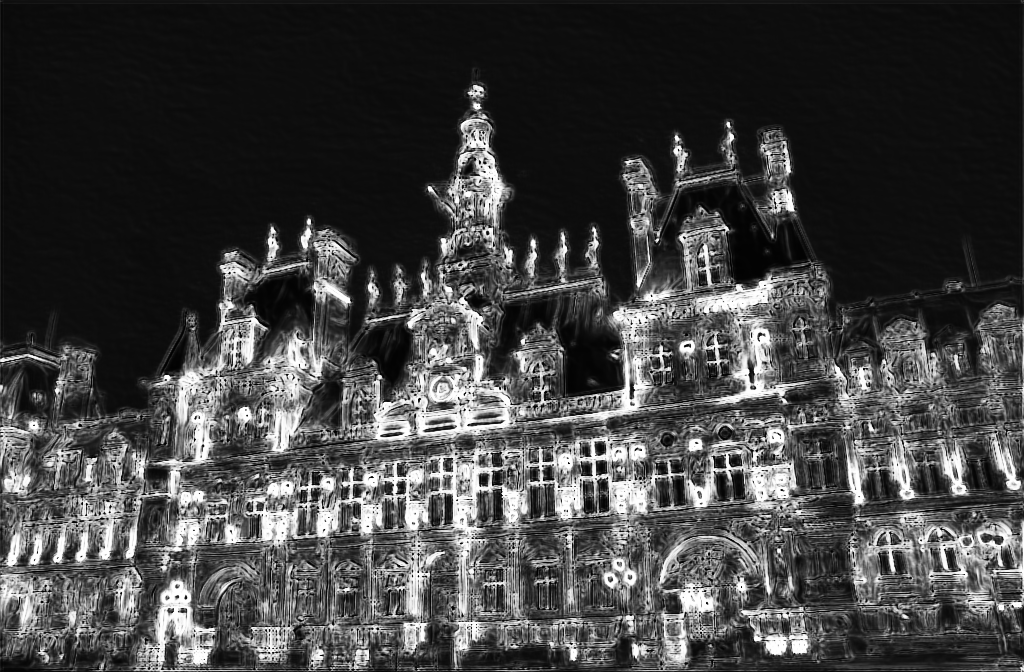} 
\includegraphics[width=1.385in]{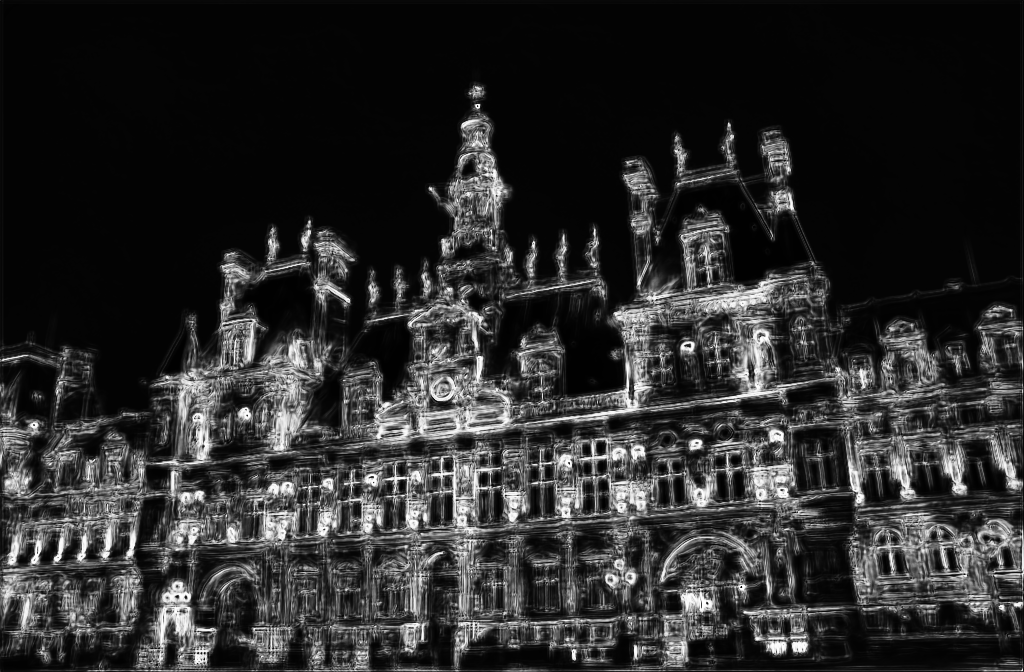}
\\
\vspace{-2.3in}
\hspace{-5.78in}
\includegraphics[width=1.385in]{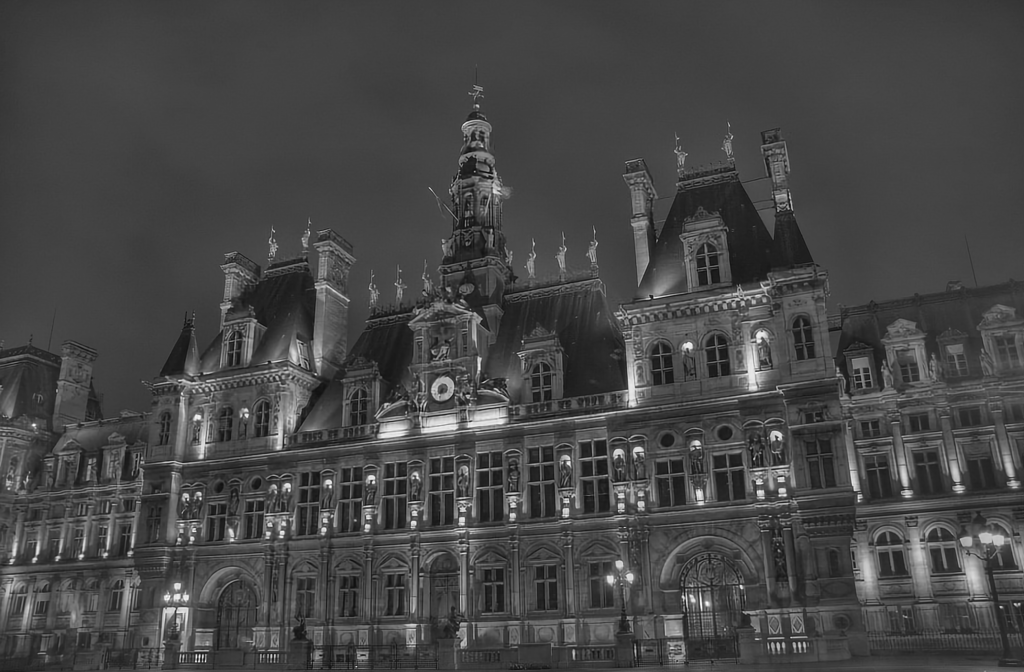}  

\vskip -0.01in \hskip -5.7in \scriptsize{(a)}

\vskip 1.28in \hskip 1.45in \scriptsize{(b) \quad\quad\quad\quad\quad\quad \quad\quad\quad\quad\quad\quad\quad\quad (c) \quad\quad\quad\quad\quad\quad\quad\quad\quad\quad\quad\quad \quad (d) \quad\quad\quad\quad\quad\quad\quad\quad\quad \quad\quad\quad\quad (e)}

\end{center}
\vskip -0.112in
\caption{The visualizations of the generated step sizes for different model variants. Figure (a) is the ground truth image. In (b)-(e), different rows indicate the step sizes generated by different model variants. Specifically, the first row is the learned step sizes of existing DUNs, second row indicates the step sizes generated by SSG-Net-G, third row is the step size maps generated by SSG-Net-B and the last row indicates the step size maps generated by SSG-Net. In addition, different columns of (b)-(e) indicate the step sizes from different phases (2-th, 6-th, 10-th, 14-th) of CS model.}
\vskip -0.15in
\label{fig:4}
\end{figure*}

\begin{table}[h]

\centering
\caption{Average running time (in seconds) of different CS algorithms for reconstructing a 256 $\times$ 256 image.}
\label{tab:69}
\vspace{-0.08in}
\begin{tabular}{p{1.98cm}<{\centering} | p{1.15cm}<{\centering}  p{1.1cm}<{\centering} | p{1.2cm}<{\centering}  p{1.2cm}<{\centering}}
\toprule

\multirow{2}*{\footnotesize{Algorithm}} & \multicolumn{2}{c|}{\footnotesize{Rate=0.01}} & \multicolumn{2}{c}{\footnotesize{Rate=0.1}}\\
\cline{2-5}

&\footnotesize{CPU}&\footnotesize{GPU}&\footnotesize{CPU}&\footnotesize{GPU}\\
\midrule

\footnotesize{TV~\cite{li2013tval3}}&\small{2.3149}&\small{---}&\small{2.6374}&\small{---}\\

\footnotesize{MH~\cite{6190204}}&\small{21.4431}&\small{---}&\small{18.9640}&\small{---}\\

\footnotesize{GSR~\cite{zhang2014group}}&\small{223.6832}&\small{---}&\small{218.9364}&\small{---}\\
\midrule

\footnotesize{CSNet~\cite{8019428}}&\small{0.1435}&\small{0.0151}&\small{0.1647}&\small{0.0159}\\

\footnotesize{SCSNet~\cite{shi2019scalable}}&\small{0.5038}&\small{0.0262}&\small{0.5180}&\small{0.0305}\\

\footnotesize{CSNet$^{+}$\footnotesize{~\cite{8765626}}}&\small{0.9146}&\small{0.0585}&\small{0.9213}&\small{0.0609}\\
\footnotesize{\hspace{-0.05in}NL-CSNet\footnotesize{~\cite{9635679}}}&\small{1.3367}&\small{0.2462}&\small{1.3614}&\small{0.2491}\\
\midrule
\footnotesize{OPINENet$^{+}$~\cite{9019857}}&\small{0.2840}&\small{0.0163}&\small{0.2974}&\small{0.0182}\\
\footnotesize{AMP-Net$^{+}$~\cite{9298950}}&\small{0.5044}&\small{0.0937}&\small{0.5185}&\small{0.1030}\\
\footnotesize{COAST~\cite{9467810}}&\small{0.7431}&\small{0.1065}&\small{0.7568}&\small{0.1175}\\
\footnotesize{MADUN~\cite{2021Memory}}&\small{0.7943}&\small{0.1144}&\small{0.8311}&\small{0.1269}\\
\midrule
\footnotesize{DUN-CSNet}&\small{0.8431}&\small{0.1267}&\small{0.8657}&\small{0.1342}\\

\bottomrule

\end{tabular}
\vspace{-0.21in}
\label{tab:69}
\end{table}

In Tables~\ref{tab:1} and~\ref{tab:3}, we find that the deep black box CS network NL-CSNet~\cite{9635679} outperforms the proposed DUN-CSNet when sampling ratio is 0.01. While with the sampling ratio increases, the proposed model achieves much better reconstruction performance against NL-CSNet. The possible explanation is provided as follows: In NL-CSNet and DUN-CSNet, the block-based sampling strategy is adopted, and the block sizes of NL-CSNet and DUN-CSNet are 32x32 and 33x33, respectively. In fact, the difference of block sizes usually leads to the inconsistency of measurement allocation. For example, when sampling rate is 0.01, the theoretical number of measurements for each 32x32 image block is 10.24. While for the 33x33 image block, the theoretical number of measurements is 10.89. Based on above, because the number of measurements must be an integer, the actual numbers of measurements for the above two different sizes of image blocks are 10 (\emph{FLOOR} operation is utilized). As above, it is clear that compared with NL-CSNet, the proposed DUN-CSNet discards more measurements in the integer operation, which weakens the performance of our DUN-CSNet. In fact, the sampling matrix of NL-CSNet is not well embedded into its reconstruction process. Contrastively, the sampling matrix of our proposed DUN-CSNet can provide the informational guidance for the image reconstruction. On the one hand, when sampling rate is 0.01, the guidance is greatly limited because the dimension of sampling matrix is very low. In this case, the limited informational guidance cannot compensate for the reconstruction loss caused by the inconsistent measurement allocation. Therefore, the algorithm NL-CSNet achieves better reconstructed quality compared with our model at sampling rate 0.01. On the other hand, with the increase of sampling rate, the dimension of sampling matrix is higher, which can provide more guidance for the reconstruction. Therefore, with the increase of sampling rate, the proposed DUN-CSNet begins to obtain better reconstruction against NL-CSNet.

\subsubsection{Comparisons with DUNs}
\vspace{-0.01in}

\begin{figure*}[t]

\begin{minipage}[t]{0.135\textwidth}
\centering
\includegraphics[width=0.95in]{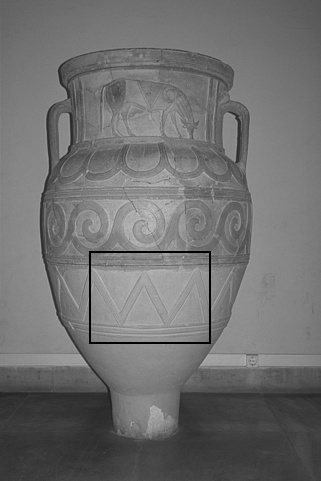}
\begin{scriptsize}
\centering
\vskip -0.52 cm \begin{tiny}GT$\backslash$PSNR$\backslash$SSIM\end{tiny}
\end{scriptsize}
\end{minipage}
\hspace{-0.012in}
\begin{minipage}[t]{0.135\textwidth}
\centering
\includegraphics[width=0.95in]{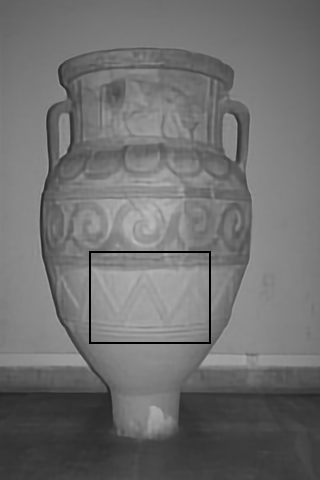}
\begin{scriptsize}
\centering
\vskip -0.52 cm \begin{tiny}CSNet$^{+}$~\cite{8765626}$\backslash$34.41$\backslash$0.8797\end{tiny}
\end{scriptsize}
\end{minipage}
\hspace{-0.012in}
\begin{minipage}[t]{0.135\textwidth}
\centering
\includegraphics[width=0.95in]{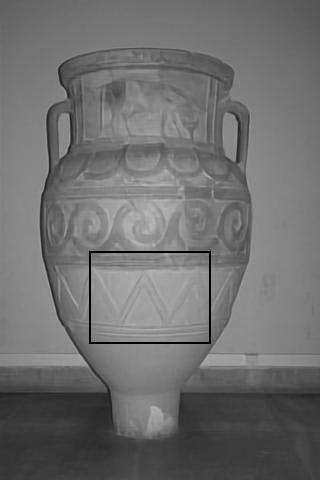}
\begin{scriptsize}
\centering
\vskip -0.52 cm \begin{tiny}NL-CSNet~\cite{9635679}$\backslash$35.44$\backslash$0.9098\end{tiny}
\end{scriptsize}
\end{minipage}
\hspace{-0.012in}
\begin{minipage}[t]{0.135\textwidth}
\centering
\includegraphics[width=0.95in]{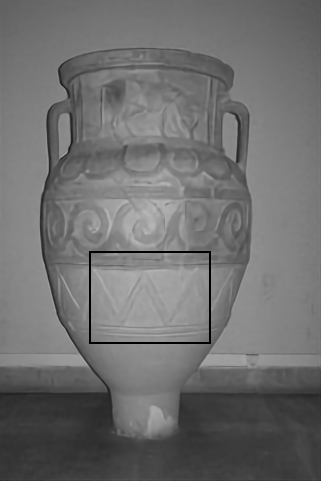}
\begin{scriptsize}
\centering
\vskip -0.52 cm \begin{tiny}OPINE-Net~\cite{9019857}$\backslash$35.24$\backslash$0.9021\end{tiny}
\end{scriptsize}
\end{minipage}
\hspace{-0.012in}
\begin{minipage}[t]{0.135\textwidth}
\centering
\includegraphics[width=0.95in]{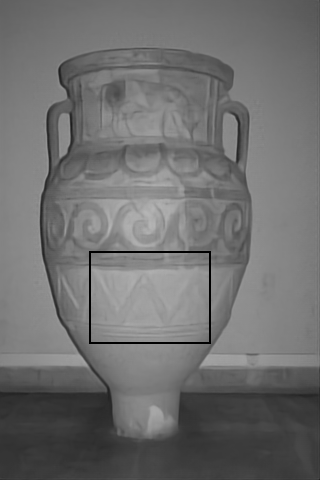}
\begin{scriptsize}
\centering
\vskip -0.52 cm \begin{tiny}AMP-Net~\cite{9298950}$\backslash$34.67$\backslash$0.8889\end{tiny}
\end{scriptsize}
\end{minipage}
\hspace{-0.012in}
\begin{minipage}[t]{0.135\textwidth}
\centering
\includegraphics[width=0.95in]{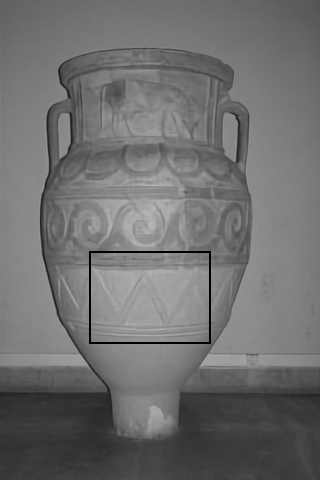}
\begin{scriptsize}
\centering
\vskip -0.52 cm \begin{tiny}MADUN~\cite{2021Memory}$\backslash$35.58$\backslash$0.9114\end{tiny}
\end{scriptsize}
\end{minipage}
\hspace{-0.012in}
\begin{minipage}[t]{0.135\textwidth}
\centering
\includegraphics[width=0.95in]{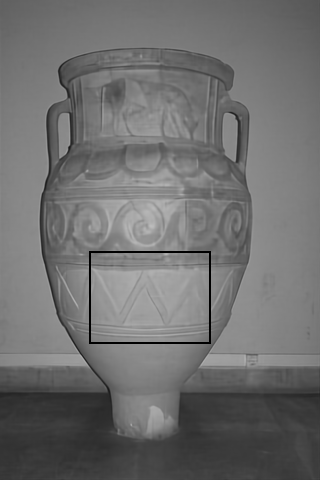}
\begin{scriptsize}
\centering
\vskip -0.52 cm \begin{tiny}DUN-CSNet$\backslash$\textbf{36.09}$\backslash$\textbf{0.9196}\end{tiny}
\end{scriptsize}
\end{minipage}

\begin{minipage}[t]{0.135\textwidth}
\centering
\includegraphics[width=0.95in]{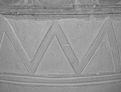}
\end{minipage}
\hspace{-0.012in}
\begin{minipage}[t]{0.135\textwidth}
\centering
\includegraphics[width=0.95in]{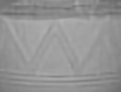}
\begin{scriptsize}
\end{scriptsize}
\end{minipage}
\hspace{-0.012in}
\begin{minipage}[t]{0.135\textwidth}
\centering
\includegraphics[width=0.95in]{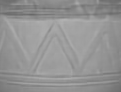}
\begin{scriptsize}
\end{scriptsize}
\end{minipage}
\hspace{-0.012in}
\begin{minipage}[t]{0.135\textwidth}
\centering
\includegraphics[width=0.95in]{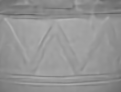}
\begin{scriptsize}
\end{scriptsize}
\end{minipage}
\hspace{-0.012in}
\begin{minipage}[t]{0.135\textwidth}
\centering
\includegraphics[width=0.95in]{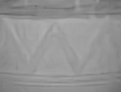}
\begin{scriptsize}
\end{scriptsize}
\end{minipage}
\hspace{-0.012in}
\begin{minipage}[t]{0.135\textwidth}
\centering
\includegraphics[width=0.95in]{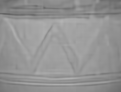}
\begin{scriptsize}
\end{scriptsize}
\end{minipage}
\hspace{-0.012in}
\begin{minipage}[t]{0.135\textwidth}
\centering
\includegraphics[width=0.95in]{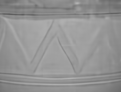}
\begin{scriptsize}
\end{scriptsize}
\end{minipage}

\vspace{-0.15in}
\caption{Visual quality comparisons of different deep network-based CS methods on one sample image from dataset BSD68 under the sampling rate 0.10.}
\vspace{-0.16in}
\label{fig:111111}
\end{figure*}

Inspired by the perspective of certain iterative optimizers, the compared DUNs usually inherit a well-designed cascaded multi-phase structure to gradually reconstruct the target image. Tables~\ref{tab:1},~\ref{tab:3}~\ref{tab:25} and~\ref{tab:2} respectively show the experimental results compared with the recent DUNs on different datasets, from which we can clearly find that the proposed CS method achieves superior reconstructed quality. In the compared DUNs, the recent schemes AMP-Net, COAST and MADUN can obtain the best reconstruction performance. For simplicity, we mainly analyze the experimental results compared with these three representative CS algorithms. Specifically, 1) On the dataset Set11, the proposed DUN-CSNet achieves on average 1.39dB, 0.86dB, 0.64dB and 0.0185, 0.0084, 0.0073 gains in PSNR and SSIM compared with these three DBNs under the given sampling ratios. 2) On the dataset Set14, our proposed framework achieves on average 1.08dB, 0.70dB, 0.58dB and 0.0189, 0.0066, 0.0061 gains in PSNR and SSIM in terms of different sampling ratios. 3) On the dataset Set5, the proposed framework achieves on average 1.15dB, 0.82dB, 0.68dB and 0.0118, 0.0073, 0.0064 gains compared against the three deep unfolding CS methods. 4) On the dataset BSD68, the proposed DUN-CSNet achieves on average 0.60dB, 0.53dB, 0.59dB and 0.0165, 0.0060, 0.0068 gains compared against the best three DUNs. More visual comparisons are shown in Figs.~\ref{fig:444}~\ref{fig:555}~\ref{fig:111111}, from which we observe that the proposed method is capable of preserving more details and retaining sharper edges compared to the recent representative deep unfolding CS methods.

\begin{figure*}[t]
\vskip -0.04in
\begin{center}
\vspace{0.04in}
\includegraphics[width=1.135in]{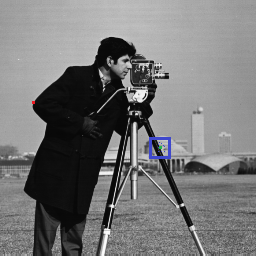}
\hspace{-0.2in}\includegraphics[width=0.55in]{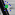}
\hspace{0.18in}
\includegraphics[width=1.135in]{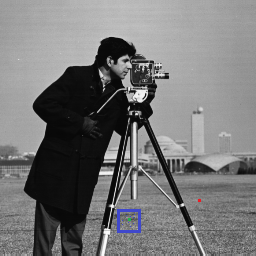}
\hspace{-0.2in}\includegraphics[width=0.55in]{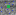}
\hspace{0.18in}
\includegraphics[width=1.135in]{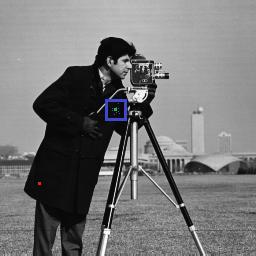}
\hspace{-0.2in}\includegraphics[width=0.55in]{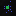}
\hspace{0.18in}
\includegraphics[width=1.135in]{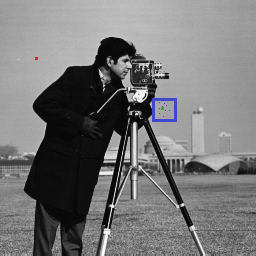}
\hspace{-0.2in}\includegraphics[width=0.55in]{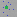}
\\
\vspace{-0.08in}
\scriptsize{(a) \quad\quad\quad\quad\quad\quad\quad\quad\quad\quad\quad\quad\quad\quad\quad\quad\quad (b) \quad\quad\quad\quad\quad\quad\quad\quad\quad\quad\quad\quad\quad\quad\quad\quad (c) \quad\quad\quad\quad\quad\quad\quad\quad\quad\quad\quad\quad\quad\quad\quad\quad\quad (d)
}\normalsize{}

\end{center}
\vskip -0.13in
   \caption{The visual results of the perception field for the non-local modules NLM and DINLM. In (a)-(d), red points are the current image patches (i.e., $\mathbf{x}_{i}$), green points are the referenced image patches in NLM, blue points are the learned positions for resampling in our proposed DINLM. The thumbnails in the bottom right corner show the enlarged views of perception.}
\vskip -0.03in
\label{fig:64}
\end{figure*}

\begin{figure*}[t]
\vskip -0.04in
\begin{center}
\includegraphics[width=1.135in]{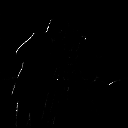}
\includegraphics[width=1.135in]{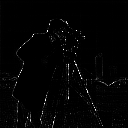}
\hspace{-0.01in}
\includegraphics[width=1.135in]{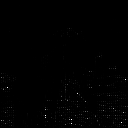}
\includegraphics[width=1.135in]{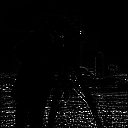}
\hspace{-0.01in}
\includegraphics[width=1.135in]{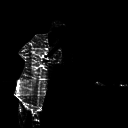}
\includegraphics[width=1.135in]{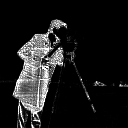}
\\
\vspace{0.04in}
\includegraphics[width=1.135in]{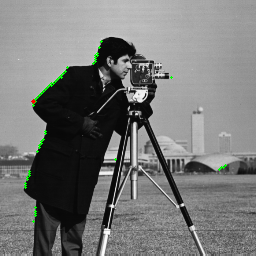}
\includegraphics[width=1.135in]{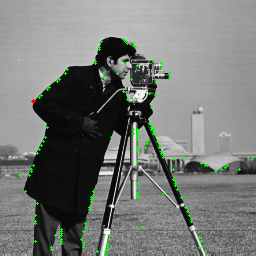}
\hspace{-0.01in}
\includegraphics[width=1.135in]{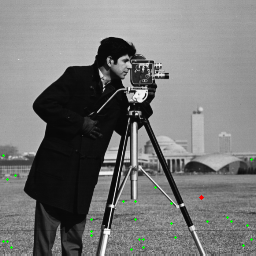}
\includegraphics[width=1.135in]{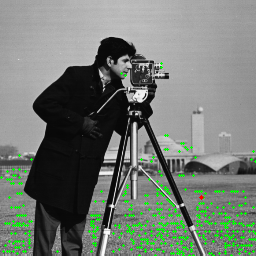}
\hspace{-0.01in}
\includegraphics[width=1.135in]{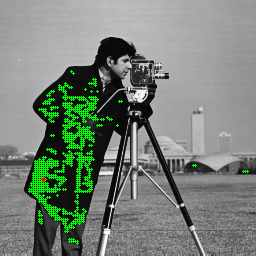}
\includegraphics[width=1.135in]{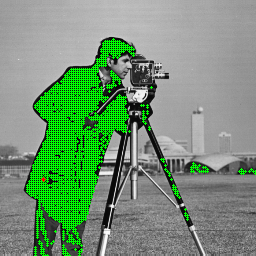}
\\
\vspace{-0.08in}
\scriptsize{(a) \quad\quad\quad\quad\quad\quad\quad\quad\quad\quad\quad (b) \quad\quad\quad\quad\quad\quad\quad\quad\quad\quad\quad (c) \quad\quad\quad\quad\quad\quad\quad\quad\quad\quad\quad (d)
\quad\quad\quad\quad\quad\quad\quad\quad\quad\quad\quad (e) \quad\quad\quad\quad\quad\quad\quad\quad\quad\quad\ (f)}\normalsize{}

\end{center}
\vskip -0.12in
   \caption{The visual comparisons of the learned affinity matrices (top) between the existing NLM and the proposed DINLM. The bottom images show their corresponding highly responsive positions on the original images. Specifically, (a), (c), (e) are the visual results of the traditional NLM, and (b), (d), (f) are the corresponding visual results of the proposed DINLM. The red points are the current locations and the green points are the corresponding positions mapped from the highlighted elements of the learned affinity matrices. (a), (b) are edge areas, (c), (d) are texture areas, and (e), (f) are smooth areas.}
\vskip -0.16in
\label{fig:6}
\end{figure*}

In the above compared CS methods, the sampling matrix is jointly optimized with the reconstruction process. However, in some practical applications, such as implementing CS in optics, or MRI system, the sampling matrix usually do not allow for such optimization. As above, to further evaluate the performance of the proposed CS framework, we conduct the experimental comparisons against some recent deep network-based CS methods that use Gaussian random sampling matrix. Specifically, we compare our proposed DUN-CSNet with twelve recent random matrix-based deep CS reconstruction algorithms, including five deep black box CS networks (ReconNet~\cite{7780424}, I-Recon~\cite{2018Convolutional}, DR$^{2}$-Net~\cite{2019DR2}, DPA-Net~\cite{9199540} and NL-CSNet~\cite{9635679}) and seven deep unfolding CS networks (IRCNN~\cite{8099783}, LDAMP~\cite{2017Learned}, ISTA-Net$^{+}$~\cite{8578294}, DPDNN~\cite{8481558}, NN~\cite{8878159}, MAC-Net~\cite{eccvcs} and iPiano-Net~\cite{SU2020115989}). In our experiments, the orthogonalized Gaussian random matrix~\cite{9199540,8999514} is utilized, and during the training process, the pre-defined random sampling matrix remains unchanged. Table~\ref{tab:45} presents the average PSNR comparisons under the given four sampling ratios (i.e., 0.10, 0.25, 0.30 and 0.40) on dataset Set11, from which we can observe that the proposed DUN-CSNet outperforms all the other compared methods in PSNR by a large margin. In addition, by comparing Tables~\ref{tab:45} and~\ref{tab:1}, we can observe that the learned sampling matrix achieves on average 1.23dB, 1.13dB, 1.24dB, 1.41dB gains (PSNR) compared with the Gaussian random sampling matrix at the given four sampling rates.

\begin{figure}[h]
\vskip -0.05in
\begin{center}
\includegraphics[width=3.4in]{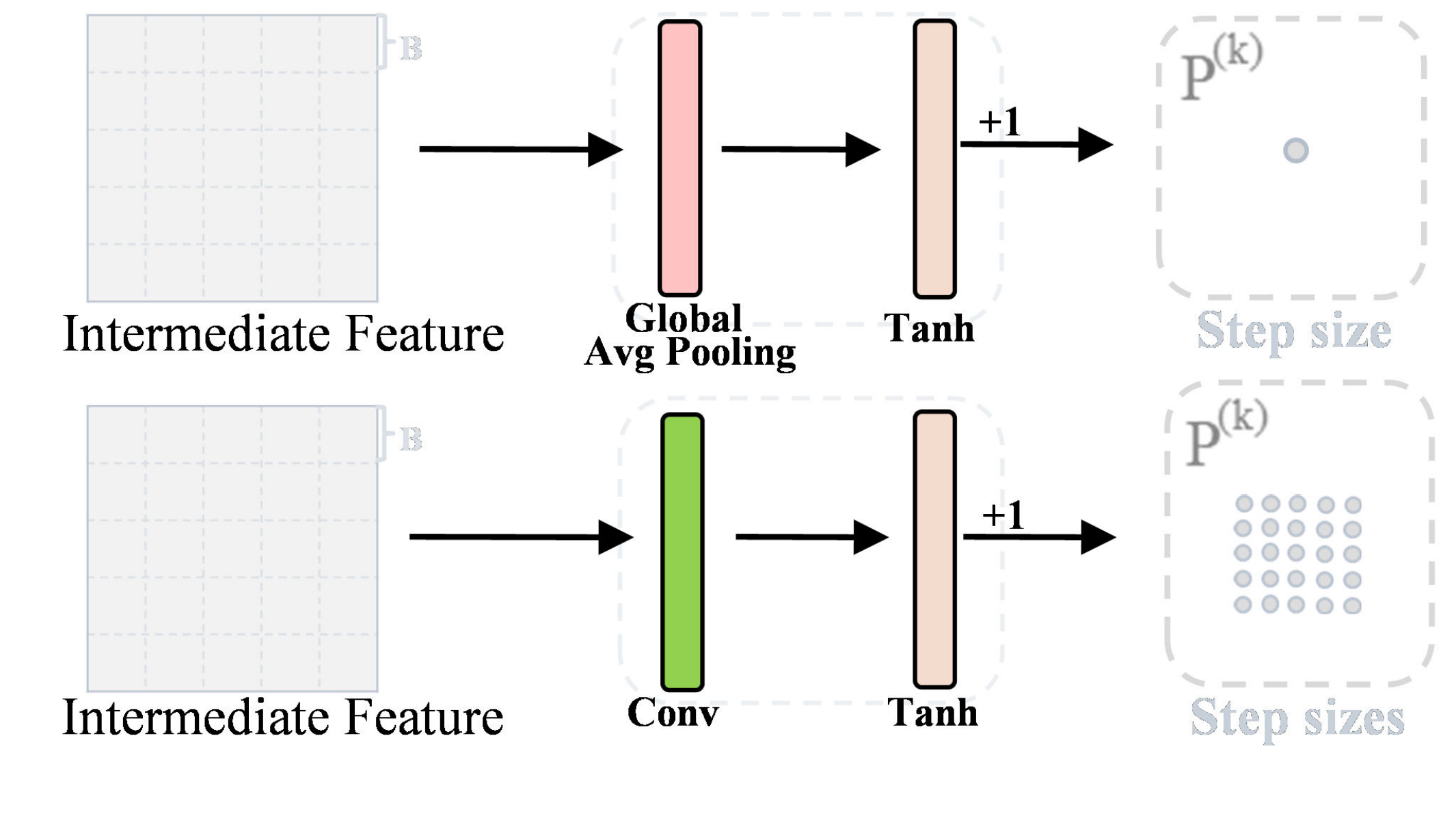}
\end{center}
\vskip -0.22in
   \caption{Top and bottom figures separately are the network structure of the normalization modules in the designed step size generation sub-networks SSG-Net-G and SSG-Net-B. The arrows indicate the flow of data representations.}
\vskip -0.08in
\label{fig:250}
\end{figure}

To verify the efficiency of the proposed DUN-CSNet, we also compare the reconstruction speed of different CS methods. Specifically, we perform all CS algorithms on the same platform with 3.30 GHz Intel i7 CPU plus NVIDIA GTX 3090 GPU. Table~\ref{tab:69} shows the average running time comparisons (in second) between different CS methods (including the optimization-based and deep network-based CS algorithms) for reconstructing a 256$\times$256 image at two sampling rates of 0.01 and 0.10. In addition, the optimization-based CS schemes are implemented based on CPU device. In contrast, we test all the deep network-based CS methods on both the CPU and GPU. The running speed comparisons show that the deep network-based methods run faster than the optimization-based methods. Furthermore, the proposed DUN-CSNet remains the same order of magnitude as the other existing deep network-based methods and achieves a faster reconstruction compared to the optimization-based CS algorithms.

\vspace{-0.1in}
\subsection{Ablation Studies and Discussions}
\vspace{-0.02in}

As mentioned above, the proposed DUN-CSNet achieves superior reconstruction performance compared to other CS methods. In order to evaluate the contribution of each part of the proposed framework, we design several counterpart variants of the proposed model, in which certain functional parts are selectively discarded or replaced. Specifically, in network CA-GDN, the introduced step size generation sub-network is analyzed in detail, and in network DN-PMN, the proposed DINLM versus vanilla NLM are discussed.

\begin{table}[b]
\vspace{-0.15in}
  \caption{The ablation results (PSNR) in network CA-GDN at various sampling rates. (``G'',``B'',``F'' respectively indicate sub-networks SSG-Net-G, SSG-Net-B and SSG-Net.)}
  \label{tab:5}
  \vspace{-0.1in}

  \begin{tabular}{p{1.08cm}<{\centering} | p{0.46cm}<{\centering}  p{0.46cm}<{\centering}  p{0.47cm}<{\centering} |  p{0.74cm}<{\centering}  p{0.74cm}<{\centering} p{0.74cm}<{\centering} p{0.74cm}<{\centering}}
    \toprule

    \small{Network}&\footnotesize{G}&\footnotesize{B}&\footnotesize{F}&\small{0.01}&\small{0.10}&\small{0.25}&\small{0.40}\\
    \midrule
    \multirow{4}*{\small{\footnotesize{CA-GDN}}} & \scriptsize{\XSolidBrush} & \scriptsize{\XSolidBrush} &\scriptsize{\XSolidBrush} & 22.35 & 28.89 & 33.04 & 36.32\\
    & \checkmark & \scriptsize{\XSolidBrush} & \scriptsize{\XSolidBrush} & 22.39 & 28.92 & 33.12 & 36.35\\
    & \scriptsize{\XSolidBrush} & $\checkmark $ & \scriptsize{\XSolidBrush} & 22.44 & 29.01& 33.18& 36.43\\

    & \scriptsize{\XSolidBrush} & \scriptsize{\XSolidBrush} & $\checkmark $  & 22.74 & 29.33 & 33.52& 36.71\\
  \bottomrule
\end{tabular}
\vskip -0.02in
\label{tab:5}
\end{table}

\begin{table}[b]
\vskip -0.06in
  \caption{The ablation results (PSNR) in network DN-PMN at various sampling rates.}
  \label{tab:6}
  \vspace{-0.08in}

  \begin{tabular}{p{1.08cm}<{\centering} | p{0.9cm}<{\centering}  p{0.9cm}<{\centering} |  p{0.74cm}<{\centering}  p{0.74cm}<{\centering}  p{0.74cm}<{\centering}  p{0.74cm}<{\centering}}
    \toprule

    \small{Network}&\footnotesize{NLM}&\footnotesize{DINLM}&\small{0.01}&\small{0.10}&\small{0.25}&\small{0.40}\\
    \midrule
    \multirow{3}*{\small{\footnotesize{DN-PMN}}} & \scriptsize{\XSolidBrush} &\scriptsize{\XSolidBrush} & 22.41& 29.03& 33.16& 36.40\\
    & \checkmark & \scriptsize{\XSolidBrush}& 22.58 & 29.17& 33.20& 36.54\\
    & \scriptsize{\XSolidBrush} & $\checkmark $ & 22.74 & 29.33 & 33.52 & 36.71\\
  \bottomrule
\end{tabular}
\label{tab:6}
\end{table}

In network CA-GDN, the proposed step size generation sub-network (SSG-Net) can densely produce the corresponding step sizes for the entire full-resolution pixels of input image. For the completeness of the experiments, another two step size generation sub-networks with different textural granularities are
designed. \textbf{1)} block content-based step size generation sub-network (SSG-Net-B): generating corresponding step sizes for different blocks of input image. \textbf{2)} global content-based step size generation sub-network (SSG-Net-G): generating corresponding step sizes for different input images. Similar with the SSG-Net as shown in Fig.~\ref{fig:2}, the sub-networks SSG-Net-B and SSG-Net-G also consist of two modules: feature extraction module and normalization module. For their feature extraction modules, the network structure retains unchanged as shown in Fig.~\ref{fig:2}, and for their normalization modules, Fig.~\ref{fig:250} shows more details of the network structure. Specifically, in normalization module of SSG-Net-B, a convolutional layer and a Tanh layer are included. It is noted that the kernel size of the convolutional layer is $B\times B$ and the stride size is $B\times B$. In normalization module of SSG-Net-G, a global pooling layer and a Tanh layer are included. In addition, because the lower bound of Tanh function is -1, we add 1 after all Tanh layers to ensure the nonnegativity of the generated step sizes.

As noted above, the output of sub-network SSG-Net-B is also a step size map, and the elements correspond to the step sizes of different non-overlapping image blocks (block size is $B\times B$). While for sub-network SSG-Net-G, the output is a single scalar value that corresponds to the step size of entire input image. Considering the three sub-networks SSG-Net, SSG-Net-B and SSG-Net-G, the ablation results on the given four testing datasets are shown in  Table~\ref{tab:5}, from which we can observe that the designed three step size generation networks can enhance the reconstructed quality to a certain extent. Besides, due to the generation of dense step sizes for the full-resolution textures, the proposed SSG-Net achieves the maximum gain. Analogously, the presented SSG-Net-G obtains the minimum gain because of its generation of single step size value. Moreover, from Table~\ref{tab:5}, we can also get that the gain brought by SSG-Net compared to SSG-Net-B is greater than that of SSG-Net-B against to SSG-Net-G.

For further intuitive analysis, Figs.~\ref{fig:5} and~\ref{fig:4} show the visualization results of step sizes generated by different variants, from which we can find that the learned step size maps generated by SSG-Net-B and SSG-Net are highly related to the content of input image, thus resulting in a content-adaptive gradient updating. Apparently, compared with the existing gradient updating strategies with the same intensity for different textures of input image, the proposed content-adaptive gradient updating is more conducive to the textural refinement of the intermediate features, which facilitates the exploring of prior knowledge by the proximal mapping network, thus enhancing the reconstructed image quality. Furthermore, from the visualization results of the generated step size maps, we can roughly observe the following two additional phenomena: 1) With the increase of the phase indexes, the elements of the generated step size maps approximately show a decreasing trend. This may be caused by the diminishing reconstructed distortion during the execution of the cascaded multi-phase framework. From another point of view, the reduction of the generated step sizes can facilitate the stable convergence of the entire cascaded model. 2) In each phase of the proposed framework, the generated step sizes of the smooth area are smaller than that of the texture area. This might mean that the smooth region is easier to learn and has a faster convergence speed compared with the texture areas.

For network DN-PMN, the proposed deformation-invariant non-local sub-module (DINLM) versus the vanilla non-local sub-module (NLM) are discussed below. Table~\ref{tab:6} shows the experimental results on the given four testing datasets, from which we can observe that both DINLM and NLM can enhance the reconstructed quality to a certain extent and the proposed DINLM further improves the reconstruction performance of NLM. For intuitive comparison, Fig.~\ref{fig:64} shows the visual results of the perception field for the non-local modules NLM and DINLM, from which we can observe that the learned resampling positions of DINLM (blue points) have a wider perception compared to the regular-shape perception positions of NLM (green points). As above, due to the optimization of the resampling indexes, the proposed DINLM has a wider perception compared against NLM. On the other hand, the affinity matrix in NLM and DINLM can actually reflect the learning ability of non-local modules for exploring the non-local self-similarity priors, and the visualizations of learned affinity matrices for NLM and DINLM are shown in Fig.~\ref{fig:6}, from which we can observe that in the affinity matrix learned from our proposed DINLM, more positions are activated (highlight area). In other words, the proposed DINLM is able to exploit richer non-local prior knowledge in certain automatically learned deformation spaces, thus enhancing CS reconstruction performance.

Compared to the existing NLM, the proposed DINLM achieves less than 0.2dB gain in PSNR (as shown in Table~\ref{tab:6}), and the possible reasons for such small improvement are analyzed as follows: \textbf{1)} a subsampling trick (similar with~\cite{2018Non}) is utilized in our proposed deformation-invariant non-local module (DINLM) for reducing computing resources, which affects the performance of our model to a certain extent. \textbf{2)} the learning efficiency and performance of DINLM is affected by the training data, parameter configuration to a certain extent, these influence factors still require more explorations. More importantly, the proposed DINLM is currently only a preliminary version, and we hope that the proposed idea can provide valuable inspirations for other researchers.

\vspace{-0.04in}
\section{Conclusion}
\label{section:a6}

In this paper, inspired by the iterative steps, i.e., gradient descent and proximal mapping, in the traditional Proximal Gradient Descent (PGD) algorithm, a novel deep unfolding network for image compressed sensing (DUN-CSNet) is proposed, which is able to exploit a novel content adaptive mechanism with deep networks to enhance the CS reconstruction performance. Specifically, for gradient descent, a well-designed step size generation sub-network (SSG-Net) is developed, which is able to dynamically allocate the corresponding step sizes for different textures of input image, realizing an adaptive gradient updating. For proximal mapping, a novel deformation-invariant deep non-local network is designed, which can adaptively build the long-range dependencies between the nonlocal patches under certain automatically learned deformation spaces, leading to a wider perception on more context priors. Extensive experiments manifest that the proposed DUN-CSNet outperforms existing state-of-the-art CS methods by large margins.

\ifCLASSOPTIONcaptionsoff
  \newpage
\fi

\bibliographystyle{IEEEtran}
\bibliography{bib/paper}

\begin{IEEEbiography}[{\includegraphics[width=1in,height=1.25in,clip,keepaspectratio]{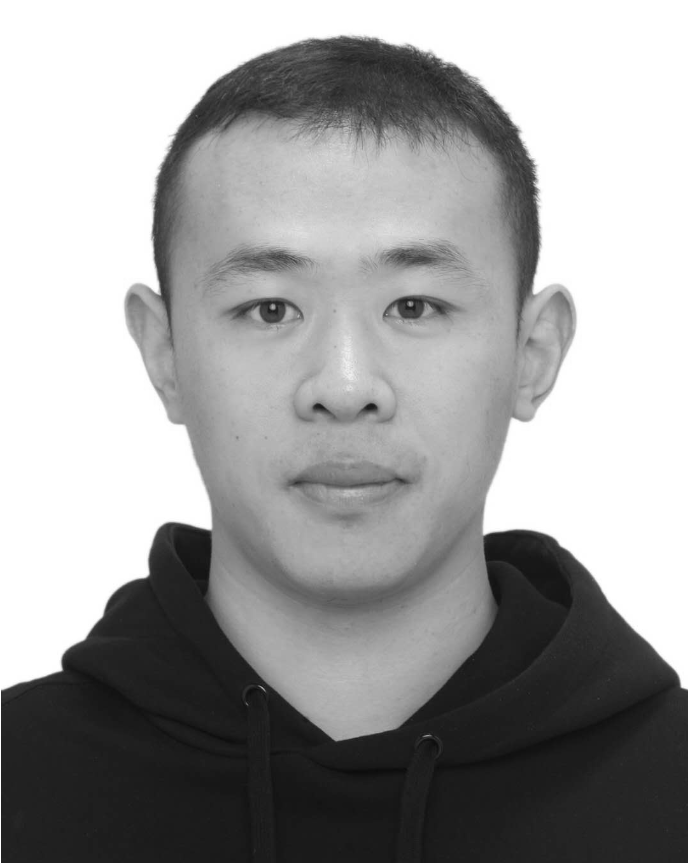}}]{Wenxue Cui}
received the B.S. degree from the Department of Mathematics, Northeast Forestry University (NEFU), Harbin, China, in 2016, and received the Ph.D. degree in computer science from Harbin Institute of Technology (HIT), Harbin, China, in 2022. He is currently an Assistant Professor with the School of Computer Science and Technology, HIT. His research interests include data compression, image and video processing, computer vision, and multimedia security.
\end{IEEEbiography}

\begin{IEEEbiography}[{\includegraphics[width=1in,height=1.25in,clip,keepaspectratio]{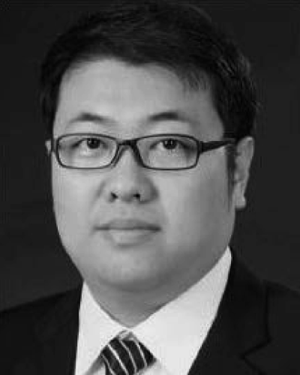}}]{Xiaopeng Fan}
received the
B.S. and M.S. degrees from the Harbin Institute
of Technology (HIT), Harbin, China, in 2001 and
2003, respectively, and the Ph.D. degree from the
Hong Kong University of Science and Technology,
Hong Kong, in 2009.

In 2009, he joined HIT, where he is currently a
Professor. From 2003 to 2005, he was with Intel
Corporation (China) as a Software Engineer. From
2011 to 2012, he was with Microsoft Research Asia
as a Visiting Researcher. From 2015 to 2016, he was
with HKUST as a Research Assistant Professor. He has authored one book
and more than 100 articles in refereed journals and conference proceedings.
His current research interests include video coding and transmission, image
processing, and computer vision. He has served as the Program Chair of
PCM2017, the Chair of the IEEE SGC2015, and the Co-Chair of MCSN2015.
He was an Associate Editor of the IEEE 1857 standard from 2012. He has
been awarded Outstanding Contributions to the Development of the IEEE
Standard 1857 by the IEEE in 2013.
\end{IEEEbiography}
\begin{IEEEbiography}[{\includegraphics[width=1in,height=1.25in,clip,keepaspectratio]{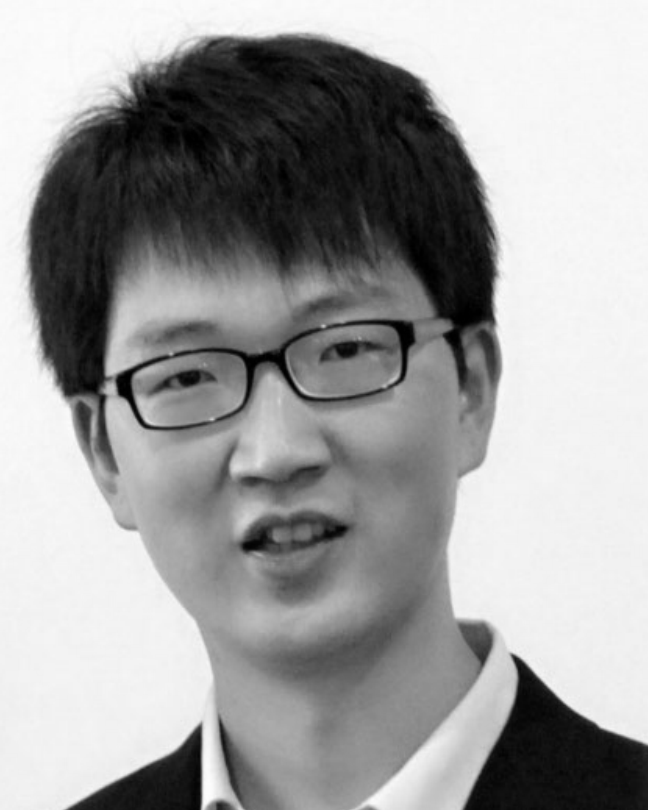}}]{Jian Zhang}
received the B.S. degree from the Department of Mathematics, Harbin
Institute of Technology (HIT), Harbin, China, in 2007, and the M.Eng. and Ph.D. degrees from the School of Computer Science and Technology, HIT, in 2009 and 2014, respectively. Currently, he is an Assistant Professor with the School of Electronic and Computer Engineering, Peking University Shenzhen Graduate School, Shenzhen, China.
His research interests include intelligent multimedia processing, deep learning, and optimization.
\end{IEEEbiography}
\begin{IEEEbiography}[{\includegraphics[width=1in,height=1.25in,clip,keepaspectratio]{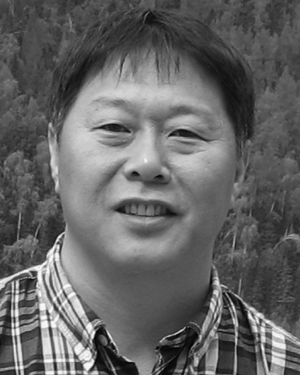}}]{Debin Zhao}
received the B.S., M.S., and Ph.D.
degrees in computer science from the Harbin Institute
of Technology (HIT), Harbin, China, in 1985,
1988, and 1998, respectively.

Since 2018, he has been with the Peng Cheng
Laboratory. He is currently a Professor with the
Department of Computer Science, HIT. He has published
over 300 technical articles in refereed journals
and conference proceedings in the areas of image and
video coding, video processing, video streaming
and transmission, and computer vision.
\end{IEEEbiography}


\vfill


\end{document}